\documentclass{naturep}

\usepackage{amssymb}
\usepackage{amsmath}
\usepackage{graphicx}
\usepackage{algorithmicx}
\usepackage{algorithm}

\usepackage{longtable}
\usepackage{booktabs}
\usepackage{amssymb}
\usepackage{pifont} 
\usepackage{caption} 

\usepackage[noend]{algpseudocode}
\usepackage{bbm}
\usepackage{lineno}
\usepackage[margin=0.6in]{geometry}
\usepackage{ragged2e}
\usepackage{setspace}
\usepackage{longtable}
\usepackage[breaklinks=true]{hyperref}
\usepackage{anyfontsize}
\usepackage{setspace}
\usepackage{float}
\usepackage{booktabs}
\usepackage{multirow}
\usepackage{xcolor}
\usepackage{blindtext}
\usepackage{tabularx}
\usepackage{caption}
\usepackage{subcaption}

\makeatletter
\let\saved@includegraphics\includegraphics
\AtBeginDocument{\let\includegraphics\saved@includegraphics}

\makeatother

\usepackage{enumitem, kantlipsum}

\title{\begin{flushleft}{\begin{spacing}{1} PathOrchestra: A Comprehensive Foundation Model for Computational Pathology with Over 100 Diverse Clinical-Grade Tasks\end{spacing}}\end{flushleft}}

\begin{document}

\maketitle
\begin{spacing}{1.8}
\vspace{-15mm}
\noindent Fang Yan$^{1*\boldsymbol{\ddag}}$, Jianfeng Wu$^{2\boldsymbol{\ddag}}$, Jiawen Li$^{3\boldsymbol{\ddag}}$, Wei Wang$^{4,5\boldsymbol{\ddag}}$, Jiaxuan Lu$^{1}$, Wen Chen$^{6}$, Zizhao Gao$^{2}$, Jianan Li$^{2}$, Heng Li$^{4,5}$, Hong Yan$^{4,5}$, Jiabo Ma$^{7}$, Minda Chen$^{6}$, Yang Lu$^{6}$, Qing Chen$^{6}$, Yizhi Wang$^{3}$, Xitong Ling$^{3}$, Xuenian Wang$^{3}$, Zihan Wang$^{9,10}$, Qiang Huang$^{9,10}$, Shengyi Hua$^{11}$, Mianxin Liu$^{1}$, Lei Ma$^{8}$, Tian Shen$^{6}$, Xiaofan Zhang$^{1,11*}$, Yonghong He$^{3*}$, Hao Chen$^{7*}$, Shaoting Zhang$^{1,11*}$, Zhe Wang$^{2,4,5*}$
\end{spacing}

\vspace{-6mm}
\begin{spacing}{1.4}
\begin{affiliations}
 \item Shanghai Artificial Intelligence Laboratory, Shanghai, 200030, China 
 \item State Key Laboratory of Holistic Integrative Management of Gastrointestinal Cancers, Department of Pathology, School of Basic Medicine and Xijing Hospital, Fourth Military Medical University, Xi’an, 710032, China 
 \item Institute of Biopharmaceutical and Health Engineering, Tsinghua Shenzhen International Graduate School, Tsinghua University, Shenzhen, 518055, China
 \item Department of Pathology, The First Affiliated Hospital of USTC, Division of Life Sciences and Medicine, University of Science and Technology of China, Hefei, 230036, China 
 \item Intelligent Pathology Institute, Division of Life Sciences and Medicine, University of Science and Technology of China, Hefei, 230036, China
 \item SenseTime Research, Shanghai, 200030, China
  \item Department of Computer Science and Engineering, Department of Chemical and Biological Engineering, Division of Life Science, Hong Kong University of Science and Technology, Hong Kong SAR, 999077, China
 \item National Biomedical Imaging Center, College of Future Technology, Peking University, Beijing 100871, China
 \item Shengqiang Technology Co.\ Ltd., Longgang District, Shenzhen 518055, China
 \item Institute of Materials Research, Shenzhen International Graduate School, Tsinghua University, Shenzhen, 518055, China
  \item Qing Yuan Research Institute, Shanghai Jiao Tong University, Shanghai, 200240, China
 \\$\boldsymbol{\ddag}$ Contributed Equally
 \\\textbf{*Corresponding author}: Fang Yan, Xiaofan Zhang, Yonghong He, Hao Chen, Shaoting Zhang, Zhe Wang 
\end{affiliations}
\end{spacing}

\clearpage

\vspace{-5mm}
\begin{spacing}{1.2}
\noindent \textbf{Abstract}\\

The complexity and variability inherent in high-resolution pathological images present significant challenges in computational pathology. While pathology foundation models leveraging artificial intelligence (AI) have catalyzed transformative advancements, their development demands large-scale datasets, considerable storage capacity, and substantial computational resources. Furthermore, ensuring their clinical applicability and generalizability requires rigorous validation across a broad spectrum of clinical tasks. In this work, we present PathOrchestra, a versatile pathology foundation model trained via self-supervised learning on a dataset comprising 300K pathological slides (262.5 TB) from 20 tissue and organ types across multiple centers. The model was rigorously evaluated on 112 clinical tasks using a combination of 61 private and 51 public datasets. These tasks encompass digital slide preprocessing, pan-cancer classification, lesion identification, multi-cancer subtype classification, biomarker assessment, gene expression prediction, and the generation of structured reports. PathOrchestra demonstrated exceptional performance across 27,755 whole slide images (WSIs) and 9,415,729 region-of-interest (ROI) images, achieving over 0.950 accuracy in 47 tasks, including pan-cancer classification across various organs, lymphoma subtype diagnosis, and bladder cancer screening. Notably, it is the first model to generate structured reports for high-incidence colorectal cancer and diagnostically complex lymphoma—areas that are infrequently addressed by foundational models but hold immense clinical potential. Overall, PathOrchestra exemplifies the feasibility and efficacy of a large-scale, self-supervised pathology foundation model, validated across a broad range of clinical-grade tasks. Its high accuracy and reduced reliance on extensive data annotation underline its potential for clinical integration, offering a pathway toward more efficient and high-quality medical services.

\end{spacing}

\newpage

\begin{spacing}{1.35}
\noindent\textbf{\large{Introduction}} 

Pathology is heralded as the gold standard for disease diagnosis, encompassing a broad range of tasks including tumor detection~\cite{tolkach2023artificial,ding2023large}, typing~\cite{lu2021data,anaya2024multiple}, grading~\cite{nagpal2020development,madabhushi2020deep}, molecular expression analysis~\cite{shamai2022deep,he2020integrating,kather2020pan,jaume2024hest}, prognosis assessment~\cite{wulczyn2021interpretable,jiang2024end}, and prediction of treatment response~\cite{foersch2023multistain,yu2016predicting,lipkova2022deep}. These tasks require handling high-resolution and morphologically diverse pathological images, posing significant diagnostic resource challenges. Computational pathology (CPath) leverages powerful algorithms to interpret and quantify pathological images, providing a more precise understanding of disease mechanisms. However, traditional AI training paradigms relying on large annotated datasets face limitations due to the impracticality of accurately annotating vast amounts of pathological data across numerous disease types. Accurately annotating vast amounts of pathological data across numerous disease types is often impractical, creating an urgent need for generalizable models that can efficiently handle multicenter data while reducing reliance on manual annotations.

The advent of self-supervised learning in computer vision (e.g., DINOv2~\cite{oquab2023dinov2}, MoCov3~\cite{chen2021empirical}, MAE~\cite{he2022masked}) and the success of large language models like the GPT series~\cite{achiam2023gpt,brown2020language} have propelled the development of data-driven pre-training methodologies. These models leverage vast collections of whole slide images (WSIs) to learn high-quality feature representations from unlabeled data, demonstrating state-of-the-art performance in downstream tasks using limited labeled data and fewer training parameters. Despite these advancements, a comprehensive evaluation of foundational pathology models is still lacking. Furthermore, the impact of pre-training data and model architecture on performance in downstream tasks remains underexplored. While technical assessments have been carried out, the clinical applicability and potential of these models remain insufficiently addressed, limiting their full integration into real-world clinical settings.

\begin{figure*}[!htbp]
	\centering
	\includegraphics[width=0.96\linewidth]{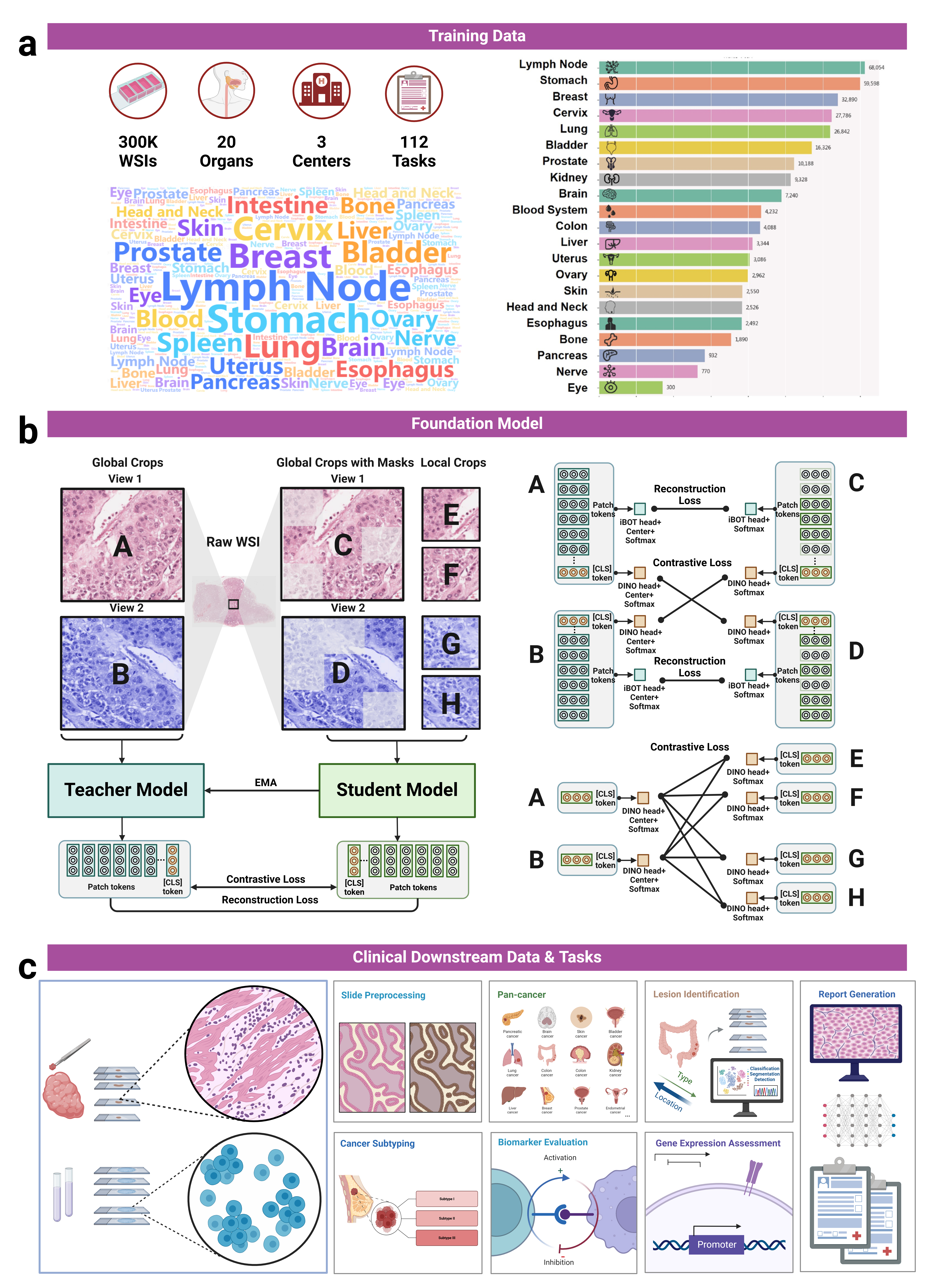}
	\caption{Overview of PathOrchestra. (Caption continued on next page.)}
	\label{fig.Fig1_Overview}
\end{figure*}

\begin{figure*}[!htbp]
	\ContinuedFloat
	\caption{(Continued) a. The foundational model was trained on 300K WSIs sourced from three centers, covering 20 human organs and tissues. b. The training strategy followed an unsupervised contrastive learning framework, utilizing a teacher model as the foundation for testing downstream tasks. c. Diverse downstream task data encompasses both histology and cytology, enabling the model to support 112 clinical tasks across seven major categories. 
	}
\end{figure*}

\begin{figure*}[!tbp]
	\centering
	\includegraphics[width=1\linewidth]{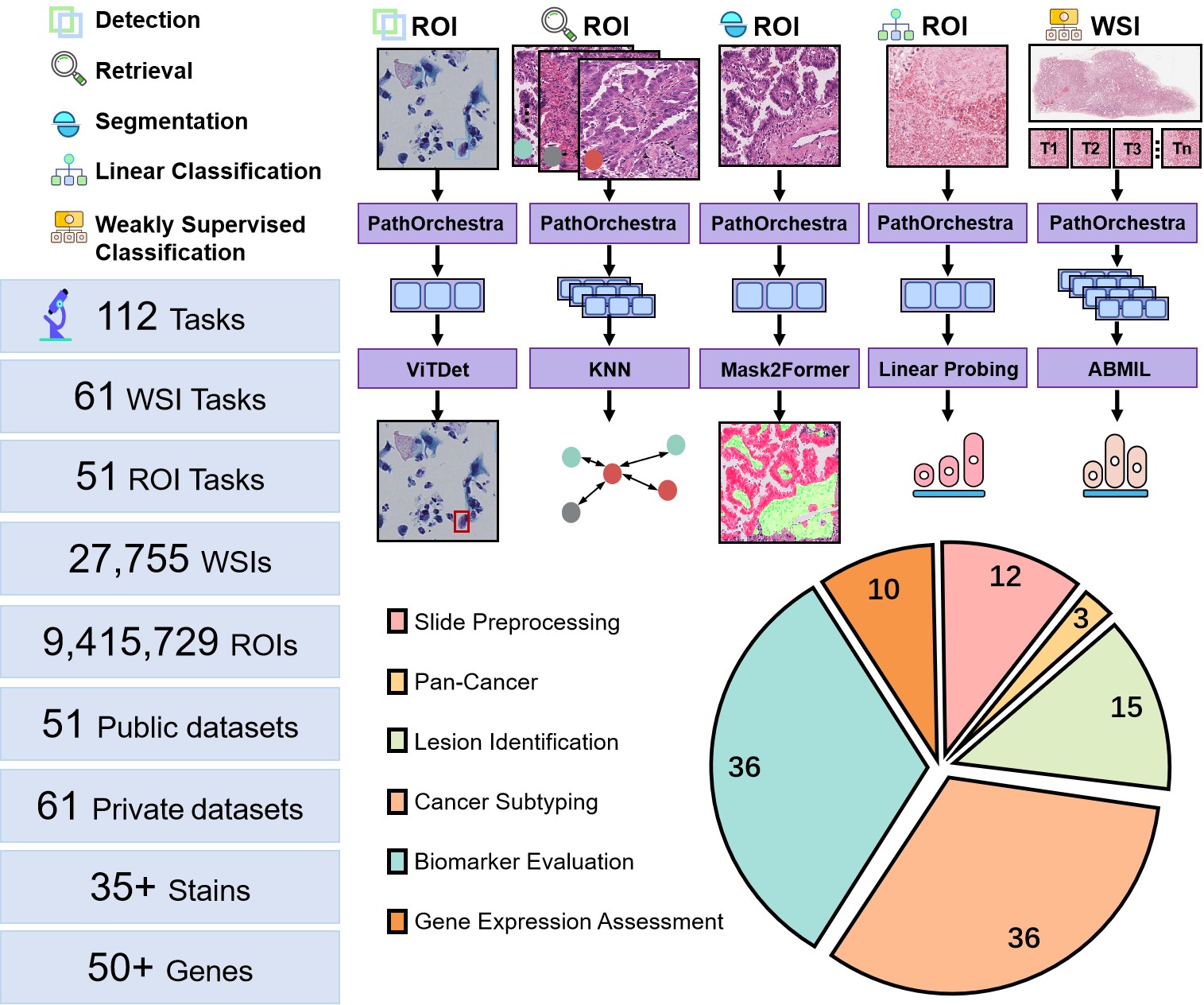}
	\caption{Overview of Downstream Tasks and Dataset Statistics. It covers a total of 112 tasks, including 61 WSI tasks and 51 ROI tasks, with a vast dataset comprising 27,755 WSIs and 9,415,729 ROIs. The datasets used span both public and private sources, with 51 public datasets and 61 private datasets, incorporating over 35 types of stains and analyzing more than 50 genes. The downstream tasks are categorized into seven groups: Slide Preprocessing (12 tasks), Pan-Cancer (3 tasks), Lesion Identification (15 tasks), Cancer Subtyping (36 tasks), Biomarker Evaluation (36 tasks), and Gene Expression Assessment (10 tasks). The methodology employed for these tasks includes detection, retrieval, segmentation, linear classification, and weakly supervised classification approaches, utilizing models like ViTDet~\cite{li2022exploring}, KNN~\cite{peterson2009k}, Mask2Former~\cite{cheng2022masked}, Linear Probing~\cite{he2022masked}, and ABMIL~\cite{ilse2018attention}.}
	\label{fig.Fig2_DownStreamOverview}
\end{figure*}

In this paper, we introduce PathOrchestra (Figure~\ref{fig.Fig1_Overview}), a pathology foundational model that establishes new benchmarks by evaluating its performance across 112 clinical downstream tasks. This model functions as a self-supervised vision encoder, pretrained on an extensive dataset of 300K hematoxylin and eosin (H\&E)-stained WSIs covering 20 different tissues and organs, sourced from both in-house collections and public datasets across three centers. We comprehensively assess PathOrchestra's applications in clinical settings (Figure~\ref{fig.Fig2_DownStreamOverview}), establishing benchmarks in real-world scenarios including digital slide preprocessing, pan-cancer classification, lesion identification and detection, multi-cancer subtype classification, biomarker evaluation, gene expression prediction, and structural reports. The model demonstrates exceptional performance among 27,755 WSIs and 9,415,729 region of interest (ROI) images, achieving over 0.950 accuracy in 47 tasks, such as pan-cancer classification across multiple organs, lymphoma subtype diagnosis, and bladder cancer screening. Furthermore, PathOrchestra is the first to generate structured reports for high-incidence colorectal cancer and diagnostically complex lymphoma, which are tasks seldom addressed by existing foundational models but hold significant clinical potential. Overall, PathOrchestra exhibits robust capabilities in addressing diagnostically challenging tasks, affirming its role as a pivotal pathology foundational model.

\noindent\textbf{\large{Methods}}

\noindent\textbf{Dataset Curation}\\
The training dataset comprises 300K WSIs, covering 20 distinct types of tissues and organ types. These WSIs were digitized from Frozen and formalin-fixed paraffin-embedded (FFPE) tissue slides stained with (H\&E), utilizing scanners from four different manufacturers (.svs, .sdpc, .kfb, .mdsx). At a magnification of 20$\times$, patch sampling was performed on these WSIs, with an average of 500 non-overlapping samples per WSI. This study was approved by the Institutional Review Boards of the Chinese Air Force Military Medical Hospital and Anhui Provincial Hospital for the retrospective analysis of internal pathological images. Prior to computational analysis and model development, all internal data including WSIs and pathology reports were de-identified to ensure patient confidentiality. Since the study involved retrospective analysis of archival pathological slides without direct patient recruitment, informed consent was not required. The training data consisted solely of pathological images without any associated patient diagnoses or personal identifiers.

\noindent\textbf{Network Architecture and Pretraining Protocol}\\
To achieve robust performance across 112 clinical-grade downstream tasks, we pre-trained our model on the curated dataset using the DINOv2 architecture~\cite{oquab2023dinov2}, which is designed to extract high-quality visual features through self-supervised learning on unlabeled data. DINOv2 employs a teacher-student framework based on Vision Transformers (ViT)~\cite{dosovitskiy2020image}, leveraging an alignment mechanism that enables the student network to learn meaningful feature representations without relying on labeled data. By eliminating the dependence on manual annotations, DINOv2 allows for the extraction of rich features from large-scale unlabeled images, significantly reducing the need for human intervention. The architecture enhances feature generalization through multi-scale, multi-view data augmentation techniques. By incorporating image crops at different scales and applying diverse augmentations, the model learns feature representations that are robust and generalize well across various downstream tasks, including classification, detection, and segmentation.

A key innovation of DINOv2 is its multi-scale feature learning strategy, which enables the model to capture both global and local information from images. This multi-scale approach improves the model's robustness when handling images at different resolutions and levels of detail, which is particularly important in histopathological analysis. Additionally, DINOv2 employs an Exponential Moving Average (EMA) mechanism~\cite{he2020momentum} for updating the teacher network's parameters, resulting in smoother and more stable feature representations. This mechanism enhances the training efficiency of the student network and helps prevent convergence to suboptimal solutions.

The architecture of DINOv2 is based on the ViT and employs a dual-network structure consisting of a teacher network and a student network. The workflow exemplifies a contrastive strategy, where the student network strategically incorporates both global and local crops, each enriched through diverse data augmentations. This dual input enables the model to capture a broader spectrum of features. The extracted features are then refined through the DINO~\cite{caron2021emerging} and iBOT~\cite{zhou2021ibot} heads, which play a critical role in enhancing the network's capacity for self-supervised learning. The goal of the student network is to align its feature representations as closely as possible with those generated by the teacher network. The teacher network’s parameters are updated using EMA from the student network’s parameters. Due to the slow update rate of EMA, the teacher network retains a smoothed version of the student network as a stable target during training. The teacher network processes only the unmasked global crops and generates target features for alignment. The DINO head processes the global features (class tokens) from both the student and teacher networks, mapping these features into a fixed-dimensional probability space to compute the cross-entropy loss. Meanwhile, the iBOT head handles partially masked image patches, predicting the features of the masked regions and improving the model’s inference ability in scenarios where information is missing.

The loss function in DINOv2 consists of three components. First, the DINO loss is used to align the outputs of the teacher and student networks on the global crops. It is computed as the cross-entropy loss between the two, as shown in Equation~\ref{dinoloss}. 

\begin{equation}
	L_{\text{DINO}} = - \sum_{i=1}^{N} p_t^i \log(p_s^i)
	\label{dinoloss}
\end{equation}

where $p_t^i$ represents the output probability distribution from the teacher network, $p_s^i$ represents the output probability distribution from the student network, and $N$ is the total number of samples. Second, the iBOT loss handles partially masked image patches, ensuring that the student network can infer the missing information by predicting the features of the masked regions, as detailed in Equation~\ref{ibotloss}. 

\begin{equation}
	L_{\text{iBOT}} = - \sum_{i \in \text{masked patches}} p_{t,i} \log(p_{s,i})
	\label{ibotloss}
\end{equation}

where $p_{t,i}$ represents the output of the teacher network at the masked patch position, and  $p_{s,i}$ represents the output of the student network at the corresponding masked patch position. Lastly, the KoLeo regularization ensures that the feature space remains evenly distributed, preventing the model from degrading into learning meaningless feature representations, as outlined in Equation~\ref{koleoloss}.

\begin{equation}
	L_{\text{KoLeo}} = - \frac{1}{n} \sum_{i=1}^{n} \log(d_{n,i})
	\label{koleoloss}
\end{equation}

where $d_{n,i}$ represents the feature distance between the $i$-th sample and its nearest neighbor in the same batch, and $n$ is the total number of samples in the batch.

The training process of DINOv2 primarily consists of the following three steps. The first step is data preprocessing and augmentation, which involves applying various augmentations such as random cropping, horizontal flipping, color jittering, grayscale conversion, and Gaussian blur to the original images. These operations generate multiple global and local crops, which are simultaneously fed into the student network for feature extraction. The second step is feature extraction and alignment. The student network extracts features from both global and local crops, while the teacher network extracts features solely from the global crops. The DINO and iBOT heads are then used to compute the alignment loss between the features extracted by the student and teacher networks. The third step involves computing the loss and updating the network. The total loss is calculated using the DINO loss, iBOT loss, and KoLeo regularization~\cite{sablayrolles2018spreading}, after which the student network’s parameters are updated through backpropagation. Simultaneously, the teacher network’s parameters are updated using the EMA mechanism, ensuring that the teacher network remains smoother and more stable. In the experiments conducted in this paper, we chose to use the teacher network as the encoder for downstream tasks. This decision is grounded in the observation that the teacher network, which is updated through EMA, exhibits smoother and more stable parameters. Consequently, this leads to enhanced generalization capabilities. Therefore, the encoder of the teacher network is more adept at handling downstream tasks that demand high robustness and applicability across different domains.

\noindent\textbf{Clinical Task Assessment Methodology}\\
We evaluated the pre-trained model on a diverse set of 112 clinically relevant downstream tasks, including pathology quality control, pan-cancer classification, lesion detection, multi-cancer subtyping, biomarker assessment, gene expression prediction, and structured report generation. 

\noindent\textbf{I. Pathology Image Preprocessing and Quality Control Tasks.}\ To assess the model's ability to handle common artifacts and variations in pathological images, we conducted several quality control tasks focusing on image quality assessment and recognition of staining methods, magnifications, and specimen types. For image quality assessment, we developed classifiers to detect various issues that can impede accurate pathological analysis. These issues included differentiating between natural images (sourced from the ImageNet-1K validation set) and pathological images (random patches from gastrointestinal H\&E and immunohistochemistry data), identifying tissue folds, bubbles, glue artifacts, contaminants such as dust, hair, and fibers, and detecting blurriness using the FocusPath-UofT dataset~\cite{hosseini2019encoding}. In terms of staining and specimen recognition, we evaluated the model's capability to recognize different staining methods, magnifications, and specimen types. We classified images based on staining type, distinguishing between H\&E and IHC stained images, as well as between H\&E stained images and immunofluorescence images from the DeepCell dataset~\cite{greenwald2022whole}. We also classified four types of IHC markers (PR, ER, Ki67, and HER2) to assist in molecular characterization, differentiating IHC markers based on nuclear (PR, ER, Ki67) and membrane (HER2) positivity. Furthermore, we developed models to distinguish images at different magnifications (10$\times$ and 20$\times$) to improve model robustness and to identify whether H\&E images were from frozen sections or FFPE samples. Specimen type recognition included classifying images based on specimen size to enhance preprocessing workflows. For these tasks, we utilized datasets comprising thousands to hundreds of thousands of images, employing linear probing strategies.

\noindent\textbf{II. Pan-cancer Classification Tasks.}\ To evaluate the model's generalization across various cancer types, we performed multi-class classification tasks using large-scale datasets from TCGA. Specifically, we conducted pan-cancer classification involving 17 subtypes by classifying 1,071 WSIs from various organs and tissues, including lung, kidney, breast, and prostate cancers. We also performed FFPE pan-cancer classification involving 32 subtypes using 3,036 FFPE WSIs and frozen pan-cancer classification with the same 32 subtypes using 3,038 frozen WSIs for comparative analysis. The multiple instance learning (ABMIL) framework~\cite{ilse2018attention} was employed with consistent training parameters to ensure robust performance.

\noindent\textbf{III. Multi-organ Lesion Identification and Analysis Tasks.}\ We assessed the model's capability in lesion detection and segmentation across multiple organs and modalities. For lesion detection, we trained models to detect breast cancer metastasis in lymph node WSIs using datasets from the CAMELYON16 and CAMELYON17 challenges. We developed models to segment regions for PD-L1 assessment in lymphoma, enhancing immunotherapy decision-making, and classified images based on the presence of tumor-infiltrating lymphocytes using the TCGA-TILs dataset. Additionally, we performed binary classification of metastatic cancer in breast tumor regions using the PCam dataset, which contains over 2 million image patches, and identified signet ring cell carcinoma in gastrointestinal sections using the DigestPath dataset. Training strategies included linear probing and fine-tuning, with images processed at appropriate magnifications. For segmentation tasks, we employed the Mask2Former model~\cite{cheng2022masked}, fine-tuned with a Vision Transformer Adapter (ViT-Adapter)~\cite{chen2022vitadapter}, to segment eight major cell types within tumor tissues using the SegPath dataset, segment glands in colorectal adenocarcinoma tissue sections using the GlaS dataset, and segment nuclei across 19 tissue types using the PanNuke dataset. We also utilized the ViTDet model~\cite{li2022exploring} for few-shot detection tasks, detecting and classifying positive cells in cervical cell pathology ROIs with over 1.3 million images and identifying abnormal cells in pleural and peritoneal fluid pathology slides using 36,544 ROI images.

\noindent\textbf{IV. Multi-cancer Subtype Classification Tasks.}\ We evaluated the model's ability to classify various cancer subtypes across multiple organs, integrating both binary and multi-class classification tasks. For binary classification, we conducted cancer versus benign tissue classification tasks in bladder, cervical, and digestive tract cancers, as well as distinguishing between lymphoma and reactive hyperplasia and between B-cell and T-cell lymphomas. We utilized the ABMIL framework and linear probing strategies to optimize performance. For multi-class subtyping, we performed organ-specific cancer subtyping tasks, including detecting cervical precancerous lesions using the TissueNet dataset, classifying tissue images into multiple categories using subsets of the LC25K dataset, and classifying benign and malignant breast tumors at high magnification using the BreaKHis dataset. Additionally, we subtyped non-small cell lung cancer and renal cell carcinoma using TCGA datasets, classified lymphoma into multiple subtypes for precise treatment, graded prostate cancer using core needle biopsies and Gleason scores, and classified gastric adenocarcinoma subtypes using the PatchGastricADC22 dataset. Training involved linear probing and ABMIL frameworks to enhance model accuracy.

\noindent\textbf{V. Biomarker Assessment Tasks.}\ We assessed the model's ability to evaluate the expression of biomarkers critical for diagnosis and treatment planning, particularly in the context of immunohistochemistry. We performed multi-IHC marker qualification by assessing the expression of 30 different IHC markers in lymphomas using 3,976 WSIs. Individual IHC marker analysis was conducted for specific markers such as CD5, CD3, CD20, CD79a, CD21, EBER, CD10, Bcl-6, Bcl-2, MUM-1, CD4, CD23, PD-1, CyclinD1, CD19, CD22, CD8, C-myc, CD56, Granzyme B, TIA-1, Perforin, CD2, CD30, CD7, CD38, ICOS, CXCL-13, ALK, PD-L1, HER2, ER, PR, Ki67, and others, with each task involving datasets ranging from 11 to 313 WSIs. We also evaluated breast cancer biomarkers by assessing HER2 scoring and ER, PR, and Ki67 status. For all tasks, we consistently employed a weakly supervised classification method using the ABMIL framework with standardized parameters.

\noindent\textbf{VI. Gene Expression Prediction Tasks.}\ To evaluate the model's ability to predict gene expression levels from pathological images using the HEST-Benchmark dataset~\cite{jaume2024hest}. We predicted expression levels of the top 50 genes across nine cancer types, including Invasive Ductal Carcinoma (IDC), Prostate Adenocarcinoma (PRAD), Skin Cutaneous Melanoma (SKCM), Colonic Adenocarcinoma (COAD), Rectal Adenocarcinoma (READ), Clear Cell Renal Cell Carcinoma (ccRCC), Hepatocellular Carcinoma (HCC), Lung Adenocarcinoma (LUAD), and Axillary Lymph Nodes in IDC (LYMPH\_IDC). By integrating and summarizing these tasks according to their major categories, we provide a cohesive overview of our evaluation of the pre-trained model across a diverse range of clinically relevant applications. This approach emphasizes the model's versatility and potential impact in various domains of pathology. Detailed descriptions of each task, including datasets, experimental setups, and training parameters, are available in the appendix for further reference.

\noindent\textbf{VII. Structured Report Generation Tasks.} \ The generation of structured reports is of significant importance in computational pathology, particularly in the diagnosis of lymphoma and colorectal cancer. This study utilizes the PathOrchestra model to generate structured reports for each patient through a combination of task subsets. In the generation of lymphoma reports, we first analyze conventional H\&E stained images for the diagnosis of lymphoma subtypes. However, this process often fails to provide definitive evidence for specific subtypes, necessitating the integration of IHC for quantitative analysis. Different lymphoma subtypes correspond to different IHC panels. This study particularly focuses on the diagnosis of Angioimmunoblastic T-cell Lymphoma (AITL) and Diffuse Large B-Cell Lymphoma (DLBCL) subtypes, combining the HE subtype prediction task with qualitative assessments of 29 IHC markers. For each patient, we employ weakly supervised learning to diagnose the H\&E stained pathology slides and conduct qualitative analyses for each IHC type, ultimately generating structured reports. In our research on colorectal cancer, we concentrate on distinguishing between tumor and non-tumor tissues through HE images and classifying cancer grades (cancerous, low-grade cancer, high-grade cancer). Additionally, for negative samples, we further categorize polyp types, including hyperplastic polyps, inflammatory polyps, polypoid hyperplasia, and no polyps. For each patient, we also utilize weakly supervised learning to train classification models for different tasks, leading to the generation of structured reports.

\noindent\textbf{Evaluation Setting}\ We utilized a range of metrics to comprehensively evaluate the performance of classification, detection, and segmentation tasks. For classification tasks, we reported balanced accuracy (ACC), weighted F1 score, and the Area Under the Receiver Operating Characteristic curve (AUC). Balanced accuracy calculates the average recall obtained on each class, effectively addressing class imbalance. The weighted F1 score combines precision and recall for each class, weighted by the number of true instances in each class. AUC measures the ability of the classifier to distinguish between classes across all threshold settings. For cell detection tasks, we mainly used Mean Average Precision (mAP), which evaluates the detection performance by considering both the precision and recall of detected objects over different confidence thresholds. For segmentation tasks, we mainly reported metrics such as Mean Pixel Accuracy (MPA), class pixel accuracy (CPA), Intersection over Union (IoU), and Dice. MPA represents the average pixel-wise accuracy across classes, CPA measures the accuracy for each individual class at the pixel level, Mlou measures the overlap between the predicted segmentation and ground truth, and Mean Dice quantifies the overall segmentation accuracy. Other details can be found in the Results section.

\noindent\textbf{\large{Results}}

\noindent\textbf{Pathology Image Preprocessing and Quality Control Tasks} \\
We classified the preprocessing tasks in CPath into two main categories which are pathology quality control tasks and general analysis tasks. The quality control tasks include wrinkle detection, bubble and adhesive identification, contamination detection, and blur detection. The general analysis tasks encompass pathology image identification, recognition of H\&E and IHC staining, identification of IHC marker types, magnification recognition, differentiation between frozen and FFPE sections, identification of biopsy versus large specimens, and recognition of IHC membrane and nuclear positivity. Preprocessing is a critical step in the CPath workflow, directly impacting the performance of subsequent algorithms. Developing a feature extractor with strong generalization capabilities can significantly enhance model performance while reducing reliance on extensive training data and annotations. All preprocessing tasks were conducted using supervised ROI classification based on the linear probing method, with model performance evaluated using area under the curve (AUC) and F1 scores. Detailed information on the preprocessing tasks and experimental setups is provided in the Methods section.

\begin{figure*}[!htbp]
	\centering
	\includegraphics[width=1\linewidth]{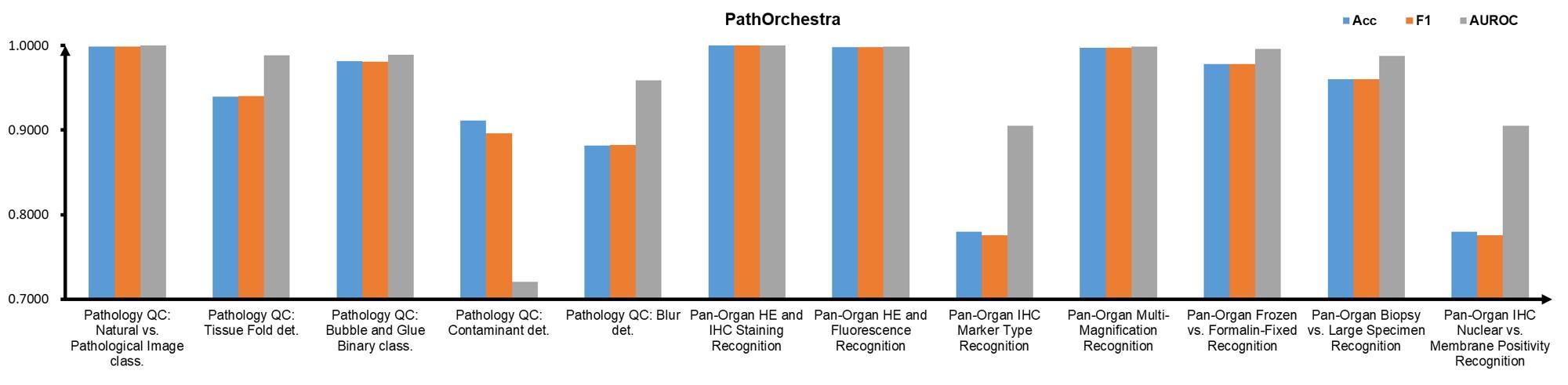}
	\caption{Performance on Pathology Image Preprocessing and Quality Control Tasks. The model's performance in quality control and general analysis tasks included 12 ROI-level classification sub-tasks, covering areas such as pathology image quality control, magnification recognition, staining type identification, specimen collection method identification, and slide preparation form identification.}

	\label{fig.Fig2_preprocess}
\end{figure*}

Among the 12 preprocessing-related tasks (Figure~\ref{fig.Fig2_preprocess}), the PathOrchestra model achieved AUC and F1 scores exceeding 0.950 in 7 subtasks. The general analysis tasks demonstrated superior performance compared to the quality control tasks, with AUC and F1 scores surpassing 0.970 for pathology image identification, recognition of H\&E and IHC staining, recognition of H\&E and fluorescence staining, magnification discrimination, and differentiation between frozen and FFPE sections. In quality control tasks such as bubble and adhesive identification, the model achieved AUC and F1 scores above 0.980. These results demonstrate the model's robust capability to effectively execute a wide range of preprocessing tasks without extensive retraining.

\noindent\textbf{Pan-cancer Classification Tasks} \\
We evaluated PathOrchestra on slide-level pan-cancer classification tasks, including a 17-class pan-cancer tissue classification using an in-house FFPE dataset, a 32-class classification using data from the TCGA FFPE dataset, and a 32-class classification using data from the TCGA frozen tissue dataset. These tasks assessed the model's performance across different cancer types, data sources, and specimen preparation methods. Pan-cancer classification models are designed to recognize and distinguish multiple cancer types, necessitating strong generalization capabilities crucial for the early detection of various cancers. All tasks were performed using weakly supervised WSI classification based on the attention-based ABMIL, with model performance evaluated using AUC and F1 scores. Detailed information on these tasks and experimental setups is provided in the Methods section.

\begin{figure*}[!htbp]
	\centering
	\includegraphics[width=1\linewidth]{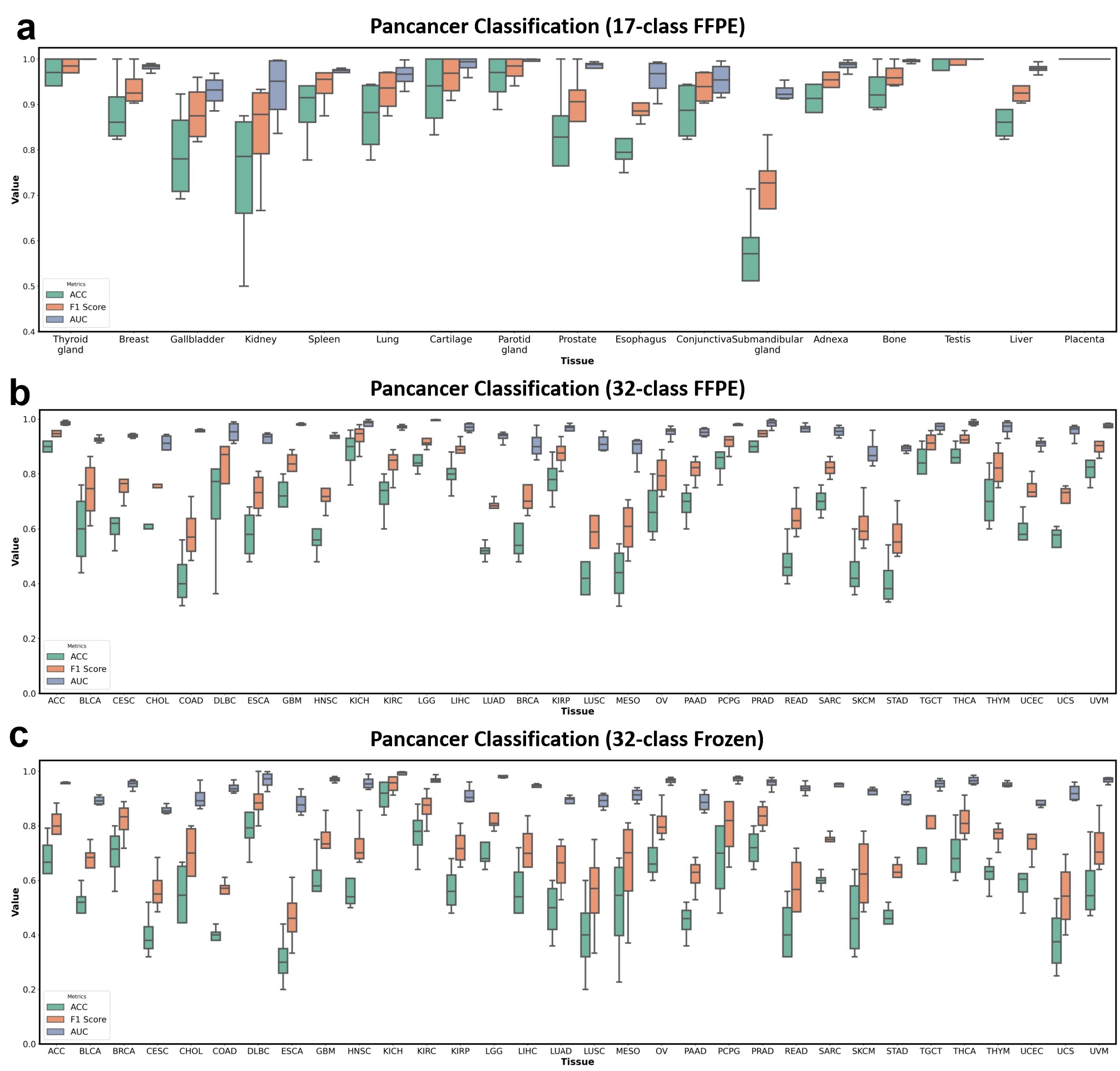}
	\caption{Results of Pan-cancer Classification Tasks. a. The performance of the model in weakly supervised WSI-level classification tasks for 17 cancer types using in-house center data is illustrated. The bar chart demonstrates the impact of the model on these predictions. b and c. Data from 32 high-incidence cancer types were selected from the TCGA database to evaluate the model's performance in pan-cancer prediction across two slide formats including FFPE and Frozen. The bar charts depict the model's impact on prediction accuracy for each format.}
	\label{fig.Fig3_Pancancer}
\end{figure*}

In the 17-class pan-cancer tissue classification task (Figure~\ref{fig.Fig3_Pancancer}a), the model achieved an average value for AUC of 0.988, ACC of 0.879, and F1 score of 0.863. For the 32-class classification using the TCGA FFPE dataset, the model attained the AUC of 0.964, ACC of 0.666, and F1 score of 0.667 (Figure~\ref{fig.Fig3_Pancancer}b). We observed that the classification performance for frozen sections was 1.4\% lower in AUC, 8.9\% lower in ACC, and 9\% lower in F1 compared to FFPE sections under similar conditions (Figure~\ref{fig.Fig3_Pancancer}c). This discrepancy may be attributed to the better preservation of tissue structure and morphology in FFPE sections, which facilitates feature extraction and classification. These results underscore the model's robust generalization capabilities in pan-cancer classification tasks, highlighting its potential as a foundational pathology model for screening multiple cancer types and improving diagnostic accuracy.

\noindent\textbf{Multi-organ Lesion Identification and Analysis Tasks} \\
Lesion identification and detection encompass 15 tasks, including WSI-level breast cancer metastasis detection (CAMELYON16, CAMELYON17), ROI-level lymphoma tumor region segmentation (PD-L1), multi-organ tumor-infiltrating lymphocyte (TIL) positive/negative screening (TCGA-TIL), breast tumor classification (PCam), colorectal malignant lesion classification (DigestPath), multi-organ cell semantic segmentation (SegPath), colorectal gland segmentation (GlaS), multi-organ nuclear instance segmentation (PanNuke), colorectal cell instance segmentation (CoNSeP), lung cancer tissue segmentation (WSSS4LUAD), cross-organ and cross-scanner adenocarcinoma segmentation (COSAS-Seg) and tissue classification (COSAS-Clas), multi-organ mitosis detection (SegPath), cervical cancer TCT inflammatory cell detection, and multi-organ PE\&A inflammatory cell detection. For WSI-level classification tasks, we utilized ABMIL for weakly supervised learning, while ROI classification was evaluated using linear probing. Cell segmentation tasks were conducted using U-Net~\cite{ronneberger2015u} architectures. Classification performance was assessed using AUC and F1 scores, while segmentation performance was evaluated using mAP, IoU, and Dice coefficients. Detailed descriptions of these tasks and experimental setups are provided in the Methods section.

\begin{figure*}[!htbp]
	\centering
	\includegraphics[width=1\linewidth]{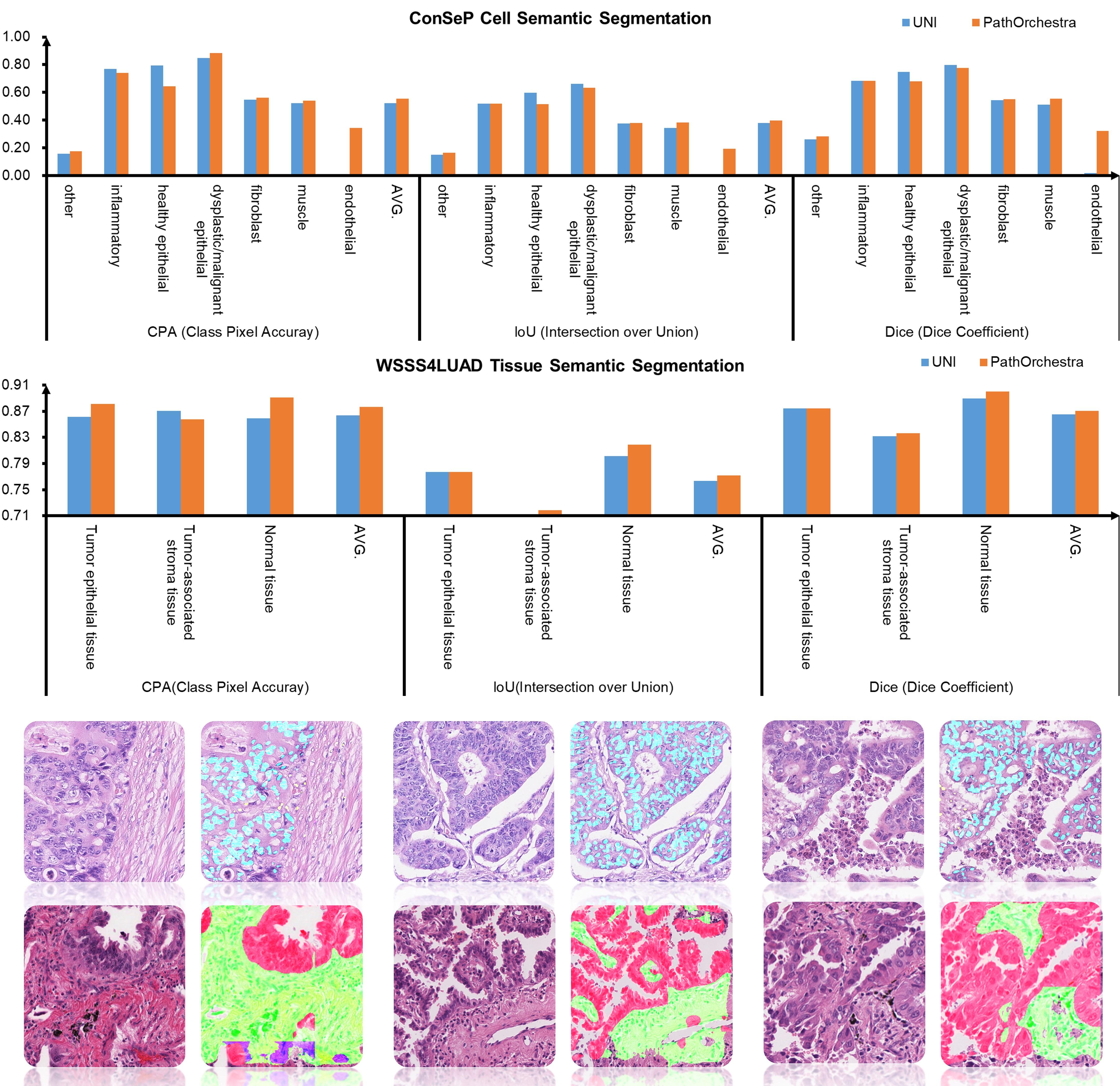}
	\caption{Cell Segmentation Results. This figure presents a comparative analysis of the semantic segmentation performance between PathOrchestra and the UNI model on two distinct datasets. The upper bar chart corresponds to the results from the ConSeP dataset for cell semantic segmentation, showcasing performance across three metrics: Class Pixel Accuracy (CPA), Intersection over Union (IoU), and Dice Coefficient (Dice). The lower bar chart illustrates the outcomes from the WSSS4LUAD tissue semantic segmentation task, assessing the performance of PathOrchestra and UNI models across three main tissue categories: tumor epithelial tissue, tumor-associated stromal tissue, and normal tissue. The images at the bottom provide a visual comparison of the segmented tissue samples, further substantiating the superiority of the PathOrchestra model in accurately segmenting and identifying complex tissue structures.}
	\label{fig.Fig8_CellSegmentation}
\end{figure*}

We evaluated the performance of PathOrchestra across multiple tasks under the multi-organ lesion identification and analysis category, with segmentation results shown in Figure~\ref{fig.Fig8_CellSegmentation}. For the CoNSeP task, which is a multi-organ nuclear instance segmentation task, we used three evaluation metrics: MPA, IoU, and Mean Dice. PathOrchestra performed poorly in this task, with an MPA of 0.556, Mlou of 0.399, and Mean Dice of 0.550, indicating substantial room for improvement in nuclear instance segmentation. Similarly, for the WSSS4LUAD task, which is a semantic segmentation task, the MPA was 0.877, Mlou was 0.771, and Mean Dice was 0.871. For the SegPath task, we segmented lesion regions in images using evaluation metrics including Precision, Recall, and Mean Dice. The experimental results showed that PathOrchestra achieved a Precision of 0.756, Recall of 0.720, and Dice of 0.717 in the SegPath task, demonstrating balanced segmentation performance and the ability to effectively identify and segment lesion regions. For the Gland segmentation task (based on the GlaS dataset), we evaluated the performance of PathOrchestra in glandular region segmentation, using Precision, Recall, and Dice as evaluation metrics. The results showed that the model achieved a Precision of 0.929, Recall of 0.868, and Dice of 0.931, indicating good performance in glandular region segmentation. For the PanNuke task, PathOrchestra achieved a Precision of 0.923, Recall of 0.833, and Dice of 0.849, demonstrating excellent performance in nuclear segmentation.

In the mitosis detection task, we also evaluated the performance of PathOrchestra under the SegPath task, using mAP (bbox, IoU=0.5), mAP (seg, IoU=0.5), and mAP (seg, IoU=0.75) as evaluation metrics. The experimental results showed that the model's mAP (bbox, IoU=0.5) and mAP (seg, IoU=0.5) were 0.885 and 0.883, respectively, while mAP (seg, IoU=0.75) was 0.658. These results indicate that PathOrchestra has limitations in accurately locating and segmenting mitotic cells, especially at higher segmentation precision requirements. For two cell detection tasks, namely TCT inflammatory cell detection and PE\&A inflammatory cell detection, the results showed that PathOrchestra achieved an mAP (bbox, IoU=0.5) of 0.351 in the TCT inflammatory cell detection task, while the mAP in the PE\&A inflammatory cell detection task was only 0.634, indicating considerable performance variation across different cell type detection tasks.

For other classification tasks, we used ACC, F1, and AUC as evaluation metrics. PathOrchestra performed excellently in the breast cancer metastasis detection task (WSI level), achieving 1.00 for ACC, F1, and AUC on the CAMELYON16 dataset, while its performance on CAMELYON17 was subpar, with an ACC of 0.840, F1 of 0.588, and AUC of 0.784. However, on the Pcam dataset, which is also a breast cancer classification model, PathOrchestra performed well, achieving an ACC of 0.916, F1 of 0.916, and AUC of 0.975. In other classification tasks, such as lymphoma PD-L1 tumor region detection, TIL positivity detection, and DigestPath tumor cell classification, PathOrchestra's ACC values were 0.851, 0.931, and 0.916, respectively, with AUCs of 0.927, 0.959, and 0.975, demonstrating good classification performance. However, in some tasks, such as COSAS screening and COSAS tissue classification, the model's AUC values were relatively low, at 0.897 and 0.624, respectively, indicating its shortcomings in complex tissue type classification tasks. Overall, PathOrchestra demonstrated some potential in multi-organ lesion identification and analysis tasks, but improvements are still needed in certain tasks to enhance the model's accuracy and robustness, particularly in more complex segmentation and cell detection tasks.

\noindent\textbf{Multi-cancer Subtype Classification Tasks} \\
Tumor classification and subtyping are critical in enhancing diagnostic accuracy and personalizing treatment plans, thereby optimizing patient prognosis and clinical outcomes. We evaluated the performance of PathOrchestra across tumor classification and subtyping tasks, including cervical cancer subtyping, bladder cancer screening (WSI and ROI), cervical TCT screening (WSI and ROI), gastrointestinal benign and malignant screening, lymphoma and reactive hyperplasia screening, lymphoma cell origin classification, colorectal cancer detection, cervical epithelial lesion detection, colorectal and lung tissue screening (LC25K), lung cancer subtyping (LC25K-lung), colorectal subtyping (LC25K-colon), breast cancer screening (BreastHis), lung cancer subtyping (TCGA-NSCLC), colorectal subtyping (TCGA-RCC), lymphoma subtyping, breast cancer subtyping (BACH), esophageal cancer screening (TCGA-ESCA), colorectal tissue classification (HunCRC-patch), prostate ISUP grading (PANDA), prostate Gleason grading, colorectal high and low-grade screening, gastric cancer subtyping (PatchGastricAD22), prostate cancer tissue classification (AGGC), glioma IDH1 screening (TCGA-IDH1), multi-organ lymph node metastasis screening, colorectal polyp screening, colorectal subtyping (HunCRC-WSI), renal cancer tissue classification (TCGA-RCC), colorectal tissue classification (CRC-100K), colorectal subtyping (institutional data), colorectal tissue classification (Kather), and brain subtyping (Ebrains). Detailed information on these tasks and experimental setups is provided in the Methods section.

\begin{figure*}[!htbp]
	\centering
	\includegraphics[width=1.0\linewidth]{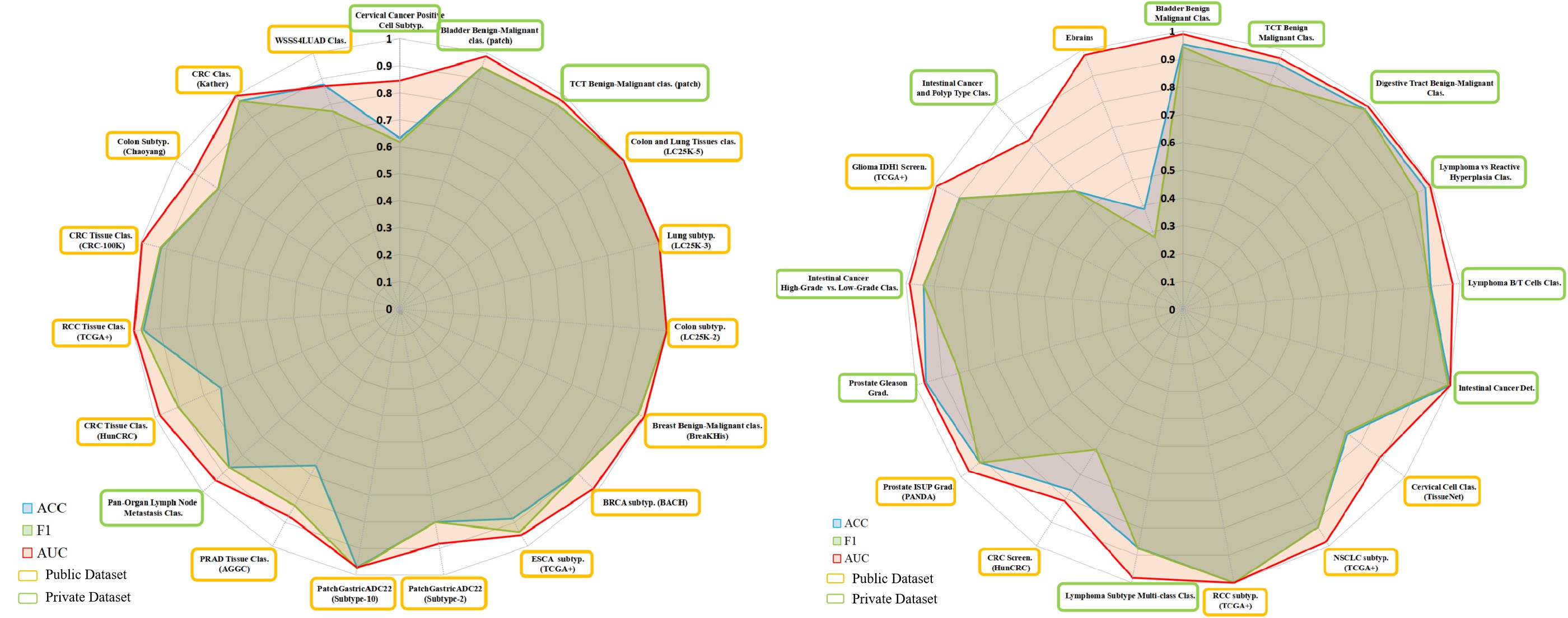}
	\caption{Results of Multi-cancer Subtype Classification. The radar charts illustrate PathOrchestra's performance across ROI-level (left) and WSI-level (right) multi-cancer subtype classification tasks. Key metrics of ACC, F1 score, and AUC are shown for each task. Public and private datasets are compared, highlighting the model's robustness across various classification challenges.}
	\label{fig.Fig5_subtypeClass}
\end{figure*}

In the evaluation of 34 tumor classification and subtype identification tasks (Figure~\ref{fig.Fig5_subtypeClass}), PathOrchestra demonstrated broad adaptability and high performance levels. Notably, in the classification tasks for cervical and bladder cancers, PathOrchestra excelled. For instance, the accuracy for benign and malignant bladder cancer classification reached 0.954, with an F1 score of 0.943 and an AUC value of 0.989, showcasing the model's exceptional performance in this task. Furthermore, in the patch classification task for bladder cancer, the model's F1 score improved to 0.954, while the accuracy and AUC values also reached 0.946 and 0.988, respectively, indicating that the model maintains efficiency even in more granular image processing scenarios. In the screening for benign and malignant gastrointestinal tumors, PathOrchestra exhibited remarkable precision, achieving an accuracy of 0.972, an F1 score of 0.971, and an AUC as high as 0.988. These results highlight the model's strong diagnostic potential in the detection and classification of gastrointestinal cancers. In lymphoma-related classification tasks, PathOrchestra also demonstrated robust performance. For example, in the task of classifying lymphoma versus reactive hyperplasia, the model achieved an accuracy of 0.975, an F1 score of 0.941, and an AUC of 0.994. In the more refined classification of B/T-cell lymphomas, PathOrchestra recorded an F1 score of 0.931 and an AUC of 0.993, reflecting the model's high accuracy in subtype classification. Similarly, in the subtype classification tasks for colorectal and lung cancers, PathOrchestra performed admirably. In the colorectal subtype classification (LC25K-2), all evaluation metrics (Accuracy, F1, and AUC) achieved 1.000, demonstrating absolute classification accuracy. Likewise, for the lung cancer subtype classification (LC25K-3), the model attained a high accuracy and F1 score of 0.999, with an AUC of 0.999, indicating strong generalization capabilities across different tissue types.

However, despite PathOrchestra's outstanding performance in most tasks, there are areas for improvement. For instance, in the IDH1 screening task for gliomas, although the AUC reached 0.940, the accuracy and F1 scores were relatively low at 0.896 and 0.894, suggesting that the model's precision in screening this specific subtype may not match its performance in other cancers. In more complex tasks, PathOrchestra's performance was comparatively weaker. For example, in the prostate cancer grading (AGGC) task, the model achieved an accuracy of only 0.660, an F1 score of 0.828, and an AUC of 0.873, reflecting limitations in handling highly complex pathological tasks. Additionally, in glioma screening, although the AUC reached 0.871, the accuracy and F1 scores were notably lower at 0.386 and 0.278, indicating a need for further optimization in brain tumor screening. In summary, PathOrchestra demonstrated exceptional accuracy and robustness across various cancer classification and subtype identification tasks, particularly excelling in bladder cancer, gastrointestinal cancers, and colorectal cancer tasks. Nevertheless, there remains room for improvement in certain complex tasks, such as prostate cancer grading and glioma screening. This indicates that while PathOrchestra performs excellently in most tasks, further optimization is necessary for specific tasks to enhance its overall diagnostic capability.

\noindent\textbf{Biomarker Assessment Tasks} \\
Biomarker evaluation tasks significantly enhance clinical and research applications by improving diagnostic accuracy, personalizing treatment plans, monitoring disease progression, and driving research innovation. We assessed the performance of PathOrchestra across 36 biomarker evaluation tasks. These tasks include qualitative analysis of multiple IHC markers for lymphoma, such as CD5, CD3, CD20, CD79a, CD21, EBER, CD10, Bcl-6, Bcl-2, MUM-1, CD4, CD23, PD-1, CyclinD1, CD19, CD22, CD8, C-myc, CD56, Granzyme B, TIA-1, Perforin, CD2, CD30, CD7, CD38, ICOS, CXCL-13, ALK, and PD-L1. For breast cancer, we evaluated HER2 scoring, as well as qualitative assessments of HER2, ER, PR, and Ki67. Details of these tasks and experimental setups are provided in the Methods section.

The model demonstrated varying levels of performance across the 36 biomarker evaluation tasks. High accuracy was achieved in tasks such as HER2 scoring and CD20 qualitative analysis, with accuracy rates exceeding 0.920. In tasks involving complex biomarker profiles, such as multi-IHC marker analysis for lymphoma, the model performed moderately well, with accuracies ranging from 0.800 to 0.880. Qualitative evaluations of markers like PD-L1 and CyclinD1 showed promising results, indicating the model's potential in handling nuanced histopathological features. However, some tasks, particularly those involving markers with subtle expression differences such as CD56 and CXCL-13, presented challenges, with accuracies around 0.700, highlighting the need for further model refinement and more comprehensive training datasets. The performance in breast cancer biomarkers, including ER and PR qualitative analysis, was robust, achieving accuracies of approximately 0.900, suggesting strong model capabilities in these specific contexts. Overall, the results underscore the potential of PathOrchestra in biomarker evaluation, demonstrating both strengths and areas for improvement.

\begin{figure*}[!htbp]
	\centering
	\includegraphics[width=0.81\linewidth]{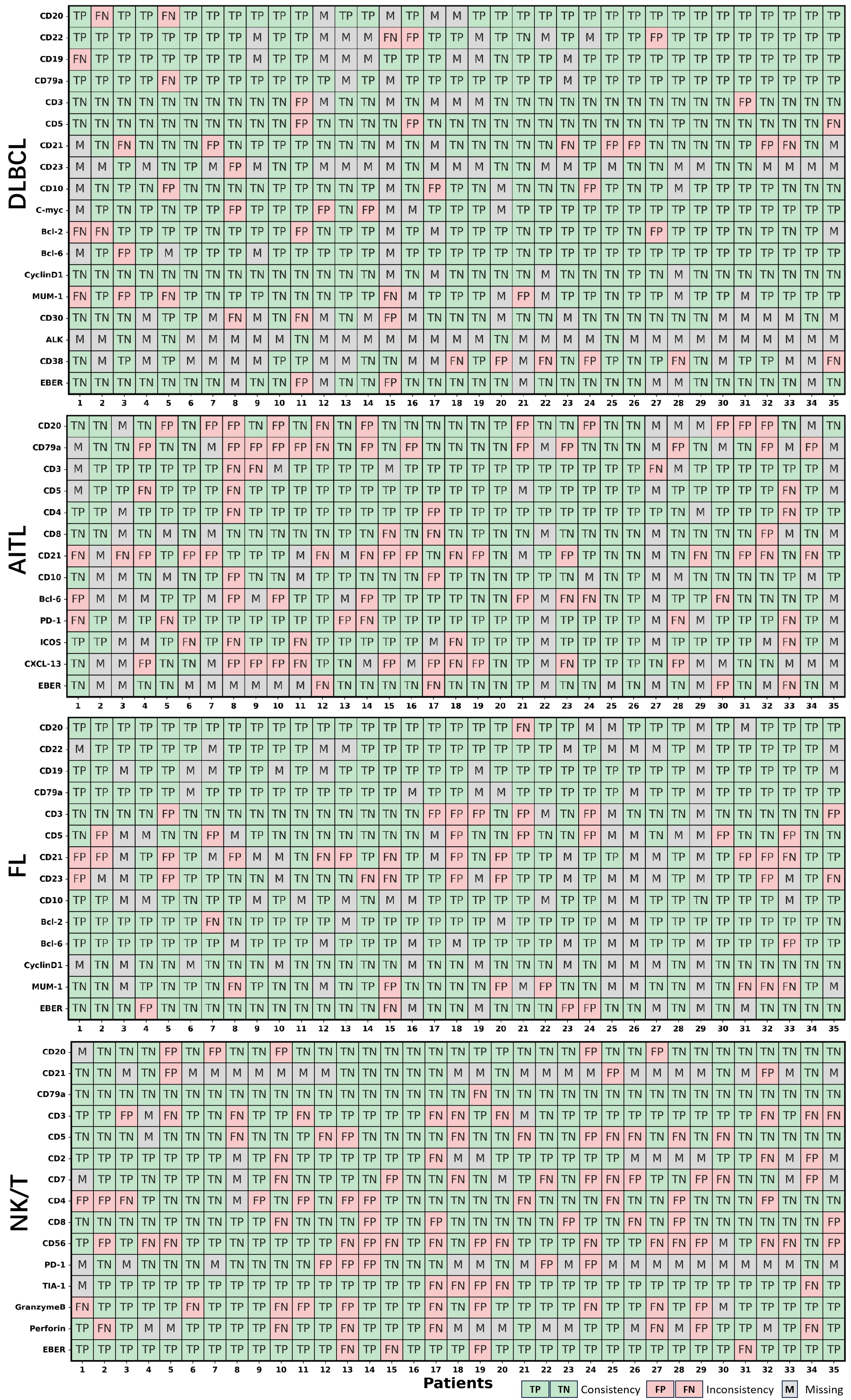}
	\caption{Performance on Biomarker Evaluation Tasks. (Caption continued on next page.)}
	\label{fig.fig-DLBCL_simple_compare}
\end{figure*}

\begin{figure*}[!htbp]
	\ContinuedFloat
	\caption{Qualitative IHC results for lymphoma patients as evaluated by pathologists. The IHC marker results generated by the AI algorithm based on PathOrchestra, are compared with those from pathologists. Green highlights indicate substantial agreement across markers. Thirty-five patients for each lymphoma disease were selected for visualization.}
	\label{fig.fig-DLBCL_simple_compare}
\end{figure*}

\noindent\textbf{Gene Expression Prediction Tasks} \\

Gene expression evaluation in computational pathology significantly enhances both clinical and research applications by enabling personalized treatments, monitoring disease progression, and driving innovation. We evaluated the performance of PathOrchestra across 10 gene expression prediction tasks, including multi-disease gene expression and specific assessments for invasive ductal carcinoma (IDC) in breast cancer, prostate cancer (PRAD), skin cancer (SKCM), colon cancer (COAD), rectal cancer (READ), clear cell renal cell carcinoma (CCRCC), liver cancer (HCC), lung adenocarcinoma (LUAD), and lymph node IDC. Detailed information regarding the gene datasets and experimental setups is provided in the Methods section.

\begin{figure*}[!htbp]
	\centering
	\includegraphics[width=0.95\linewidth]{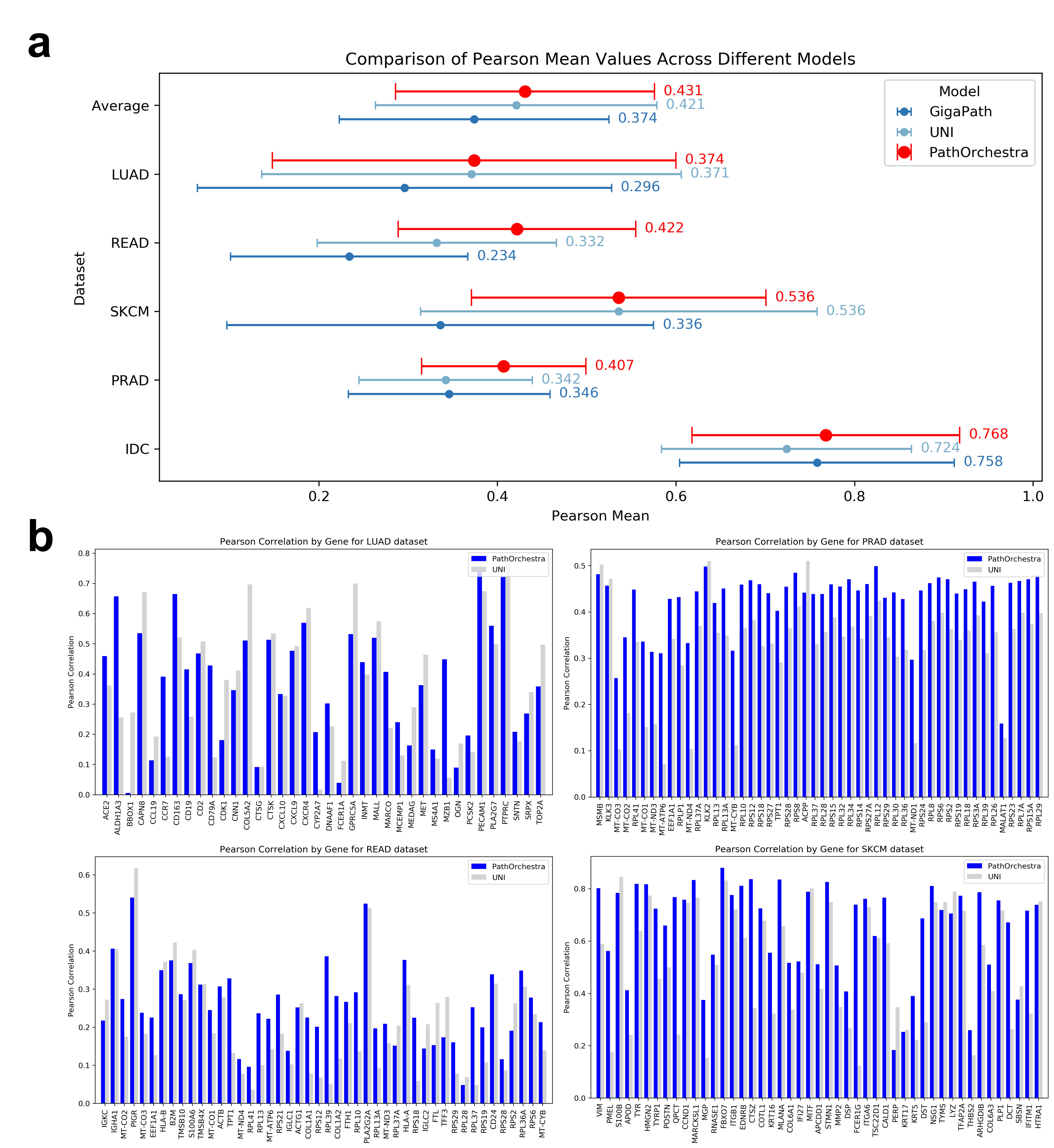}
	\caption{Comparison of Gene Expression Evaluation Results. a. The performance advantage of the model over GigaPath and UNI in gene expression prediction was evaluated across five cancer types: LUAD, READ, SKCM, PRAD, and IDC. b. A detailed comparison of the model and UNI was conducted on LUAD, PRAD, READ, and SKCM, focusing on their predictive capabilities across multiple gene categories within each cancer type.}
	\label{fig.Fig6_Gene}
\end{figure*}

\begin{figure*}[!htbp]
	\centering
	\includegraphics[width=0.83\linewidth]{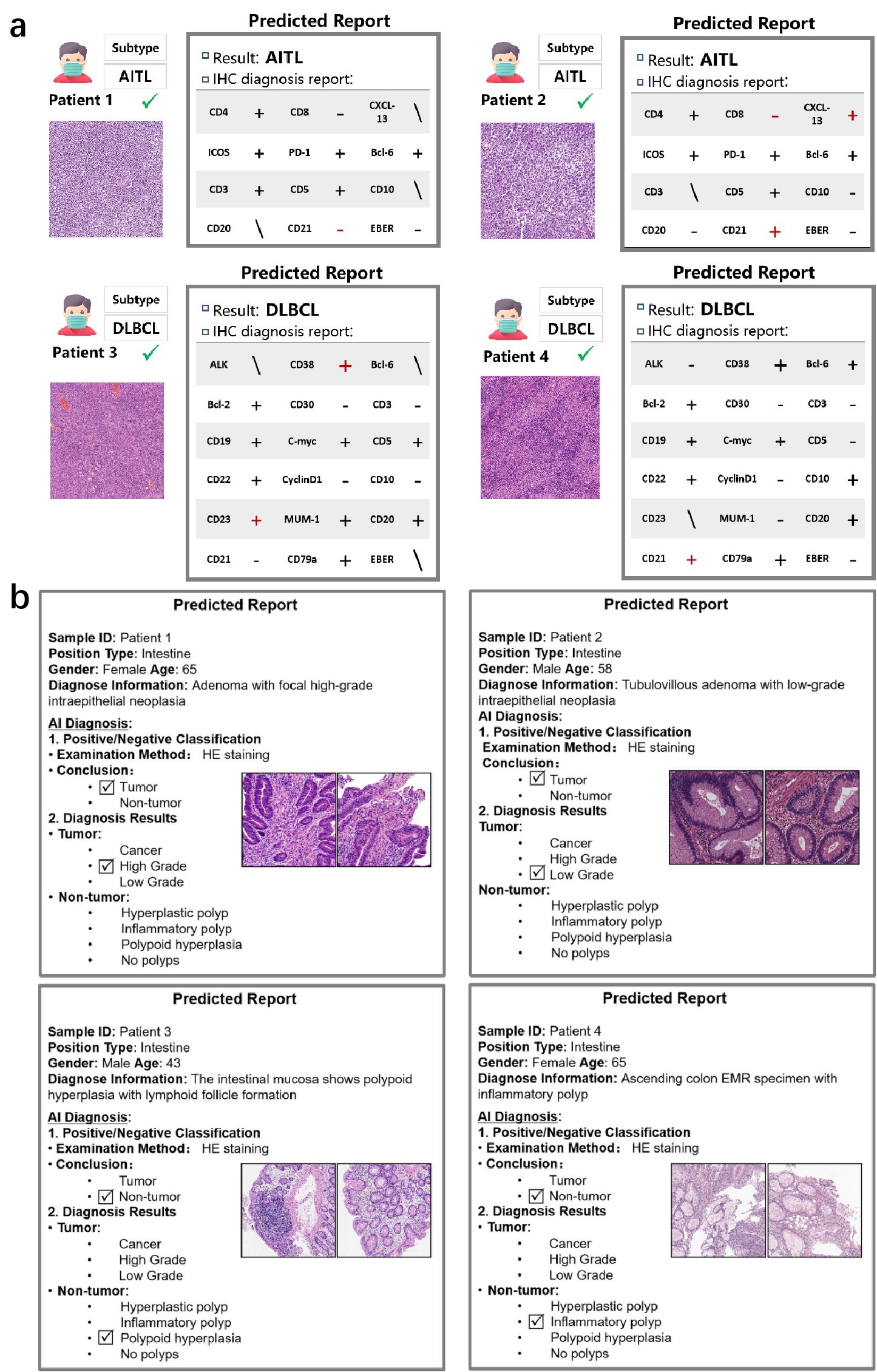}
	\caption{Structured Report Results for Lymphoma and Intestine. (Continued on next page.)}
	\label{fig.Fig7_StructuredReport}
\end{figure*}

\begin{figure*}[!htbp]
	\ContinuedFloat
	\caption{(Continued) a. Four patients diagnosed with lymphoma, including AITL and DLBCL subtypes, were selected to demonstrate the structured report generated by the PathOrchestra model. This report integrates predictions for lymphoma subtypes and qualitative IHC tasks across 30 combined tasks. It includes subtype diagnosis based on H\&E-stained images and the corresponding IHC results, with the IHC panels varying for each subtype. b. The pathology report for colorectal cancer was streamlined to focus on predicting malignancy (positive and negative) and tumor grading (cancer, low-grade, high-grade) based on H\&E-stained images. Additionally, for negative samples, the model further classifies the type of polyps.}
\end{figure*}

As illustrated in Figure~\ref{fig.Fig6_Gene}, we compared the performance of PathOrchestra with other foundational models, GigaPath and UNI, in predicting gene expression across multiple cancers. PathOrchestra demonstrated superior performance in five cancer types: LUAD, READ, SKCM, PRAD, and IDC. On average, the model outperformed UNI by 1\% and GigaPath by 5.7\% across these five cancer types. In the PRAD dataset, PathOrchestra achieved 6.5\% higher accuracy compared to UNI and 6.1\% higher than GigaPath. For the IDC dataset, although all three models performed above 0.700, PathOrchestra exceeded UNI by 4.4\% and GigaPath by 1\%. Additionally, we conducted a detailed comparison of gene prediction capabilities between PathOrchestra and UNI across LUAD, PRAD, READ, and SKCM. The proposed model showed relatively stable performance across multiple gene categories, with particularly strong results in the PRAD dataset, significantly surpassing UNI for genes such as MT-ATP6, PRL34, PRL12, PRL26, and PRL29. These findings demonstrate that PathOrchestra is capable of predicting gene expression profiles from H\&E-stained images, identifying distinct gene expression patterns across various cancers.

\noindent\textbf{Structured Report Generation Tasks} \\
Generating pathology reports is one of the most challenging clinical tasks in computational pathology, primarily due to the large number of WSIs per patient, the complexity of diagnostic content, and the difficulty in predicting subtasks. By selecting appropriate disease types and breaking down the pathology report into a structured format, we can use artificial intelligence algorithms to generate pathology reports from diagnostic and auxiliary examination perspectives. This approach can effectively save time of pathologists while providing reasonable diagnostic explanations, thus holding significant value for clinical practice. This study focuses on two diseases, lymphoma and colorectal cancer, employing a subtask combination method to generate structured reports for each patient based on the PathOrchestra model.

In the task of determining lymphoma subtypes, conventional HE image diagnoses often do not provide sufficient evidence to identify specific subtypes, usually requiring quantitative analysis from IHC, with different subtype suspected patients corresponding to different immunohistochemical panels. This paper emphasizes the AITL and DLBCL subtypes, combining the HE subtype prediction task with the qualitative assessment of 29 IHC markers, and diagnosing multiple differently stained pathology slides for each patient, resulting in structured reports. As seen in example from Figure~\ref{fig.Fig7_StructuredReport}a, the model demonstrates the ability to diagnose lymphoma subtypes and provides detailed diagnostic explanations via immunohistochemistry. For instance, in Patient 1, the algorithm incorrectly predicted CD21 as negative, and due to data loss during the collection process, it was unable to provide effective diagnoses for CD20, CXCL-13, and CD10, while the qualitative assessments of the other IHC markers were consistent with the actual diagnoses by pathologists. The results of subtype diagnosis and accompanying IHC reports indicate that our model is capable of making comprehensive diagnoses for this complex disease.

Colorectal cancer is a prevalent disease with a complex initial report. This study focuses solely on the assessment of HE images to distinguish between tumor and non-tumor, as well as to classify cancer grades (cancerous, low-grade cancer, high-grade cancer). Additionally, for negative samples, we further classify polyp types (including hyperplastic polyps, inflammatory polyps, polypoid hyperplasia, and no polyps). As illustrated in example from Figure~\ref{fig.Fig7_StructuredReport}b, the model exhibits the ability to screen for colorectal cancer and determine the severity of the tumor. For instance, Patient 1 was diagnosed by the algorithm as a tumor type and identified the tumor grade as high-grade. This diagnosis aligns with the actual result of adenoma with focal high-grade intraepithelial neoplasia. Thus, our model demonstrates the capability to generate preliminary diagnostic reports for this common disease.

\noindent\textbf{\large{Discussion}} \\
The recent emergence of foundational pathology models has revolutionized computational pathology, particularly in addressing rare diseases, reducing training annotations, and enhancing the training efficiency of downstream tasks as well as accelerating the development of complex, structured tasks. In this study, we introduced PathOrchestra, a self-supervised vision encoder trained on an unprecedented scale of 300K WSIs encompassing 20 different human tissues and organs. By evaluating its performance across 112 clinically relevant downstream tasks, which triples the task volume of the current leading UNI model, we have demonstrated its robust generalization capabilities and potential for integration into clinical practice.

There is no effective consensus on benchmarks for evaluating foundational pathology models, with unclear definitions on the perspective (technical or clinical), specific tasks, and the scope of evaluation (number of tasks, classifications, representative tasks, etc.). Although previous studies have performed multi-faceted evaluations, such as the UNI model covering 34 types of downstream tasks, these evaluations are predominantly from a technical standpoint. How to effectively utilize foundational pathology models and integrate them with actual clinical needs to maximize their value and promote their development at both technical and application levels remains a critical issue. A significant contribution of this study is the in-depth exploration of how large pathology models can be integrated with clinical applications to more effectively promote their practical implementation. By addressing a wide array of tasks, including digital slide preprocessing, pan-cancer classification, lesion identification, multi-cancer subtyping, biomarker assessment, gene expression prediction, and structured report generation, we have showcased the model's versatility and its potential to meet real-world clinical needs. 

The multifaceted capabilities of PathOrchestra in structured report generation, pan-cancer classification, and gene expression prediction signify its vital role in advancing clinical integration and enhancing diagnostic practices for complex diseases. First, our exploration into structured report generation for complex diseases such as lymphoma and colorectal cancer represents a significant step toward practical clinical integration. By decomposing pathology reports into structured formats and combining multiple subtasks including subtype diagnosis and qualitative IHC marker analysis, PathOrchestra effectively generates comprehensive reports that can enhance diagnostic accuracy and efficiency. This approach not only alleviates the workload of pathologists but also contributes to standardized reporting practices. Second, the impressive performance of PathOrchestra in pan-cancer classification tasks underscores its ability to generalize across diverse cancer types, specimen preparation methods, and data sources. Achieving AUCs exceeding 0.930 in both in-house and TCGA datasets, the model demonstrates robust diagnostic capabilities that are crucial for early cancer detection and personalized treatment planning. Third, its superior performance in gene expression prediction tasks outperforming models like GigaPath and UNI by significant margins, highlights its potential in advancing molecular pathology by providing insights into gene expression profiles directly from H\&E-stained images.

The reasons for the model's enhanced performance in downstream tasks can be summarized in three key aspects. First, the model was trained on 300K high-quality WSI pathology images ensuring rich feature learning with a 304M parameter model. Second, the data was sourced from three major centers, encompassing diverse populations from Asia, the Americas, and other regions, enhancing the model's adaptability and generalization, leading to more reliable performance across various clinical contexts. Finally, the implementation of a self-supervised learning strategy effectively mined features from unlabeled data, improving the model's representation capabilities for pathology images and its transfer learning efficiency across different tasks. 

Despite above advancements, several limitations warrant discussion. First, while the model demonstrates strong performance in tumor-related tasks, its capability in handling inflammatory conditions and distinguishing between neoplastic and reactive processes remains limited. The pan-cancer evaluations primarily focused on tumor types, and future work should incorporate a broader spectrum of pathological conditions, including inflammatory and infectious diseases, to enhance the model's diagnostic utility. Second, the imbalance in training data across different tissue types and disease categories may influence the model's performance. Although we utilized a vast dataset, ensuring balanced representation is essential for unbiased learning and generalization. Future studies will investigate the effects of data balance and explore strategies such as data augmentation or synthetic data generation to address this issue. Third, while the model excels in numerous tasks, integrating it seamlessly into clinical workflows poses challenges. Clinical environments demand not only high accuracy but also interpretability, ease of use, and compliance with regulatory standards. Further research is needed to enhance the model's interpretability, possibly through explainable AI techniques, and to develop user-friendly interfaces that facilitate adoption by healthcare professionals. Fourth, the unique characteristics of pathological data such as the gigapixel scale of WSIs and the necessity for multi-resolution analysis, call for specialized model architectures and training strategies. Future work should focus on optimizing the model for WSI-level analysis, incorporating hierarchical and multi-scale representations to capture both global tissue architecture and cellular-level details. Fifth, this paper does not achieve professional structured report generation, particularly for colorectal cancer, which is highly complex in clinical practice, involving the comprehensive handling of 40 subtasks and the joint diagnosis of dozens of slides from the same patient. Additionally, it must address more detailed yet equally important subtasks, such as reporting vascular invasion and margin assessment in the future. Finally, the exploration of multimodal approaches represents another promising avenue. Integrating data from various modalities, such as genomics, proteomics, and radiology, alongside pathology images, could provide a more comprehensive understanding of disease processes. Additionally, developing models capable of natural language processing tasks such as interpreting clinical notes or generating narrative reports, could further enhance the model's clinical applicability.

In conclusion, PathOrchestra represents a substantial advancement in the development of foundational pathology models, demonstrating exceptional performance across a wide range of clinically relevant tasks. By addressing both technical and clinical perspectives, this work contributes to bridging the gap between computational pathology research and real-world clinical practice. Ongoing efforts to refine the model, expand its capabilities, and facilitate its integration into healthcare settings hold the promise of significantly improving diagnostic accuracy, personalized treatment, and ultimately patient outcomes in the future.

\noindent\textbf{Computing Hardware and Software} \\ For all experiments and analyses conducted in this study (unless otherwise specified), we utilized Python (version 3.9) and PyTorch (version 2.0.0, CUDA 11.8) (\href{https://pytorch.org/}{pytorch.org}). These experiments can be replicated using the open-source libraries detailed below. To train PathOrchestra via DINOv2, we adapted the Vision Transformer implementation from the Timm library (version 0.9.2) maintained by Hugging Face (\href{https://huggingface.co/}{huggingface.co}) for the encoder backbone, and employed the original DINOv2 self-supervised learning algorithm (\href{https://github.com/facebookresearch/dinov2}{github.com/facebookresearch/dinov2}) for pretraining. This setup utilized 4$\times$8 80GB NVIDIA A100 GPU nodes configured for multi-GPU, multi-node training with distributed data-parallel (DDP). All additional computations for downstream experiments were conducted on single 24GB NVIDIA 3090 GPUs, 4090 GPUs, and 80GB NVIDIA A100 GPUs. WSI processing was supported by OpenSlide (version 4.3.1) and openslide-python (version 1.2.0). For training weakly-supervised models, we adapted the training scaffold code from the ABMIL codebase (\href{https://github.com/AMLab-Amsterdam/AttentionDeepMIL}{github.com/AMLab-Amsterdam/AttentionDeepMIL}). For semantic segmentation training, we used the original Mask2Former implementation (\href{https://github.com/facebookresearch/Mask2Former}{github.com/facebookresearch/Mask2Former}). Pillow (version 9.3.0) and OpenCV-python were employed for basic image processing tasks. Matplotlib (version 3.7.1) and Seaborn (version 0.12.2) were used for creating plots and figures. Figure~\ref{fig.Fig1_Overview} in this paper were created using the BioRender platform (\href{https://www.biorender.com/}{biorender.com/}).

\noindent\textbf{\large{Data Availability}} \\
Public datasets can be available at: 
FocusPath-UofT\footnote{\href{https://sites.google.com/view/focuspathuoft/database}{https://sites.google.com/view/focuspathuoft/database}}, 
CAMELYON16 \footnote{\href{https://camelyon16.grand-challenge.org}{https://camelyon16.grand\-challenge.org}}, 
CAMELYON17 \footnote{\href{https://camelyon17.grand-challenge.org}{https://camelyon17.grand\-challenge.org}}, 
TCGA-TILs \footnote{\href{https://zenodo.org/records/6604094}{https://zenodo.org/records/6604094}}, 
Pcam \footnote{\href{https://github.com/basveeling/pcam}{https://github.com/basveeling/pcam}}, 
GlaS \footnote{\href{https://www.kaggle.com/datasets/sani84/glasmiccai2015-gland-segmentation}{https://www.kaggle.com/datasets/sani84/glasmiccai2015-gland-segmentation}}, 
PanNuke \footnote{\href{https://link.springer.com/chapter/10.1007/978-3-030-23937-4\_2}{https://link.springer.com/chapter/10.1007/978-3-030-23937-4\_2}},
CoNSeP \footnote{\href{https://paperswithcode.com/dataset/consep}{https://paperswithcode.com/dataset/consep}},
COSAS \footnote{\href{https://cosas.grand-challenge.org/teams/}{https://cosas.grand-challenge.org/teams/}}, 
TissueNet \footnote{\href{https://www.drivendata.org/competitions/67/competition-cervical-biopsy/page/255/}{https://www.drivendata.org/competitions/67/competition-cervical-biopsy/page/255/}},
LC25K \footnote{\href{https://github.com/tampapath/lung\_colon\_image\_set}{https://github.com/tampapath/lung\_colon\_image\_set}},
BreakHis \footnote{\href{https://web.inf.ufpr.br/vri/databases/breast-cancer-histopathological-database-breakhis/}{https://web.inf.ufpr.br/vri/databases/breast-cancer-histopathological-database-breakhis/}}, 
TCGA-NSCLC \footnote{\href{https://portal.gdc.cancer.gov/}{https://portal.gdc.cancer.gov/}}, 
TCGA-RCC \footnote{\href{https://portal.gdc.cancer.gov/}{https://portal.gdc.cancer.gov/}},
BACH \footnote{\href{https://iciar2018-challenge.grand-challenge.org/Dataset/}{https://iciar2018-challenge.grand-challenge.org/Dataset/}},
TCGA-ESCA \footnote{\href{https://zenodo.org/record/7548828}{https://zenodo.org/record/7548828}},
HunCRC \footnote{\href{https://www.nature.com/articles/s41597-022-01450-y}{https://www.nature.com/articles/s41597-022-01450-y}},
PANDA \footnote{\href{https://panda.grand-challenge.org/data/}{https://panda.grand-challenge.org/data/}}, PatchGastricADC22 \footnote{\href{https://zenodo.org/records/6550925}{https://zenodo.org/records/6550925}},
AGGC \footnote{\href{https://aggc22.grand-challenge.org}{https://aggc22.grand-challenge.org}},
TCGA-IDH1 \footnote{\href{https://www.nature.com/articles/s41597-022-01450-y}{https://www.nature.com/articles/s41597-022-01450-y}},
CRC-100K \footnote{\href{https://zenodo.org/records/1214456}{https://zenodo.org/records/1214456}}, 
Chaoyang \footnote{\href{https://bupt-ai-cz.github.io/HSA-NRL/}{https://bupt-ai-cz.github.io/HSA-NRL/}},
WSSS4LUAD \footnote{\href{https://wsss4luad.grand-challenge.org/}{https://wsss4luad.grand-challenge.org/}},
Kather \footnote{\href{https://zenodo.org/records/53169}{https://zenodo.org/records/53169}},
Ebrains \footnote{\href{https://search.kg.ebrains.eu/instances/Dataset/8fc108ab-e2b4-406f-8999-60269dc1f994}{https://search.kg.ebrains.eu/instances/Dataset/8fc108ab-e2b4-406f-8999-60269dc1f994}},
HEST \footnote{\href{https://github.com/mahmoodlab/HEST}{https://github.com/mahmoodlab/HEST}},
DeepCell \footnote{\href{https://datasets.deepcell.org/data}{https://datasets.deepcell.org/data}},
SegPath
\footnote{\href{https://dakomura.github.io/SegPath/}{https://dakomura.github.io/SegPath/}}.

\noindent\textbf{\large{Code Availability}} \\
The model weights to train PathOrchestra will be made available for non-commercial academic use once the article is published.

\noindent\textbf{\large{Author Contributions}}\\
This paper was primarily written by FY, JFW, JWL, and WW. FY, JWL, WC, JXL, and TS conceived the study and designed the experiments. FY, WC, MDC, YL, QC, TS performed data collection and processing. FY, SYH performed model development. ZZG, JNL, HL, HY, YZW, XTL, XNW, ZHW, QH performed experimental analysis and interpreted the results. JFW, WW, MXL, and LM reviewed and revised the manuscript. XFZ, YHH, HC, STZ and ZW supervised the research.

\noindent\textbf{\large{Acknowledgements}}\\
This work was supported by Shanghai Artificial Intelligence Laboratory.

\noindent\textbf{\large{Competing Interests}}\\
The authors have declared that no competing interests exist.
\end{spacing}

\clearpage

\begin{nolinenumbers}
\clearpage
\section*{References} 
\vspace{2mm}

\begin{spacing}{0.9}
\bibliographystyle{IEEEtran}
\bibliography{sample}

\begin{thebibliography}{10}
\providecommand{\url}[1]{#1}
\csname url@samestyle\endcsname
\providecommand{\newblock}{\relax}
\providecommand{\bibinfo}[2]{#2}
\providecommand{\BIBentrySTDinterwordspacing}{\spaceskip=0pt\relax}
\providecommand{\BIBentryALTinterwordstretchfactor}{4}
\providecommand{\BIBentryALTinterwordspacing}{\spaceskip=\fontdimen2\font plus
\BIBentryALTinterwordstretchfactor\fontdimen3\font minus
  \fontdimen4\font\relax}
\providecommand{\BIBforeignlanguage}[2]{{%
\expandafter\ifx\csname l@#1\endcsname\relax
\typeout{** WARNING: IEEEtran.bst: No hyphenation pattern has been}%
\typeout{** loaded for the language `#1'. Using the pattern for}%
\typeout{** the default language instead.}%
\else
\language=\csname l@#1\endcsname
\fi
#2}}
\providecommand{\BIBdecl}{\relax}
\BIBdecl

\bibitem{tolkach2023artificial}
Y.~Tolkach, L.~M. Wolgast, A.~Damanakis, A.~Pryalukhin, S.~Schallenberg,
  W.~Hulla, M.~L. Eich, W.~Schroeder, A.~Mukhopadhyay, M.~Fuchs \emph{et~al.},
  ``Artificial intelligence for tumour tissue detection and histological
  regression grading in oesophageal adenocarcinomas: a retrospective algorithm
  development and validation study,'' \emph{The Lancet Digital Health}, vol.~5,
  no.~5, pp. 265--275, 2023.

\bibitem{ding2023large}
K.~Ding, M.~Zhou, H.~Wang, O.~Gevaert, D.~Metaxas, and S.~Zhang, ``A
  large-scale synthetic pathological dataset for deep learning-enabled
  segmentation of breast cancer,'' \emph{Scientific Data}, vol.~10, no.~1, pp.
  231--241, 2023.

\bibitem{lu2021data}
M.~Y. Lu, D.~F. Williamson, T.~Y. Chen, R.~J. Chen, M.~Barbieri, and
  F.~Mahmood, ``Data-efficient and weakly supervised computational pathology on
  whole-slide images,'' \emph{Nature Biomedical Engineering}, vol.~5, no.~6,
  pp. 555--570, 2021.

\bibitem{anaya2024multiple}
J.~Anaya, J.-W. Sidhom, F.~Mahmood, and A.~S. Baras, ``Multiple-instance
  learning of somatic mutations for the classification of tumour type and the
  prediction of microsatellite status,'' \emph{Nature Biomedical Engineering},
  vol.~8, no.~1, pp. 57--67, 2024.

\bibitem{nagpal2020development}
K.~Nagpal, D.~Foote, F.~Tan, Y.~Liu, P.-H.~C. Chen, D.~F. Steiner, N.~Manoj,
  N.~Olson, J.~L. Smith, A.~Mohtashamian \emph{et~al.}, ``Development and
  validation of a deep learning algorithm for gleason grading of prostate
  cancer from biopsy specimens,'' \emph{JAMA Oncology}, vol.~6, no.~9, pp.
  1372--1380, 2020.

\bibitem{madabhushi2020deep}
A.~Madabhushi, M.~D. Feldman, and P.~Leo, ``Deep-learning approaches for
  gleason grading of prostate biopsies,'' \emph{The Lancet Oncology}, vol.~21,
  no.~2, pp. 187--189, 2020.

\bibitem{shamai2022deep}
G.~Shamai, A.~Livne, A.~Pol{\'o}nia, E.~Sabo, A.~Cretu, G.~Bar-Sela, and
  R.~Kimmel, ``Deep learning-based image analysis predicts pd-l1 status from
  h\&e-stained histopathology images in breast cancer,'' \emph{Nature
  Communications}, vol.~13, no.~1, p. 6753, 2022.

\bibitem{he2020integrating}
B.~He, L.~Bergenstr{\aa}hle, L.~Stenbeck, A.~Abid, A.~Andersson, {\AA}.~Borg,
  J.~Maaskola, J.~Lundeberg, and J.~Zou, ``Integrating spatial gene expression
  and breast tumour morphology via deep learning,'' \emph{Nature Biomedical
  Engineering}, vol.~4, no.~8, pp. 827--834, 2020.

\bibitem{kather2020pan}
J.~N. Kather, L.~R. Heij, H.~I. Grabsch, C.~Loeffler, A.~Echle, H.~S. Muti,
  J.~Krause, J.~M. Niehues, K.~A. Sommer, P.~Bankhead \emph{et~al.},
  ``Pan-cancer image-based detection of clinically actionable genetic
  alterations,'' \emph{Nature Cancer}, vol.~1, no.~8, pp. 789--799, 2020.

\bibitem{jaume2024hest}
G.~Jaume, P.~Doucet, A.~H. Song, M.~Y. Lu, C.~Almagro-P{\'e}rez, S.~J. Wagner,
  A.~J. Vaidya, R.~J. Chen, D.~F. Williamson, A.~Kim \emph{et~al.}, ``Hest-1k:
  A dataset for spatial transcriptomics and histology image analysis,''
  \emph{arXiv preprint arXiv:2406.16192}, 2024.

\bibitem{wulczyn2021interpretable}
E.~Wulczyn, D.~F. Steiner, M.~Moran, M.~Plass, R.~Reihs, F.~Tan,
  I.~Flament-Auvigne, T.~Brown, P.~Regitnig, P.-H.~C. Chen \emph{et~al.},
  ``Interpretable survival prediction for colorectal cancer using deep
  learning,'' \emph{NPJ Digital Medicine}, vol.~4, no.~1, pp. 71--81, 2021.

\bibitem{jiang2024end}
X.~Jiang, M.~Hoffmeister, H.~Brenner, H.~S. Muti, T.~Yuan, S.~Foersch, N.~P.
  West, A.~Brobeil, J.~Jonnagaddala, N.~Hawkins \emph{et~al.}, ``End-to-end
  prognostication in colorectal cancer by deep learning: a retrospective,
  multicentre study,'' \emph{The Lancet Digital Health}, vol.~6, no.~1, pp.
  33--43, 2024.

\bibitem{foersch2023multistain}
S.~Foersch, C.~Glasner, A.-C. Woerl, M.~Eckstein, D.-C. Wagner, S.~Schulz,
  F.~Kellers, A.~Fernandez, K.~Tserea, M.~Kloth \emph{et~al.}, ``Multistain
  deep learning for prediction of prognosis and therapy response in colorectal
  cancer,'' \emph{Nature Medicine}, vol.~29, no.~2, pp. 430--439, 2023.

\bibitem{yu2016predicting}
K.-H. Yu, C.~Zhang, G.~J. Berry, R.~B. Altman, C.~R{\'e}, D.~L. Rubin, and
  M.~Snyder, ``Predicting non-small cell lung cancer prognosis by fully
  automated microscopic pathology image features,'' \emph{Nature
  Communications}, vol.~7, no.~1, p. 12474, 2016.

\bibitem{lipkova2022deep}
J.~Lipkova, T.~Y. Chen, M.~Y. Lu, R.~J. Chen, M.~Shady, M.~Williams, J.~Wang,
  Z.~Noor, R.~N. Mitchell, M.~Turan \emph{et~al.}, ``Deep learning-enabled
  assessment of cardiac allograft rejection from endomyocardial biopsies,''
  \emph{Nature Medicine}, vol.~28, no.~3, pp. 575--582, 2022.

\bibitem{oquab2023dinov2}
M.~Oquab, T.~Darcet, T.~Moutakanni, H.~Vo, M.~Szafraniec, V.~Khalidov,
  P.~Fernandez, D.~Haziza, F.~Massa, A.~El-Nouby \emph{et~al.}, ``Dinov2:
  Learning robust visual features without supervision,'' \emph{arXiv preprint
  arXiv:2304.07193}, 2023.

\bibitem{chen2021empirical}
X.~Chen, S.~Xie, and K.~He, ``An empirical study of training self-supervised
  vision transformers,'' in \emph{Proceedings of the IEEE/CVF International
  Conference on Computer Vision (CVPR)}, 2021, pp. 9640--9649.

\bibitem{he2022masked}
K.~He, X.~Chen, S.~Xie, Y.~Li, P.~Doll{\'a}r, and R.~Girshick, ``Masked
  autoencoders are scalable vision learners,'' in \emph{Proceedings of the
  IEEE/CVF Conference on Computer Vision and Pattern Recognition (CVPR)}, 2022,
  pp. 16\,000--16\,009.

\bibitem{achiam2023gpt}
J.~Achiam, S.~Adler, S.~Agarwal, L.~Ahmad, I.~Akkaya, F.~L. Aleman, D.~Almeida,
  J.~Altenschmidt, S.~Altman, S.~Anadkat \emph{et~al.}, ``Gpt-4 technical
  report,'' \emph{arXiv preprint arXiv:2303.08774}, 2023.

\bibitem{brown2020language}
T.~Brown, B.~Mann, N.~Ryder, M.~Subbiah, J.~D. Kaplan, P.~Dhariwal,
  A.~Neelakantan, P.~Shyam, G.~Sastry, A.~Askell \emph{et~al.}, ``Language
  models are few-shot learners,'' \emph{Advances in Neural Information
  Processing Systems}, vol.~33, pp. 1877--1901, 2020.

\bibitem{dosovitskiy2020image}
A.~Dosovitskiy, ``An image is worth 16x16 words: Transformers for image
  recognition at scale,'' \emph{International Conference on Learning
  Representations (ICLR)}, 2021.

\bibitem{he2020momentum}
K.~He, H.~Fan, Y.~Wu, S.~Xie, and R.~Girshick, ``Momentum contrast for
  unsupervised visual representation learning,'' in \emph{Proceedings of the
  IEEE/CVF Conference on Computer Vision and Pattern Recognition (CVPR)}, 2020,
  pp. 9729--9738.

\bibitem{caron2021emerging}
M.~Caron, H.~Touvron, I.~Misra, H.~J{\'e}gou, J.~Mairal, P.~Bojanowski, and
  A.~Joulin, ``Emerging properties in self-supervised vision transformers,'' in
  \emph{Proceedings of the IEEE/CVF International Conference on Computer Vision
  (ICCV)}, 2021, pp. 9650--9660.

\bibitem{zhou2021ibot}
J.~Zhou, C.~Wei, H.~Wang, W.~Shen, C.~Xie, A.~Yuille, and T.~Kong, ``ibot:
  Image bert pre-training with online tokenizer,'' \emph{arXiv preprint
  arXiv:2111.07832}, 2021.

\bibitem{sablayrolles2018spreading}
A.~Sablayrolles, M.~Douze, C.~Schmid, and H.~J{\'e}gou, ``Spreading vectors for
  similarity search,'' \emph{International Conference on Learning
  Representations (ICLR)}, 2019.

\bibitem{hosseini2019encoding}
M.~S. Hosseini, Y.~Zhang, and K.~N. Plataniotis, ``Encoding visual sensitivity
  by maxpol convolution filters for image sharpness assessment,'' \emph{IEEE
  Transactions on Image Processing}, vol.~28, no.~9, pp. 4510--4525, 2019.

\bibitem{greenwald2022whole}
N.~F. Greenwald, G.~Miller, E.~Moen, A.~Kong, A.~Kagel, T.~Dougherty, C.~C.
  Fullaway, B.~J. McIntosh, K.~X. Leow, M.~S. Schwartz \emph{et~al.},
  ``Whole-cell segmentation of tissue images with human-level performance using
  large-scale data annotation and deep learning,'' \emph{Nature Biotechnology},
  vol.~40, no.~4, pp. 555--565, 2022.

\bibitem{ilse2018attention}
M.~Ilse, J.~Tomczak, and M.~Welling, ``Attention-based deep multiple instance
  learning,'' in \emph{International Conference on Machine Learning
  (ICML)}.\hskip 1em plus 0.5em minus 0.4em\relax PMLR, 2018, pp. 2127--2136.

\bibitem{cheng2022masked}
B.~Cheng, I.~Misra, A.~G. Schwing, A.~Kirillov, and R.~Girdhar,
  ``Masked-attention mask transformer for universal image segmentation,'' in
  \emph{Proceedings of the IEEE/CVF Conference on Computer Vision and Pattern
  Recognition (CVPR)}, 2022, pp. 1290--1299.

\bibitem{chen2022vitadapter}
Z.~Chen, Y.~Duan, W.~Wang, J.~He, T.~Lu, J.~Dai, and Y.~Qiao, ``Vision
  transformer adapter for dense predictions,'' \emph{International Conference
  on Learning Representations (ICLR)}, 2023.

\bibitem{li2022exploring}
Y.~Li, H.~Mao, R.~Girshick, and K.~He, ``Exploring plain vision transformer
  backbones for object detection,'' in \emph{Proceedings of the IEEE/CVF
  Conference on European Conference on Computer Vision (ECCV)}.\hskip 1em plus
  0.5em minus 0.4em\relax Springer, 2022, pp. 280--296.

\bibitem{peterson2009k}
L.~E. Peterson, ``K-nearest neighbor,'' \emph{Scholarpedia}, vol.~4, no.~2, p.
  1883, 2009.

\bibitem{ronneberger2015u}
O.~Ronneberger, P.~Fischer, and T.~Brox, ``U-net: Convolutional networks for
  biomedical image segmentation,'' in \emph{Medical Image Computing and
  Computer-assisted Intervention (MICCAI)}.\hskip 1em plus 0.5em minus
  0.4em\relax Springer, 2015, pp. 234--241.

\bibitem{roetzer2022digital}
T.~Roetzer-Pejrimovsky, A.-C. Moser, B.~Atli, C.~C. Vogel, P.~A. Mercea,
  R.~Prihoda, E.~Gelpi, C.~Haberler, R.~H{\"o}ftberger, J.~A. Hainfellner
  \emph{et~al.}, ``The digital brain tumour atlas, an open histopathology
  resource,'' \emph{Scientific Data}, vol.~9, no. 1--10, p.~55, 2022.

\bibitem{janesick2023high}
A.~Janesick, R.~Shelansky, A.~D. Gottscho, F.~Wagner, S.~R. Williams,
  M.~Rouault, G.~Beliakoff, C.~A. Morrison, M.~F. Oliveira, J.~T. Sicherman
  \emph{et~al.}, ``High resolution mapping of the tumor microenvironment using
  integrated single-cell, spatial and in situ analysis,'' \emph{Nature
  Communications}, vol.~14, no. 1--10, p. 8353, 2023.

\bibitem{erickson2022spatially}
A.~Erickson, M.~He, E.~Berglund, M.~Marklund, R.~Mirzazadeh, N.~Schultz,
  L.~Kvastad, A.~Andersson, L.~Bergenstr{\aa}hle, J.~Bergenstr{\aa}hle
  \emph{et~al.}, ``Spatially resolved clonal copy number alterations in benign
  and malignant tissue,'' \emph{Nature}, vol. 608, no. 7922, pp. 360--367,
  2022.

\bibitem{valdeolivas2023charting}
A.~Valdeolivas, B.~Amberg, N.~Giroud, M.~Richardson, E.~J. G{\'a}lvez,
  S.~Badillo, A.~Julien-Laferri{\`e}re, D.~Turos, L.~V. von Voithenberg,
  I.~Wells \emph{et~al.}, ``Charting the heterogeneity of colorectal cancer
  consensus molecular subtypes using spatial transcriptomics,'' \emph{bioRxiv},
  pp. 2023--01, 2023.

\bibitem{meylan2022tertiary}
M.~Meylan, F.~Petitprez, E.~Becht, A.~Bougo{\"u}in, G.~Pupier, A.~Calvez,
  I.~Giglioli, V.~Verkarre, G.~Lacroix, J.~Verneau \emph{et~al.}, ``Tertiary
  lymphoid structures generate and propagate anti-tumor antibody-producing
  plasma cells in renal cell cancer,'' \emph{Immunity}, vol.~55, no.~3, pp.
  527--541, 2022.

\bibitem{giraud2022trem1}
J.~Giraud, D.~Chalopin, E.~Ramel, T.~Boyer, A.~Zouine, M.-A. Derieppe,
  N.~Larmonier, O.~Adotevi, B.~L. Bail, J.-F. Blanc \emph{et~al.}, ``Trem1+
  regulatory myeloid cells expand in steatohepatitis-hcc and associate with
  poor prognosis and therapeutic resistance to immune checkpoint blockade,''
  \emph{bioRxiv}, pp. 2022--11, 2022.

\bibitem{liu2022single}
T.~Liu, C.~Liu, M.~Yan, L.~Zhang, J.~Zhang, M.~Xiao, Z.~Li, X.~Wei, and
  H.~Zhang, ``Single cell profiling of primary and paired metastatic lymph node
  tumors in breast cancer patients,'' \emph{Nature Communications}, vol.~13,
  no.~1, p. 6823, 2022.

\end{thebibliography}
\end{spacing}
\end{nolinenumbers}

\clearpage
\noindent\textbf{\large{APPENDIX I}}

\noindent\textbf{Clinical Downstream Task Details}\\
112 clinically relevant downstream tasks encompass pathology quality control (QC), pan-cancer classification, lesion detection, multi-cancer subtyping, biomarker assessment, gene expression prediction, and structured report generation. We provide detailed descriptions of selected tasks below, including datasets, experimental setups, and training parameters.

\noindent\textbf{I. Pathology Image Preprocessing and Quality Control Tasks}\\
\noindent\textit{1. Pathology QC: Natural vs. Pathological Image Classification}. We developed a binary classifier to distinguish between natural images and pathological images. Natural images were sourced from the ImageNet-1K validation set (ILSVRC2012\_val), while pathological images were randomly selected patches from gastrointestinal H\&E and IHC image data. The dataset was split into training and testing sets in a 6:4 ratio, with an approximate distribution of natural images to H\&E images to IHC images of 5:3:2. Pathological images, originally sized at 512 $\times$ 512 pixels, were resized to 256 $\times$ 256 pixels for model input. The training set contained 13,113 positive (pathological) and 13,119 negative (natural) samples, while the test set included 8,752 positive and 8,746 negative samples. We employed linear probing as the training strategy, training a linear classifier on top of the frozen pre-trained features for up to 1,000 iterations. All training operations were conducted on NVIDIA Tesla V100 GPUs.

\noindent\textit{2. Pathology QC: Tissue Fold Detection}. In the fold detection task, overlapping tissue sections due to folding obscure the complete tissue structure, making pathological analysis challenging. These folds may also exhibit uneven staining, further complicating the recognition and analysis of pathological regions. We selected 512$\times$512 pixel patch images from gastrointestinal tissue sections, resized to 256$\times$256 pixels for model input. The positive training set contained 12,803 patches, the positive test set had 5,156 patches, the negative training set included 11,445 patches, and the negative test set had 8,057 patches. Linear Probing was used as the training strategy, with training parameters set to a maximum of 1,000 iterations. All training operations were conducted on NVIDIA Tesla V100 GPUs.

\noindent\textit{3. Pathology QC: Bubble and Glue Classification}. Bubbles and glue artifacts in tissue sections can interfere with the observation of tissue structures and staining patterns, leading to diagnostic challenges. For this task, we selected 256$\times$256 pixel patch images from H\&E sections. The training set included 1,610 bubble images and 4,317 glue images, while the test set consisted of 1,232 bubble images and 5,273 glue images. We employed Linear Probing as the training strategy, with training parameters set to a maximum of 1,000 iterations. All training operations were conducted on NVIDIA Tesla V100 GPUs.

\noindent\textit{4. Pathology QC: Contaminant Detection}. Contaminants in pathological slides, such as dust, hair, and fibers, can significantly impede the accurate observation of tissue structures, potentially leading to diagnostic errors. Detecting and eliminating contaminated areas during slide preprocessing is essential for enhancing the accuracy and efficiency of pathological analysis. We selected 256$\times$256 pixel patch images from H\&E slides, with the training set containing 497 contaminated images and 4,638 uncontaminated images, and the test set comprising 389 contaminated images and 4,511 uncontaminated images. Linear Probing was used as the training strategy, with training parameters set to a maximum of 1,000 iterations. All training operations were conducted on NVIDIA Tesla V100 GPUs.

\noindent\textit{5. Pathology QC: Blur Detection}. In this quality control task, the focus was on identifying blurry images, utilizing the FocusPath-UofT~\cite{hosseini2019encoding} dataset for training and testing. The dataset was generated by scanning each slide along the Z-axis, producing 16 images with varying degrees of defocus (score range [-8, +8]). Images with scores between [-3, 3] were defined as clear, while those with scores in the ranges [-8, -4] and [4, 8] were classified as blurry. The dataset was divided into training and testing sets in a 6:4 ratio. The positive training set contained 291 patches, and the positive test set had 195 patches. The negative training set consisted of 226 patches, while the negative test set included 152 patches. Linear Probing was utilized as the training strategy, with training parameters set to a maximum of 1,000 iterations. All training operations were conducted on NVIDIA Tesla V100 GPUs.

\noindent\textit{6. Pan-Organ H\&E and IHC Staining Recognition}. This binary classification task aimed to distinguish between H\&E and IHC patch images. The ratio of H\&E to IHC images was approximately 6:4, and the dataset was divided into training and testing sets with the same ratio. The positive training set contained 7,871 patches, the positive test set had 5,248 patches, the negative training set included 5,242 patches, and the negative test set consisted of 3,504 patches. We employed Linear Probing as the training strategy, with training parameters set to a maximum of 1,000 iterations. All training operations were conducted on NVIDIA Tesla V100 GPUs.

\noindent\textit{7. Pan-Organ H\&E and Fluorescence Recognition}. The task focused on developing a model capable of distinguishing between ROI-level H\&E stained images and immunofluorescence images from DeepCell~\cite{greenwald2022whole}, enhancing the model's recognition capabilities during pathological preprocessing. We used a dataset comprising 15,022 images, including 8,000 H\&E and 7,022 fluorescence images, divided into 12,017 training images and 3,005 test images. Training parameters were set to a maximum of 1,000 iterations for linear probing, with a batch size of 256 images. Training was conducted on systems equipped with A100 GPUs to ensure efficiency. All images were resized to 256 pixels to meet the model's input requirements.

\noindent\textit{8. IHC Marker Type Recognition}. This task involved classifying four types of IHC markers: PR, ER, Ki67, and HER2. Recognizing IHC markers is fundamental in pathology, with the model expected to identify IHC staining, assisting pathologists in accurately determining the molecular characteristics of tumors. The dataset comprised 2,689 ROI images, split into training and testing sets in a 6:4 ratio. The training set contained 305 PR images, 461 ER images, 294 HER2 images, and 552 Ki67 images, while the test set included 204 PR images, 308 ER images, 196 HER2 images, and 369 Ki67 images. We used Linear Probing as the training strategy, with training parameters set to a maximum of 1,000 iterations. The ROIs were cropped at 10x magnification with a size of 256$\times$256 pixels. All training operations were conducted on NVIDIA Tesla V100 GPUs.

\noindent\textit{9. Pan-Organ Multi-Magnification Recognition}. This task aimed to design a model capable of distinguishing between H\&E images at different magnifications, enhancing the model's recognition capabilities during pathological preprocessing. The dataset comprised 286,712 images, including 143,017 10x and 143,695 20x H\&E images, split into 229,369 training images and 57,343 test images. During training, the maximum number of iterations for linear probing was set to 1,000, with a batch size of 256 images. All training operations were conducted on A100 GPUs, ensuring computational efficiency. Images were processed at 10x and 20x magnifications and resized to 256 pixels.

\noindent\textit{10. Pan-Organ Frozen vs. Formalin-Fixed Recognition}. This task involved distinguishing between H\&E images from FFPE and frozen sections to enhance the model's recognition capabilities during pathological preprocessing. The dataset comprised 289,469 images, including 146,452 frozen images and 143,017 FFPE images, divided into 231,575 training images and 57,894 test images. Linear probing was employed for training, with parameters set to 1,000 iterations and a batch size of 256 images. All training operations were conducted on A100 GPUs. Images were processed at 20x magnification and resized to 256 pixels.

\noindent\textit{11. Pan-Organ Biopsy vs. Large Specimen Recognition}. This task focuses on distinguishing between biopsy and large specimen H\&E images at the ROI level, aiming to enhance the model's recognition capabilities during pathological preprocessing. The dataset includes 726 images (449 large specimen images and 277 biopsy images), divided into 580 training images and 146 test images. Training parameters were set to a maximum of 1,000 iterations for linear probing, with a batch size of 256 images. All training operations were conducted on systems equipped with A100 GPUs. Images were processed at 20x magnification and resized to 256 pixels.

\noindent\textit{12. IHC Nuclear vs. Membrane Positivity Recognition}. In this task, the model classifies IHC markers based on nuclear positivity (PR, ER, Ki67) and membrane positivity (HER2). The dataset includes 1,812 ROI images, divided into a training set of 1,612 images and a test set of 400 images. Training parameters were set to a maximum of 1,000 iterations for linear probing, with a batch size of 256 images. All training operations were conducted on systems equipped with NVIDIA Tesla V100 GPUs. Images were sampled at 10x magnification and resized to 256×256 pixels.

\noindent\textbf{II. Pan-cancer Classification Tasks}\\
\noindent\textit{13. Pan-cancer Classification (17-Subtype)}. The pan-cancer classification task aims to enhance the model's generalization capability across various cancer types through large-scale and diverse data training. This approach not only helps improve the model's accuracy and robustness in identifying different cancers but also supports broader clinical applications. In the pan-cancer 17-classification task, we used a total of 1,071 WSIs samples from different organs and tissues, including 79 from the lung, 75 from the kidney, 71 from bone tissue, 70 each from the breast, gallbladder, appendage, parotid gland, spleen, thyroid, prostate, esophagus, and testis, 69 from cartilage, 53 from the placenta, 37 from the conjunctiva, 30 from the liver, and 27 from the submandibular gland. We employed ABMIL as the model framework, with training parameters set to 50 epochs, a batch size of 1, and a learning rate of 0.00002. The dataset was divided into training, validation, and test sets in an 8:1:1 ratio. All training operations were conducted on systems equipped with A100 GPUs. The images were sampled at a 20x magnification, with each image sized at 256$\times$256 pixels.

\noindent\textit{14. FFPE Pan-cancer Classification (32-Subtype)}. We selected 32 diseases from TCGA, involving 3,036 FFPE WSIs, for the pan-cancer 16-class classification task. Specifically, these include 100 WSIs each from the CESC, SARC, KIRP, THCA, THYM, PCPG, SKCM, PRAD, LUSC, READ, ACC, KIRC, LIHC, TGCT, LGG, and 95 WSIs each from the STAD, 91 WSIs from USC, 87 WSIs from MESO, 80 WSIs from UVM, 44 WSIs from DLBC, 39 WSIs from CHOL. We used ABMIL as the model framework with training parameters set to 50 epochs, a batch size of 1, and a learning rate of 0.00002. The dataset was divided into training, validation, and test sets in an 8:1:1 ratio. All training operations were conducted on systems equipped with A100 GPUs. The images were sampled at a magnification of 20x, with each image having a size of 256$\times$256 pixels.

\noindent\textit{15. Frozen Pan-cancer Classification (32-Subtype)}. We selected 32 diseases from TCGA, involving 3,038 FFPE WSIs, for the pan-cancer 16-class classification task. Specifically, these include 100 WSIs each from the CESC, KICH, KIRP, THCA, PCPG, SKCM, PRAD, LUSC, READ, KIRC, LIHC, TGCT, SARC, STAD, BLCA, COAD, LUAD, OV, PAAD, ESCA, LGG, and 99 WSIs each from GBM and BRCA, 98 WSIs each from the HNSC and THYM, 97 WSIs from UCEC, 96 WSIs from the ACC, 88 WSIs from the MESO, 71 WSIs from the CHOL, 70 WSIs from the UVM, 63 WSIs from the UCS, and 59 WSIs from the DLBC. We employed ABMIL as the model framework, with training parameters set to 50 epochs, a batch size of 1, and a learning rate of 0.00002. The dataset was divided into training, validation, and test sets in an 8:1:1 ratio. All training operations were conducted on systems equipped with A100 GPUs. The images were sampled at a magnification of 20x, with each image sized at 256$\times$256 pixels.

\noindent\textbf{III. Multi-organ Lesion Identification and Analysis Tasks}\\

\noindent\textit{ 16. Breast Metastasis Detection (CAMELYON16)}. This task aims to detect breast cancer metastasis in lymph node WSI images. The dataset includes 402 WSIs, split into 244 training images, 28 validation images, and 130 test images. The ABMIL model was trained with 50 epochs, a batch size of 1, and a learning rate of 0.0002. Training was performed on NVIDIA GeForce RTX 3090 GPUs, with patches preprocessed at 10x magnification and resized to 224$\times$224 pixels.

\noindent\textit{17. Breast Metastasis Detection (CAMELYON17)}. This task is derived from the CAMELYON17 challenge, focusing on the detection of lymph node metastases in breast cancer patients. The dataset includes over 1,000 WSIs from five medical centers, with 500 WSIs used in this study, split into 400 training and 100 test images. Training was conducted on NVIDIA GeForce RTX 3090 GPUs, with patches processed at 20x magnification and resized to 224$\times$224 pixels.

\noindent\textit{18. Lymphoma PD-L1 Tumor Region Detection}. The goal of this task is to segment effective regions for quantitative PD-L1 assessment in lymphoma. The dataset includes 8,540 image data points, divided into 4,290 tumor region ROIs and 4,250 non-tumor region ROIs. The dataset was split into training and test sets, with 6,832 training samples and 1,708 test samples. Training parameters were set to 1,000 iterations and a batch size of 256. All training operations were conducted on A100 GPUs, with images sampled at 20x magnification and resized to 256$\times$256 pixels.

\noindent\textit{19. TIL Negative vs. Positive Classification}. The TCGA-TILs dataset is specifically designed for studying Tumor Infiltrating Lymphocytes (TILs), containing images with or without TILs. The original images are sourced from TCGA, and the dataset comprises a total of 304,097 image patches. All images are 100 x 100 pixels with a 20x resolution. If there are at least two TILs present in an image, it is labeled as TIL-positive. In the Tumor Infiltrating Lymphocyte (TIL) negative vs positive classification task based on the TCGA-TILs dataset, we processed a total of 8,540 image data, including 4,290 tumor region ROIs and 4,250 non-tumor region ROIs. The dataset is divided into training and test sets, with 6,832 training samples and 1,708 test samples. Training parameters are set to a maximum of 1,000 iterations and a batch size of 256. All training operations are conducted on systems equipped with A100 GPUs. The images are sampled at a magnification of 20x, with each image sized at 256 $\times$ 256 pixels.

\noindent\textit{20. Breast Cancer Classification (PCam)}. The PCam dataset is used for binary classification of metastatic cancer in breast tumor regions. The dataset includes 2,948,912 image patches, with 148,219 tumor region ROIs and 146,693 non-tumor region ROIs. The training set contained 262,144 samples, while the test set had 32,768 samples. Training parameters were set to 1,000 iterations and a batch size of 256. All training operations were conducted on A100 GPUs, with images sampled at 20x magnification and resized to 256$\times$256 pixels.

\noindent\textit{21. Ring Cell Classification (DigestPath)}. Signet ring cell carcinoma is a special type of cancer commonly associated with gastric, colorectal, and breast cancers. This carcinoma is typically more aggressive, prone to early dissemination to other organs, and likely to metastasize to lymph nodes. This task involves the binary classification of signet ring cell carcinoma in gastrointestinal tract sections into positive and negative categories. Accurate identification and classification of signet ring cell carcinoma can aid clinicians in early detection and diagnosis of this highly aggressive cancer, thereby facilitating the development of more effective treatment plans and improving patient survival rates. In the training set, there are 1,074 positive patches and 1,611 negative patches. In the test set, there are 717 positive patches and 1075 negative patches. ROIs are cropped at 10x resolution, with a size of 256$\times$256 pixels. We use Linear Probing as the training strategy, with training parameters set to a maximum of 1000 iterations. All training operations are conducted on systems equipped with NVIDIA Tesla V100 GPUs.

\noindent\textit{22. Multi-Class Cell Type Segmentation (SegPath: Multi-Class)}. We utilized the Mask2Former~\cite{cheng2022masked} model for ROI-level cell type segmentation, specifically fine-tuning Mask2Former with a ViT-Adapter for the SegPath-based cell segmentation task. The SegPath dataset is designed for semantic segmentation of H\&E-stained images, encompassing eight major cell types within tumor tissues. The dataset comprises 158,687 ROIs, each with an original size of 984$\times$984 pixels at 20x magnification. The cell types and their respective quantities are as follows: endothelium (10,647 ROIs), epithelium (26,509 ROIs), leukocyte (24,805 ROIs), lymphocyte (12,273 ROIs), myeloid cell (14,135 ROIs), plasma cell (13,231 ROIs), red blood cell (25,909 ROIs), and smooth muscle (31,178 ROIs). The dataset was split into training and test sets in a 9:1 ratio. All training operations were conducted on systems equipped with NVIDIA GeForce RTX 3090 GPUs, with training parameters set to a learning rate of 5$\times$10$^{-4}$ and a batch size of 16.

\noindent\textit{ 23. Gland Segmentation (GlaS)}. For the Gland Segmentation in Colon Histology Images Challenge, we employed the Mask2Former model for ROI-level segmentation, fine-tuned with a ViT-Adapter. The GlaS dataset includes 165 images from 16 H\&E-stained T3 or T4 stage colorectal adenocarcinoma tissue sections, focusing on the segmentation of glands. The task utilized 1,900 ROI images, each originally sized at 224$\times$224 pixels at 20x magnification. The dataset was divided into 988 training images and 912 test images. All training was conducted on systems equipped with NVIDIA GeForce RTX 3090 GPUs, with training parameters set to a learning rate of 5$\times$10$^{-4}$ and a batch size of 16.

\noindent\textit{ 24. Nuclei Segmentation (PanNuke)}. For the PanNuke dataset-based cell segmentation task, we used Mask2Former for ROI-level cell type segmentation, fine-tuning it with a ViT-Adapter. PanNuke is a semi-automatically generated instance segmentation and classification dataset for cell nuclei, covering 19 different tissue types with detailed labels for each cell nucleus, including an instance segmentation mask. We utilized 7,901 ROI images, each sized at 224x224 pixels, with the training set comprising 2,124 images and the test set 532 images. All training was conducted on systems equipped with NVIDIA GeForce RTX 3090 GPUs, with training parameters set to a learning rate of 5$\times$10$^{-4}$ and a batch size of 16.

\noindent\textit{25. Nuclear Segmentation in Colorectal Cancer (CoNSeP)}. In the colorectal nuclear segmentation task using the CoNSep dataset, we analyzed 41 H\&E-stained image tiles, each sized at 1,000$\times$1,000 pixels at 40x magnification. These images were extracted from 16 colorectal adenocarcinoma WSIs, each from different patients, and scanned using the Omnyx VL120 scanner. The dataset includes seven nuclei types: other, inflammatory, healthy epithelium, dysplastic/malignant epithelium, fibroblast, muscle, and endothelial cells, with pixel-level predictions derived from image-level annotations. The total dataset consists of 656 images, with 224 in the training set and 432 in the test set. Training parameters were set to 200 epochs, a batch size of 4, and a learning rate of 0.00005. All training was conducted on systems equipped with NVIDIA GeForce RTX 4090 GPUs. Images were sampled at 40x magnification, with each image sized at 224$\times$224 pixels.

\noindent\textit{26. Semantic Segmentation of Adenocarcinoma (COSAS)}. In the semantic segmentation task based on the COSAS dataset, we processed 190 patches scanned by three different instruments. The task involved adenocarcinoma classification at the patch level and segmentation into normal regions and adenocarcinoma at the pixel level. The dataset consisted of 190 ROI images, divided into training and test sets at a 4:1 ratio. Training parameters were set to a learning rate of 1$\times$10$^{-4}$, a batch size of 32, and 100 epochs. All training was conducted on systems equipped with NVIDIA GeForce RTX 4090 GPUs. Images were sampled at 10x magnification, with each image sized at 224$\times$224 pixels.

\noindent\textit{27. Cell Type Segmentation (COSAS)}. In this semantic segmentation task, we focused on cell type segmentation at both patch and pixel levels within the COSAS dataset. The task included three cancer types: gastric adenocarcinoma, colorectal adenocarcinoma, and pancreatic ductal adenocarcinoma, with pixel-level classification into four categories, including normal regions. The dataset comprised 180 ROI images, divided into training and test sets at a 4:1 ratio. Training parameters were set to a learning rate of 1$\times$10$^{-4}$, a batch size of 32, and 100 epochs. All training was conducted on systems equipped with NVIDIA GeForce RTX 4090 GPUs. Images were sampled at 5x magnification, with each image sized at 224$\times$224 pixels.

\noindent\textit{28. Mitosis Detection (SegPath: Single-Class)}. We utilized the Mask2Former~\cite{cheng2022masked} model for ROI-level mitosis detection, specifically fine-tuning Mask2Former with a ViT-Adapter to enhance performance for the SegPath-based mitosis detection task. The SegPath dataset is designed for semantic segmentation of H\&E-stained images, containing a large number of annotated mitotic cells within tumor tissues. The dataset comprises 158,687 ROIs, each with an original size of 984$\times$984 pixels at 20x magnification. The dataset was split into training and test sets in a 9:1 ratio. All training operations were conducted on systems equipped with NVIDIA GeForce RTX 3090 GPUs, with training parameters set to a learning rate of 5$\times$10$^{-4}$ and a batch size of 16.

\noindent\textit{29. TCT Tumor Lesion Detection}. This task involved detecting and classifying positive cells in cervical cell pathology ROI regions. The dataset comprised 1,378,941 ROI images, with 965,337 in the training set and 413,604 in the test set. ViTDet~\cite{li2022exploring} was used as the model framework, with training parameters set to a learning rate of 0.0001, a batch size of 32, and 2 epochs. All training was conducted on systems equipped with NVIDIA Tesla V100 GPUs. Images were sampled at 20x magnification, with each image sized at 512$\times$512 pixels.

\noindent\textit{30. PE\&A Neoplastic Cell Few-shot Detection}. This task involved detecting abnormal cells in pleural and peritoneal fluid pathology slides, processing a total of 36,544 ROI images. The dataset was split into training and test sets, with 32,281 training images and 4,263 test images. ViTDet was employed as the model framework, with training parameters set to a learning rate of 0.0001, a batch size of 32, and 6 epochs. All training was conducted on systems equipped with NVIDIA Tesla V100 GPUs. Images were sampled at 20x magnification, with each image sized at 512$\times$512 pixels.

\noindent\textbf{IV. Multi-cancer Subtype Classification Tasks}\\
\noindent\textit{31. Cervical Cancer Positive Cell Subtyping}. The task involves subtype classification of lesion cells in cervical pathology slides into four categories with ASC-US, LSIL, ASC-H, and HSIL. The dataset contains a total of 215,296 lesion cell images, with 149,859 used for training, none for validation, and 65,437 for testing. The model was trained using an MLP architecture with a learning rate of 0.0002, batch size of 256, and 6 epochs. All training was conducted on NVIDIA Tesla V100 GPUs. The images were processed at 20x magnification and resized to 224$\times$224 pixels.

\noindent\textit{32. Bladder Benign-Malignant Classification}. The significance of bladder cancer screening lies in the early detection and diagnosis of bladder cancer, thereby improving treatment outcomes and reducing patient mortality. Timely screening facilitates the early implementation of effective treatment plans, improving patient prognosis and quality of life. This task involved binary classification of bladder tissue sections into cancerous and benign categories. The experiment included 235 cancerous sections and 89 benign sections. ABMIL was used for model training, with training parameters set to 50 epochs, a learning rate of 0.00002, and a batch size of 1. ROIs were cropped at 10x magnification, with a size of 256$\times$256 pixels. All training was conducted on systems equipped with NVIDIA Tesla V100 GPUs.

\noindent\textit{33. TCT Benign-Malignant Classification (WSI)}. The task focuses on slice-level diagnosis of cervical cell pathology slides to determine whether they exhibit tumor lesions, classified into two categories: negative and positive. The dataset comprises a total of 1,497 WSIs, with 1,123 used for training, none for validation, and 374 for testing. The model was trained using the ABMIL algorithm with a learning rate of 0.0002, batch size of 1, and 10 epochs. All training was conducted on NVIDIA Tesla V100 GPUs. The images were processed at 20x magnification and resized to 224$\times$224 pixels.

\noindent\textit{34. Bladder Benign-Malignant Classification (Patch)}. This task involved binary classification of malignant and benign patch images extracted from bladder biopsy slides. The dataset included 235 cancer slides and 89 benign slides. The WSI data was split into training and test sets in a 3:2 ratio, followed by patch extraction at 10x magnification. The positive training set comprised 3,512 patches, and the positive test set comprised 2,082 patches. Linear Probing was used as the training strategy, with training parameters set to a maximum of 1,000 iterations. All training was conducted on systems equipped with NVIDIA Tesla V100 GPUs.

\noindent\textit{35. TCT Benign-Malignant Classification (Patch)}. In this task, we aim to determine whether ROIs (Regions of Interest) from cervical cell pathology slides contain abnormal cells. The dataset comprises a total of 1,378,941 ROI images, including both diseased and non-diseased samples. The data is divided into training and test sets, with the training set containing 149,859 ROI images and the test set comprising 65,437 ROI images. For training, we employed a linear probing approach with parameters set to a learning rate of 0.0001, a batch size of 256, and a total of 2 epochs. All training operations were conducted on systems equipped with NVIDIA Tesla V100 GPUs. The images were processed at 20x magnification and resized to 224$\times$224 pixels.

\noindent\textit{36. Digestive Tract Benign-Malignant Classification}. The primary objective of this task is to facilitate the early detection of gastrointestinal cancers (such as gastric and colorectal cancers) and other gastrointestinal diseases (including polyps and ulcers) by classifying tissue sections as either malignant or benign. Specifically, the dataset includes 330 malignant sections (110 cancerous, 110 high-grade, and 110 low-grade) and 495 benign sections. We utilized ABMIL for model training, with training parameters set to 50 epochs, a learning rate of 0.00002, and a batch size of 1. ROIs were cropped at 10x magnification, each resized to 256$\times$256 pixels. All training operations were conducted on systems equipped with NVIDIA Tesla V100 GPUs.

\noindent\textit{37. Lymphoma vs Reactive Hyperplasia Classification}. Differentiating between lymphoma and reactive hyperplasia is crucial for precise treatment, as these conditions often share similar morphological features, leading to potential misdiagnoses. The use of artificial intelligence in this context can significantly enhance diagnostic accuracy and consistency, thereby optimizing patient treatment strategies. This task involved WSI-level binary classification to distinguish between lymphoma and reactive hyperplasia, utilizing a total of 688 H\&E-stained WSIs, including 368 cases of reactive hyperplasia and 320 cases of lymphoma. ABMIL was employed as the model framework, with training parameters set to 50 epochs, a batch size of 1, and a learning rate of 0.00002. The dataset was divided into training, validation, and test sets in an 8:1:1 ratio. All training operations were conducted on systems equipped with A100 GPUs, with images sampled at 20x magnification and resized to 256$\times$256 pixels.

\noindent\textit{38. Lymphoma B/T Cells Classification}. B-cell and T-cell lymphomas exhibit significant differences in their response to targeted therapies and prognoses. While traditional diagnostic methods rely heavily on immunohistochemical analysis, leveraging deep learning to identify the cell of origin from H\&E-stained images offers substantial clinical benefits. For this study, we selected a dataset of 645 H\&E-stained WSIs, including 354 B-cell lymphoma images and 291 T-cell lymphoma images. The ABMIL framework was utilized, with training parameters set to 50 epochs, a batch size of 1, and a learning rate of 0.00002. The dataset was divided into training, validation, and test sets in an 8:1:1 ratio. All training procedures were conducted on systems equipped with A100 GPUs, with image magnification set at 20x and resized to 256$\times$256 pixels.

\noindent\textit{39. Intestinal Cancer Detection}. The early detection of colorectal cancer is of critical importance as it can substantially improve treatment outcomes and survival rates. This task aims to classify colorectal tissue sections into cancerous and non-cancerous (benign) categories, thereby aiding in early diagnosis and treatment, which ultimately helps reduce mortality rates. The dataset includes 110 cancerous sections and 165 benign sections. We utilized ABMIL for model training, with training parameters set to 50 epochs, a learning rate of 0.00002, and a batch size of 1. ROIs were cropped at 10x magnification and resized to 256$\times$256 pixels. All training operations were conducted on systems equipped with NVIDIA Tesla V100 GPUs.

\noindent\textit{40. Cervical Cell Classification (TissueNet)}. This task focuses on the detection of cervical precancerous lesions using a dataset comprising thousands of cervical tissue microscope slides collected. Each image is classified based on the most severe epithelial lesion present. The categories include benign (normal or subnormal), low malignant potential (low-grade squamous intraepithelial lesion), high malignant potential (high-grade squamous intraepithelial lesion), and invasive cancer (invasive squamous cell carcinoma). The dataset includes 1,005 WSIs from four different scanning formats, with 804 WSIs in the training set and 201 WSIs in the test set. All training operations were conducted on systems equipped with NVIDIA GeForce RTX 3090 GPUs. The image patches were preprocessed at 20x magnification and cropped to 224$\times$224 pixels.

\noindent\textit{41. Colon and Lung Tissues Classification (LC25K-5)}. The LC25K dataset comprises 25,000 color images categorized into five classes, each containing 5,000 images. The classes include colorectal adenocarcinoma, benign colorectal tissue, lung adenocarcinoma, lung squamous cell carcinoma, and benign lung tissue. Each image is sized at 768×768 pixels. For this task, the dataset was split into a training set of 17,500 images and a test set of 7,500 images. The training parameters were set to a maximum of 1,000 iterations with a batch size of 256. All training operations were conducted on a system equipped with an A100 GPU. Images were downsampled to 256$\times$256 pixels at 20x magnification.

\noindent\textit{42. Lung subtyping. (LC25K-3)}. The LC25K dataset contains 25,000 color images, divided into five categories, each with 5,000 images. To differentiate between lung adenocarcinoma, lung squamous cell carcinoma, and benign lung tissue, we constructed a subset containing 15,000 images, with 5,000 images per category. All images are sized at 768$\times$768 pixels. In the experiment, the dataset was divided into training and testing sets, with the training set comprising 10,498 images and the testing set containing 4,502 images. The training parameters were set to a maximum of 1,000 iterations with a batch size of 256. All training operations were conducted on systems equipped with A100 GPUs. The images were downsampled to a resolution of 20x, with a size of 256$\times$256 pixels.

\noindent\textit{43. Colon Subtyping (LC25K-2)}. Within the LC25K dataset, a subset of 10,000 images was used to distinguish between colon cancer and benign colon tissue. Each image was originally sized at 768$\times$768 pixels. The dataset was split into a training set of 7,200 images and a test set of 2,998 images. The training parameters were set to a maximum of 1,000 iterations with a batch size of 256. All training operations were conducted on systems equipped with A100 GPUs. Images were downsampled to 256$\times$256 pixels at 20x magnification.

\noindent\textit{44. Breast Benign-Malignant Classification (BreaKHis)}. This task focuses on classifying histopathological images from breast tumor surgical specimens to identify benign and malignant breast tumors. Benign categories include adenosis (A), fibroadenoma (F), phyllodes tumor (PT), and tubular adenoma (TA); malignant categories include ductal carcinoma (DC), lobular carcinoma (LC), mucinous carcinoma (MC), and papillary carcinoma (PC). The dataset comprises 1,693 images at 400x magnification, with 547 benign and 1,146 malignant images. The dataset was divided into 1,148 training images and 545 test images. The maximum number of iterations for linear probing was set to 1,000, with a batch size of 256 images. All training operations were conducted on systems equipped with A100 GPUs to ensure efficient processing. All images were resized to 256 pixels to meet model input requirements.

\noindent\textit{45. NSCLC Subtyping (TCGA+)}. The TCGA NSCLC subtyping task aims to classify non-small cell lung cancer (NSCLC) into its two major subtypes using H\&E-stained histopathological images. The subtypes include lung adenocarcinoma (LUAD), which originates from alveolar epithelial cells and is characterized by glandular structures and mucus production, and lung squamous cell carcinoma (LUSC), originating from bronchial epithelial cells and characterized by squamous cell proliferation and keratinization. The dataset consists of 922 WSIs, divided into 736 for the training set, 89 for the validation set, and 97 for the test set. The ABMIL model was trained with 50 epochs, a batch size of 1 image, and a learning rate of 0.0002. All training operations were conducted on systems equipped with NVIDIA GeForce RTX 3090 GPUs. Image patches were preprocessed at 10x magnification and resized to 224$\times$224 pixels to meet model input requirements.

\noindent\textit{46. RCC subtyping (TCGA+)}. The TCGA RCC subtyping task focuses on the classification of Renal Cell Carcinoma (RCC) into its three major subtypes using H\&E-stained histopathological images. These subtypes include Clear Cell Renal Cell Carcinoma (ccRCC), the most common RCC subtype characterized by tumor cells with clear cytoplasm; Papillary Renal Cell Carcinoma (pRCC), identified by its papillary structures; and Chromophobe Renal Cell Carcinoma (chRCC), which features tumor cells with pale or clear cytoplasm and distinct nuclear chromatin patterns. We utilized a total of 1,043 WSIs for this task, divided into 848 for the training set, 97 for the validation set, and 98 for the test set. The ABMIL model was employed for training, with parameters set to 50 epochs, a batch size of 1 image, and a learning rate of 0.0002. All training operations were conducted on systems equipped with NVIDIA GeForce RTX 3090 GPUs. To meet the model's input requirements, all image patches were preprocessed at 10x magnification and resized to 224$\times$224 pixels.

\noindent\textit{47. Lymphoma Subtype Multi-class Classification}. Diagnosing lymphoma subtypes presents significant challenges due to morphological similarities, diverse immunophenotypes, and the necessity to integrate clinical and molecular data for comprehensive analysis. These factors place high demands on the expertise of pathologists. Artificial intelligence has greatly enhanced the efficiency and accuracy of diagnosing lymphoma subtypes (e.g., Follicular Lymphoma (FL), Angioimmunoblastic T-cell Lymphoma (AITL), Diffuse Large B-Cell Lymphoma (DLBCL), Natural Killer/T-cell Lymphoma (NKT)), enabling a more precise integration of various information sources for subtype identification. This, in turn, supports the development of personalized treatment plans and improves prognosis evaluation. In this study, we utilized 363 H\&E-stained WSIs, comprising 109 FL, 77 AITL, 75 DLBCL, 59 NKT, and 43 reactive hyperplasia cases. The ABMIL framework was used for training, with parameters set to 50 epochs, a batch size of 1, and a learning rate of 0.00002. The dataset was divided into training, validation, and test sets in an 8:1:1 ratio. All training operations were conducted on systems equipped with A100 GPUs, with image magnification set at 20x and resized to 256$\times$256 pixels.

\noindent\textit{48. BRCA Subtyping (BACH)}. The BRCA subtyping task involves the classification of breast cancer pathology images into four primary cancer types using H\&E-stained histological images. These images are derived from the ICIAR 2018 Breast Cancer Histology Image Grand Challenge, annotated and extracted from H\&E FFPE diagnostic histopathology WSIs. The data are categorized into four types: normal (100 images), benign (100 images), in situ carcinoma (100 images), and invasive carcinoma (100 images). The dataset consists of 400 ROI images, with 280 images in the training set and 120 images in the test set. All training operations were conducted on systems equipped with NVIDIA GeForce RTX 3090 GPUs. Image patches were sampled at 20x magnification (0.42 mpp) and had an original size of 2048$\times$1536 pixels. To meet the model's input requirements, these patches were resized as needed for the training process.

\noindent\textit{ 49. ESCA Subtyping (TCGA+)}. The ESCA subtyping task involves using image retrieval techniques to classify esophageal cancer subtypes. This task utilizes datasets from UKK, WNS, and TCGA. The query images are encoded into low-dimensional feature representations by an encoder and then compared with other images using a K-Nearest Neighbors (KNN) algorithm. The dataset includes 367,330 ROI images, extracted using QuPath, each sized at 256x256 pixels with a 20x magnification. The dataset was divided into 189,143 ROI images for training and 178,187 ROI images for testing. Evaluation metrics include Acc@K (where K = 1, 3, 5), representing the accuracy of retrieving images with the same label as the query image among the top K results, and MVAcc@5, which is considered correct if the majority vote of the labels of the 5 retrieved images matches the query image's label. All training operations were conducted on systems equipped with NVIDIA GeForce RTX 3090 GPUs.

\noindent\textit{50. CRC Screening (HunCRC)}. The CRC screening task is designed to identify and classify four categories associated with colorectal cancer (CRC) from WSI images. These categories include: Normal Tissue, Low-Grade Adenoma, High-Grade Adenoma, and Colorectal Adenocarcinoma. We utilized a total of 200 WSIs, divided into 160 WSIs for training, 20 for validation, and 20 for testing. The ABMIL model was employed for training, with parameters set to 50 epochs, a batch size of 1 image, and a learning rate of 0.0002. All patches were preprocessed at 10x magnification and resized to 224$\times$224 pixels to meet the model's input requirements. Training was conducted on systems equipped with NVIDIA GeForce RTX 3090 GPUs.

\noindent\textit{51. Prostate ISUP Grading (PANDA)}. The PANDA classification task involves grading prostate cancer using core needle biopsies, originating from the PANDA challenge dataset, which includes 10,616 WSIs from Radboud University Medical Center and Karolinska Institute. The dataset was divided into training and test sets in an 8:2 ratio, with 8,494 WSIs in the training set and 2,124 WSIs in the test set. The ABMIL model was used for training, with parameters set to 50 epochs, a learning rate of 0.0002, and a batch size of 1. Each WSI was sampled at 10x magnification and resized to 256$\times$256 pixels. All training operations were conducted on systems equipped with NVIDIA GeForce RTX 4090 GPUs.

\noindent\textit{52. Prostate Gleason Grading (in-house)}. The prostate Gleason grading task uses a private dataset focused on multi-label classification of prostate Gleason scores using H\&E-stained images. This dataset contains 1,042 WSIs categorized into four classes: negative, grade 3, grade 4, and grade 5. For example, a score of 4+3 indicates a primary grade of 4 with a secondary grade of 3. The dataset was split into training, validation, and test sets in a 7:2:1 ratio, comprising 737 WSIs in the training set, 200 in the validation set, and 105 in the test set. All images were sampled at 20x magnification and resized to 224$\times$224 pixels. Training was conducted over 40 epochs with a learning rate of 0.0002 and a batch size of 32, using a single NVIDIA RTX 4090 GPU.

\noindent\textit{53. Intestinal Cancer High-Grade vs. Low-Grade Classification}. Intestinal intraepithelial neoplasia can be classified into high-grade and low-grade based on the morphological characteristics and structural abnormalities of the diseased cells. This task aims to perform binary classification of intestinal intraepithelial neoplasia into high-grade and low-grade categories based on morphological characteristics and structural abnormalities. The dataset includes 110 cases for each category. We utilized the ABMIL model for training, with parameters set to 50 epochs, a learning rate of 0.00002, and a batch size of 1. ROIs were cropped at 10x magnification, resized to 256$\times$256 pixels, and training operations were conducted on systems equipped with NVIDIA Tesla V100 GPUs.

\noindent\textit{54. Classification of Tubular Adenocarcinoma Types (PatchGastricADC22 Subtype-2)}. In this sub-task based on the PatchGastricADC22 dataset, we focused on the classification of moderately differentiated tubular adenocarcinoma and well-differentiated tubular adenocarcinoma using histopathological images and related diagnostic reports. The dataset comprises 133,826 image patches, with 73,336 patches for moderately differentiated tubular adenocarcinoma and 60,490 patches for well-differentiated tubular adenocarcinoma. The data were divided into a training set of 107,060 patches and a test set of 26,766 patches. Training parameters were set to a learning rate of 0.0001, a batch size of 64, and 50 epochs. All training operations were conducted on systems equipped with NVIDIA Tesla A100 GPUs, with images sampled at 20x magnification and sized at 224$\times$224 pixels.

\noindent\textit{55. Subtype Classification of Gastric Adenocarcinoma (PatchGastricADC22 Subtype-10)}. In the task based on the PatchGastricADC22 dataset, we used a dataset containing histopathological images and related diagnostic reports of gastric adenocarcinoma endoscopic biopsy specimens. This dataset includes various subtypes of gastric adenocarcinoma and comprises a total of 262,777 image patches. Specific ROI categories include well-differentiated tubular adenocarcinoma, moderately differentiated tubular adenocarcinoma, papillary adenocarcinoma, moderately to poorly differentiated adenocarcinoma, poorly differentiated adenocarcinoma (non-solid type), poorly differentiated adenocarcinoma (solid type), well to moderately differentiated tubular adenocarcinoma, signet ring cell carcinoma, mucinous adenocarcinoma, and others. The dataset consists of 262,777 ROIs, divided into training and test sets in a 4:1 ratio. The training parameters are set to a learning rate of 0.0001, a batch size of 64, and 50 epochs. All training operations are conducted on systems equipped with NVIDIA GeForce RTX 3090 GPUs. The images are sampled at 20X magnification, with each image having a size of 224$\times$224 pixels.

\noindent\textit{56. PRAD Tissue Classification (AGGC)}. The PRAD tissue classification task involves using the AGGC dataset to retrieve images based on low-dimensional features extracted by an encoder, comparing them with features from other images using a KNN algorithm. This dataset is from the Automated Gleason Grading Challenge 2022 and is annotated for prostate cancer Gleason grading. A total of 1,126,447 ROI images were used for this task, with the training set containing 780,933 ROIs and the test set containing 345,514 ROIs. Evaluation metrics include Acc@K (where K = 1, 3, 5), representing the accuracy of retrieving images with the same label as the query image among the top K results, and MVAcc@5, which is considered correct if the majority vote of the labels of the 5 retrieved images matches the query image's label. All training operations were conducted on systems equipped with NVIDIA GeForce RTX 3090 GPUs, with patches preprocessed at 20x magnification and cropped to 256$\times$256 pixels.

\noindent\textit{57. Glioma IDH1 Screening (TCGA+)}. The Glioma IDH1 screening task focuses on the detection of isocitrate dehydrogenase 1 (IDH1) mutations in gliomas. This task supports binary classification to distinguish between IDH1-mutant and IDH1-wildtype gliomas. IDH1-mutant gliomas, characterized by the presence of IDH1 gene mutations, generally have different prognoses and treatment responses compared to IDH1-wildtype gliomas, which lack these mutations and tend to be more aggressive. A total of 1,698 WSIs were utilized, with 1,358 allocated for the training set, 170 for the validation set, and 170 for the test set. The ABMIL model was trained for 50 epochs with a batch size of 1 image and a learning rate of 0.0002. All patches were preprocessed at 20x magnification and resized to 256$\times$256 pixels. The training was conducted on NVIDIA GeForce RTX 3090 GPUs.

\noindent\textit{58. Pan-Organ Lymph Node Metastasis Classification}. This task involves a binary classification aimed at detecting malignancy in patches extracted from multi-organ lymph node metastasis slides. Patch images were categorized as either cancerous or benign. The original dataset comprised 219 cancerous WSIs and 703 benign WSIs, which were divided into training and test sets in a 3:2 ratio. Patches were extracted at 10x magnification based on regions marked by pathologists, with the number of patches per WSI varying according to the positive and negative regions identified. Each patch was originally sized at 512$\times$512 pixels and resized to 256$\times$256 pixels for model input. The positive training set included 6,589 patches, while the positive test set had 10,020 patches. The negative training and test sets contained 19,256 and 12,075 patches, respectively. Linear Probing was used as the training strategy, with a maximum of 1,000 iterations. Training was conducted on NVIDIA Tesla V100 GPUs.

\noindent\textit{59. Intestinal Cancer and Polyp Type Classification}. This task classifies benign intestinal sections into four categories including negative, hyperplastic polyps, inflammatory polyps, and polypoid hyperplasia. Each category consisted of 100 examples. The ABMIL model was employed for training, with parameters set to 50 epochs, a learning rate of 0.00002, and a batch size of 1. ROIs were cropped at 10x magnification and resized to 256$\times$256 pixels. Training was conducted on NVIDIA Tesla V100 GPUs.

\noindent\textit{60. CRC Tissue Classification (HunCRC)}. The CRC tissue classification task is based on patch images from colorectal cancer (CRC) patients, aiming to automatically identify nine distinct categories associated with CRC: adenocarcinoma, high-grade dysplasia, low-grade dysplasia, inflammation, tumor necrosis, suspicious infiltration, resection margin, technical artifacts, and normal tissue. The dataset included 101,398 patch images, with 70,978 in the training set and 30,420 in the test set. All patches were sampled at 20x magnification (0.48 mpp) and originally sized at 512$\times$512 pixels. Training was conducted on NVIDIA GeForce RTX 3090 GPUs.

\noindent\textit{61. RCC Tissue Classification (TCGA+)}. The RCC tissue classification task focused on clear cell renal cell carcinoma (CCRCC) and utilized 52,713 ROI images ranging from 256$\times$256 to 300$\times$300 pixels at approximately 0.25 mpp. These images were annotated and extracted from H\&E FFPE diagnostic histopathology WSIs of CCRCC samples from TCGA (502 samples) and Helsinki University Hospital (HEL) (64 samples). The images were categorized into six classes: cancer (13,057 ROIs), normal (8,652 ROIs), stroma (5,460 ROIs), red blood cells (996 ROIs), empty background (16,026 ROIs), and other textures (8,522 ROIs). Due to label imbalance and ambiguity in the other category, only cancer, normal, and stroma labels were considered. The dataset comprised 21,095 ROIs from TCGA and 6,074 ROIs from HEL, with 19,018 images used for training and 8,150 for testing. Linear Probing was employed for model training, conducted on NVIDIA GeForce RTX 3090 GPUs.

\noindent\textit{62. CRC Tissue Classification (CRC-100K)}. This task utilized the NCT-CRC-HE-100K dataset, consisting of 100,000 non-overlapping histological image patches from human colorectal cancer (CRC) and normal tissues, all stained with H\&E and sized at 224$\times$224 pixels with a resolution of 20x. The dataset is derived from 86 H\&E-stained human cancer tissue slides, including primary CRC tumor sections and CRC liver metastases. Normal tissue categories were augmented with non-tumor regions from gastrectomy specimens to increase variability. The test set was derived from the CRC-VAL-HE-7K dataset, which included 7,180 image patches from 50 colorectal adenocarcinoma patients, with images of the same size and resolution. The task involved nine tissue categories: adipose tissue (ADI), background (BACK), debris (DEB), lymphocytes (LYM), mucus (MUC), smooth muscle (MUS), normal colon mucosa (NORM), cancer-associated stroma (STR), and colorectal adenocarcinoma epithelium (TUM). Training parameters were set with a maximum of 1,000 iterations and a batch size of 256. Training was conducted on A100 GPUs, with images sampled at 20x magnification and resized to 256$\times$256 pixels.

\noindent\textit{63. Colon Subtyping (Chaoyang)}. The Chaoyang dataset includes colon tissue slice images, encompassing four categories: normal, serrated, adenocarcinoma, and adenoma. A total of 6,160 samples were used to evaluate the model's performance on this task, with each image patch sized at 512$\times$512 pixels. The original WSI images were scanned at 20x magnification. The dataset was divided into a training set with 4,021 images and a test set with 2,139 images. Training parameters included a maximum of 1,000 iterations and a batch size of 256. Training was conducted on A100 GPUs.

\noindent\textit{64. Pixel-Level Tissue Type Prediction in LUAD (WSSS4LUAD)}. In the WSSS4LUAD dataset task, we used image-level annotations to achieve pixel-level predictions for three critical tissue types including tumor epithelium, tumor-associated stroma, and normal tissue. The dataset comprised 1,176 ROI images, with 1,129 allocated for the training set and 47 for the test set. The training parameters were set to a learning rate of 5.00E-05, a batch size of 4, and 200 epochs. Training was conducted on NVIDIA GeForce RTX 4090 GPUs, with each image sized at 224$\times$224 pixels.

\noindent\textit{65. Colorectal Cancer Segmentation (Kather)}. This task aimed to achieve 8-class classification of colorectal cancer (CRC) using the Kather pathology dataset, a publicly available resource widely used for CRC research. The dataset contains 5,000 H\&E-stained histopathological images of human colorectal cancer, representing eight different tissue types: tumor epithelium, simple stroma, complex stroma, immune cells, debris, normal mucosal glands, adipose tissue, and background. The images were manually annotated and sliced at 20x magnification, with each ROI sized at 150$\times$150 pixels. The training set consisted of 3,500 images, and the test set included 1,500 images. Training was conducted on NVIDIA GeForce RTX 3090 GPUs.

\noindent\textit{66. Brain Subtype Classification (Ebrains)}. The Ebrains task using the subset of the work~\cite{roetzer2022digital} involved multi-task learning classification using the ABMIL model after feature extraction. The dataset comprised 963 WSIs, covering 30 classification categories (Anaplastic astrocytoma with IDH1-mutant,
Anaplastic astrocytoma with IDH1-wildtype,
Anaplastic oligodendroglioma with IDH1-mutant \& 1p/19q codeleted,
Glioblastoma with IDH1-mutant,
Glioblastoma with IDH1-wildtype,
Gliosarcoma,
Oligodendroglioma with IDH1-mutant \& 1p/19q codeleted,
Pilocytic astrocytoma,
Schwannoma,
Medulloblastoma with non-WNT/non-SHH,
Anaplastic ependymoma,
Ependymoma,
Ganglioglioma,
Diffuse large B-cell lymphoma of the CNS,
Langerhans cell histiocytosis,
Anaplastic meningioma,
Angiomatous meningioma,
Atypical meningioma,
Fibrous meningioma,
Meningothelial meningioma,
Secretory meningioma,
Transitional meningioma,
Haemangioblastoma,
Haemangioma,
Haemangiopericytoma,
Lipoma,
Metastatic tumours,
Diffuse astrocytoma with IDH1-mutant,
Adamantinomatous craniopharyngioma, 
and Pituitary adenoma). It was divided into a training set with 642 WSIs and a test set with 321 WSIs. Training parameters included a learning rate of 1.00E-04, a batch size of 32, and 10 epochs. Training was conducted on NVIDIA GeForce RTX 4090 GPUs, with images sampled at 40x magnification and resized to 224x224 pixels.

\noindent\textbf{V. Biomarker Assessment Tasks}\\
\noindent\textit{67. Multi-IHC qualification}. IHC is a critical technique for detecting the expression of specific proteins or antigens in tissue sections, using antibodies tagged with dyes or fluorophores to visualize the distribution and intensity of target proteins within cells or tissues. In this task, we aimed to comprehensively assess the expression of 30 different IHC markers in lymphomas. A total of 6,045 WSI samples were collected, including 3,282 positive samples and 3,123 negative samples. A weakly supervised classification method using the ABMIL model framework was employed. Training parameters were set to 50 epochs, a batch size of 1, and a learning rate of 0.00002. The dataset was split into training, validation, and test sets in an 8:1:1 ratio. All training was conducted on A100 GPUs, with images sampled at 20x magnification and resized to 256$\times$256 pixels.

\noindent\textit{68. CD5 IHC Qualification}. CD5 is a crucial T-cell surface antigen with significant clinical relevance in the classification and diagnosis of both B-cell and T-cell lymphomas. Notably, while CD5 is typically absent in most cases of diffuse large B-cell lymphoma, its positive expression can indicate a more aggressive subtype with a poorer prognosis. CD5 positivity is generally observed as distinct staining along the tumor cell membrane. Qualitative analysis of CD5 expression involves assessing both the intensity and distribution pattern of immunohistochemical staining. Positive expression is typically marked by brown staining, with the evaluation considering the intensity and proportion of positively stained tumor cells. This task aims to qualitatively analyze CD5 expression using 448 WSI samples. A weakly supervised classification approach was employed, using ABMIL as the model framework. Training parameters were set to 50 epochs, with a batch size of 1 and a learning rate of 0.00002. The dataset was split into training, validation, and test sets in an 8:1:1 ratio. All training operations were conducted on A100 GPU systems, with images magnified at 20x and resized to 256$\times$256 pixels.

\noindent\textit{69. CD3 IHC Qualification}. CD3 is a specific surface antigen of T cells and serves as an essential marker for assessing T-cell components in lymphomas. Its expression is critical for diagnosing and differentiating T-cell lymphomas from B-cell lymphomas and other non-lymphomatous lesions. CD3 positivity usually manifests as brown staining and the assessment considers both staining intensity and the proportion of positive cells. A higher staining intensity and a larger proportion of positive cells typically indicate a more mature T-cell component. This task involves a qualitative analysis of CD3 expression across 474 WSI samples. A weakly supervised classification method was employed using ABMIL as the model framework. Training parameters included 50 epochs, a batch size of 1, and a learning rate of 0.00002. The dataset was divided into training, validation, and test sets in an 8:1:1 ratio. All training operations were conducted on A100 GPU systems, with images sampled at 20x magnification and resized to 256$\times$256 pixels.

\noindent\textit{70. CD20 IHC Qualification}. CD20 is a specific antigen expressed on the surface of B cells and plays a crucial role in the assessment and diagnosis of B-cell lymphomas. CD20 is strongly positive in most B-cell lymphomas, aiding in confirming B-cell origin and supporting the diagnostic process. When combined with clinical context and other immunomarkers, CD20 expression provides significant insights for final diagnosis and treatment planning. CD20 positivity typically presents as distinct brown staining on the tumor cell membrane, with assessment focusing on staining intensity and the proportion of positive tumor cells. High staining intensity and a large proportion of positive cells usually suggest a more mature B-cell component, while CD20-negative or weak expression may indicate an atypical or low-expression B-cell lymphoma subtype or residual or recurrent disease post-treatment. This task aims to perform a qualitative analysis of CD20 expression using a dataset of 473 WSI samples. A weakly supervised classification approach was employed with ABMIL as the model framework. Training parameters were set to 50 epochs, with a batch size of 1 and a learning rate of 0.00002. The dataset was split into training, validation, and test sets in an 8:1:1 ratio. All training operations were conducted on A100 GPU systems, with images sampled at 20x magnification and resized to 256$\times$256 pixels.

\noindent\textit{71. CD79a IHC Qualification}. CD79a is a B-cell-specific antigen commonly co-expressed with CD20, serving as a critical marker in the diagnosis of B-cell lymphomas. CD79a is expressed across various stages of B-cell differentiation, making it clinically significant for detecting both early and mature B-cell lymphomas. Unlike CD20, CD79a can remain positive in certain B-cell lymphomas that lack CD20 expression, making it a valuable supplementary marker for confirming B-cell lineage. Given its broad and consistent expression in B-cell lymphomas, accurate qualitative evaluation of CD79a is crucial for lymphoma diagnosis, classification, and treatment planning. CD79a positivity typically presents as brown staining in the membrane of tumor cells. Strong positive expression suggests a clear B-cell origin, while negative or weak expression may necessitate further analysis with other B-cell markers like CD20. This task involves a qualitative analysis of CD79a expression using a dataset of 316 WSI samples. A weakly supervised classification approach was employed with ABMIL as the model framework. Training parameters were set to 50 epochs, with a batch size of 1 and a learning rate of 0.00002. The dataset was divided into training, validation, and test sets in an 8:1:1 ratio. All training operations were conducted on A100 GPU systems, with images sampled at 20x magnification and resized to 256$\times$256 pixels.

\noindent\textit{72. CD21 IHC Qualification}. CD21 is a specific antigen expressed on follicular dendritic cells (FDCs) and is essential for assessing follicular structures and the follicular dendritic cell network in lymphoid tissues. CD21 has unique clinical significance in lymphoma diagnosis, for example, follicular lymphoma typically exhibits CD21 positivity, indicating the preservation of the follicular dendritic cell network within the tumor, which aids in differentiating it from lymphomas like diffuse large B-cell lymphoma that lack distinct follicular structures. In the qualitative assessment of CD21 expression, pathologists focus on staining of the follicular dendritic cell network, which generally appears as brown staining on cell membranes within the follicular center or boundary, displaying a reticular or band-like pattern. This task involves qualitative analysis of CD21 expression using a dataset of 377 WSI samples. A weakly supervised classification approach was employed with ABMIL as the model framework. Training parameters were set to 50 epochs, with a batch size of 1 and a learning rate of 0.00002. The dataset was divided into training, validation, and test sets in an 8:1:1 ratio. All training operations were conducted on A100 GPU systems, with images sampled at 20x magnification and resized to 256$\times$256 pixels.

\noindent\textit{73. EBER Qualification}. EBER (Epstein-Barr virus-encoded RNA) is an RNA sequence encoded by the Epstein-Barr virus (EBV) and serves as a crucial marker for detecting EBV infection in tissues. In lymphoma, detecting EBER is particularly important for specific types of lymphomas such as nasal NK/T-cell lymphoma, Hodgkin lymphoma, and certain diffuse large B-cell lymphomas. EBER positivity indicates that tumor cells are associated with EBV infection. Given the high specificity of EBER expression, qualitative assessment is vital for diagnosing EBV-associated lymphomas, predicting prognosis, and guiding treatment strategies. EBER positivity typically manifests as distinct brown or purple speckled staining within the nuclei of tumor cells. When evaluating EBER expression and stained cell composition, it is important to consider staining intensity, the proportion of positive cells, and the distribution of these cells. Strong, widespread EBER expression often suggests a close association between the lymphoma and EBV infection, while EBER negativity may indicate that the tumor is not linked to EBV infection. This task involves qualitative analysis of EBER expression using a dataset of 325 WSI samples. A weakly supervised classification approach was employed, using ABMIL as the model framework. Training parameters were set to 50 epochs, a batch size of 1, and a learning rate of 0.00002. The dataset was divided into training, validation, and test sets in an 8:1:1 ratio. All training operations were conducted on A100 GPU systems, with images sampled at 20x magnification and resized to 256$\times$256 pixels.

\noindent\textit{74. CD10 IHC Qualification}. CD10 is a critical B-cell marker in lymphoma diagnosis, indicating that the tumor originates from germinal center B cells, which is essential for distinguishing these lymphomas from other B-cell or T-cell lymphomas. Some T-cell lymphomas may also express CD10. Qualitative assessment of CD10 expression typically reveals positive expression as brown staining on the tumor cell membrane, sometimes extending into the cytoplasm. Evaluation should consider staining intensity, the proportion of positive cells, and the uniformity of staining distribution. Strong positive expression and widespread distribution of positive cells suggest that the tumor has germinal center B-cell characteristics, while negative or uneven expression may necessitate further analysis using other B-cell markers and clinical information. This task involves qualitative analysis of CD10 expression using a dataset of 542 WSI samples. A weakly supervised classification method was employed with ABMIL as the model framework. Training parameters were set to 50 epochs, a batch size of 1, and a learning rate of 0.00002. The dataset was divided into training, validation, and test sets in an 8:1:1 ratio. All training operations were conducted on A100 GPU systems, with images magnified at 20x and resized to 256$\times$256 pixels.

\noindent\textit{75. Bcl-6 IHC Qualification}. Bcl-6 is a nuclear transcription factor predominantly expressed in germinal center B cells and serves as a critical marker in the evaluation and diagnosis of various B-cell lymphomas. And some T-cell lymphomas may also express Bcl-6. Positive Bcl-6 expression suggests that the tumor likely originates from germinal center B cells, making it highly relevant for diagnosing specific lymphoma subtypes. Bcl-6 expression also helps distinguish germinal center-derived B-cell lymphomas from other B-cell or T-cell lymphomas, providing essential insights for tumor classification and prognostic evaluation. Qualitative assessment of Bcl-6 expression focuses on nuclear staining of tumor cells. Uniform and strong nuclear positivity typically indicates germinal center B-cell characteristics, whereas weak or uneven staining may suggest atypical or heterogeneous lymphoma subtypes. This task involves qualitative analysis of Bcl-6 expression using a dataset of 527 WSI samples. A weakly supervised classification approach was employed with ABMIL as the model framework. Training parameters were set to 50 epochs, a batch size of 1, and a learning rate of 0.00002. The dataset was divided into training, validation, and test sets in an 8:1:1 ratio. All training operations were conducted on A100 GPU systems, with images magnified at 20x and resized to 256$\times$256 pixels.

\noindent\textit{76. Bcl-2 IHC Qualification}. Bcl-2 is an anti-apoptotic protein primarily located in the cytoplasm and plays a crucial role in the diagnosis, prognosis, and therapeutic decision-making for lymphomas. In follicular lymphoma, strong Bcl-2 positivity is a key diagnostic feature, contrasting with the physiological apoptosis inhibition seen in germinal center B cells, and often indicates the presence of FL. Additionally, in diffuse large B-cell lymphoma, Bcl-2 expression is a significant prognostic marker, with Bcl-2 positive DLBCL typically associated with poorer outcomes. Qualitative assessment of Bcl-2 expression focuses on the staining intensity and distribution within the cytoplasm and membrane of tumor cells. Strong positivity generally suggests that tumor cells have robust anti-apoptotic capabilities, which are linked to increased disease aggressiveness and potential treatment resistance. Conversely, negative or weak expression might indicate a less aggressive biological behavior of the tumor. This task involves qualitative analysis of Bcl-2 expression using a dataset of 471 WSI samples. A weakly supervised classification approach was employed with ABMIL as the model framework. Training parameters were set to 50 epochs, a batch size of 1, and a learning rate of 0.00002. The dataset was split into training, validation, and test sets in an 8:1:1 ratio. All training operations were conducted on A100 GPU systems, with images magnified at 20x and resized to 256$\times$256 pixels.

\noindent\textit{77. MUM-1 IHC Qualification}. MUM-1 (Multiple Myeloma Oncogene 1) is a nuclear transcription factor predominantly expressed in post-germinal center B cells, plasma cells, and certain T cells. Its accurate assessment plays a crucial role in the classification, diagnosis, and prognostic evaluation of lymphomas, particularly in identifying non-germinal center-type DLBCL and related subtypes. Qualitative evaluation of MUM-1 expression primarily focuses on the nuclear staining of tumor cells. Positive MUM-1 expression is typically characterized by brown nuclear staining. Strong positivity often indicates that the tumor cells are at a later stage of differentiation and is associated with a poorer prognosis. Conversely, negative or weak expression may suggest early B-cell differentiation or a lymphoma subtype not reliant on MUM-1 expression. This task involves qualitative analysis of MUM-1 expression using a dataset of 425 WSI samples. A weakly supervised classification approach was employed with ABMIL as the model framework. Training parameters were set to 50 epochs, a batch size of 1, and a learning rate of 0.00002. The dataset was divided into training, validation, and test sets in an 8:1:1 ratio. All training operations were conducted on A100 GPU systems, with images magnified at 20x and resized to 256$\times$256 pixels.

\noindent\textit{78. CD4 IHC Qualification}. CD4 is a glycoprotein predominantly expressed on the surface of helper T cells (Th cells), as well as some monocytes and macrophages. In lymphoma diagnosis, CD4 positivity suggests that the tumor cells may originate from helper T cells. Qualitative assessment of CD4 expression typically reveals positive staining as brown coloration on the cell membrane. The evaluation considers both staining intensity and the proportion of positive cells. Strong CD4 expression with a high proportion of positive cells often indicates a helper T cell phenotype. In contrast, negative or low CD4 expression may suggest that the tumor originates from other T cell subtypes or non-T cell lineages. This task involves qualitative analysis of CD4 expression using a dataset of 124 WSI samples. A weakly supervised classification approach was employed, using ABMIL as the model framework. Training parameters were set to 50 epochs, with a batch size of 1 and a learning rate of 0.00002. The dataset was divided into training, validation, and test sets in an 8:1:1 ratio. All training operations were conducted on A100 GPU systems, with images magnified at 20x and resized to 256$\times$256 pixels.

\noindent\textit{79. CD23 IHC Qualification}. CD23 is a low-affinity IgE receptor primarily expressed on the surface of mature B cells, follicular dendritic cells, and activated T cells. It plays a critical role in the diagnosis and classification of lymphomas, particularly in distinguishing chronic lymphocytic leukemia/small lymphocytic lymphoma (CLL/SLL) from other B-cell lymphomas. In the qualitative assessment of CD23 expression, pathologists focus on the staining patterns observed on the tumor cell membrane and cytoplasm. CD23 positivity typically presents as brown membrane staining, which may extend into the cytoplasm. High staining intensity and widespread distribution generally indicate that the tumor is likely CLL/SLL or a closely related B-cell lymphoma. Conversely, negative or weak expression suggests other types of B-cell lymphomas or non-B-cell-derived tumors. This study aims to qualitatively analyze CD23 expression using a dataset of 133 WSI samples. A weakly supervised classification approach was employed, utilizing ABMIL as the model framework. Training parameters were set to 50 epochs, with a batch size of 1 and a learning rate of 0.00002. The dataset was divided into training, validation, and test sets in an 8:1:1 ratio. All training operations were conducted on systems equipped with A100 GPUs, with images magnified at 20x and resized to 256$\times$256 pixels.

\noindent\textit{80. PD-1 IHC Qualification}. Programmed cell death protein 1 (PD-1) is an immune checkpoint molecule primarily expressed on the surface of activated T cells, B cells, and natural killer cells. Its expression is particularly significant in the diagnosis and classification of certain T-cell lymphomas and Hodgkin lymphoma. Positive PD-1 expression suggests that tumor cells may possess immune evasion mechanisms linked to an immunosuppressive tumor microenvironment. In evaluating PD-1 expression, pathologists focus on the staining patterns observed in tumor-infiltrating T cells or the tumor cell membrane. Typically, PD-1 positivity presents as brown membranous staining, with the intensity and proportion of positive cells being key factors in the assessment. A high proportion of strongly positive PD-1 staining often indicates potential responsiveness to PD-1/PD-L1 checkpoint inhibitors, making this information crucial for treatment decisions. Conversely, negative or low PD-1 expression may suggest a poor response to immunotherapy. This study aimed to qualitatively analyze PD-1 expression using 91 WSI samples. A weakly supervised classification approach was employed, utilizing ABMIL as the model framework. Training parameters were set to 50 epochs, with a batch size of 1 and a learning rate of 0.00002. The dataset was divided into training, validation, and test sets in an 8:1:1 ratio. All training operations were conducted on systems equipped with A100 GPUs, with images at 20x magnification and resized to 256$\times$256 pixels.

\noindent\textit{81. Cyclin D1 IHC Qualification}. Cyclin D1 is a key cell cycle regulatory protein that plays a crucial role in promoting the transition from the G1 phase to the S phase, thus driving cell proliferation. Its expression holds significant clinical relevance in lymphoma diagnosis, particularly in mantle cell lymphoma. Positive Cyclin D1 expression is typically characterized by intense brown nuclear staining in tumor cells. Strong and widespread Cyclin D1 expression strongly supports an MCL diagnosis, whereas negative or weak expression necessitates further analysis in conjunction with other clinical and pathological features. This study aimed to qualitatively analyze Cyclin D1 expression using 266 WSI samples. A weakly supervised classification approach was employed, utilizing ABMIL as the model framework. Training parameters were set to 50 epochs, with a batch size of 1 and a learning rate of 0.00002. The dataset was divided into training, validation, and test sets in an 8:1:1 ratio. All training operations were conducted on systems equipped with A100 GPUs, with images at 20x magnification and resized to 256$\times$256 pixels.

\noindent\textit{82. CD19 IHC Qualification}. CD19 is a transmembrane protein extensively expressed on the surface of B cells, playing a crucial role in the diagnosis and classification of lymphomas. The qualitative assessment of CD19 expression is vital for confirming the diagnosis of B-cell lymphomas and guiding treatment decisions. Pathologists primarily evaluate CD19 expression by observing the staining patterns on the tumor cell membrane. Given that CD19 is strongly positive in nearly all B-cell lymphomas, any loss or reduction in its expression may indicate certain atypical B-cell lymphoma subtypes or potential technical issues. This study focuses on the qualitative analysis of CD19 expression using 161 WSI samples. A weakly supervised classification approach was employed, utilizing ABMIL as the model framework. Training parameters were set to 50 epochs, with a batch size of 1 and a learning rate of 0.00002. The dataset was divided into training, validation, and test sets in an 8:1:1 ratio. All training operations were conducted on systems equipped with A100 GPUs, with images magnified at 20x and resized to 256$\times$256 pixels.

\noindent\textit{83. CD22 IHC Qualification}. CD22 is a transmembrane glycoprotein expressed on the surface of B cells and plays a crucial role in the diagnosis and classification of lymphomas by confirming the B-cell origin of B-cell lymphomas. Typically, CD22 is used alongside other B-cell markers such as CD19, CD20, and CD79a to distinguish B-cell lymphomas from T-cell or other non-B-cell lymphomas. Positive CD22 expression indicates that the tumor cells are of B-cell origin and is commonly observed in most B-cell lymphomas. When qualitatively assessing CD22 expression, positive results are usually characterized by brown staining on the tumor cell membrane. Strong and widespread positivity typically suggests a clear B-cell origin, whereas negative or weak expression may indicate certain special B-cell lymphoma subtypes or the need to rule out non-B-cell lymphomas. This study aims to qualitatively analyze CD22 expression using 143 WSI samples. A weakly supervised classification approach was employed, with ABMIL selected as the model framework. Training parameters were set to 50 epochs, with a batch size of 1 and a learning rate of 0.00002. The dataset was divided into training, validation, and test sets in an 8:1:1 ratio. All training operations were conducted on systems equipped with A100 GPUs, with images magnified at 20x and resized to 256$\times$256 pixels.

\noindent\textit{84. CD8 IHC Qualification}. CD8 is a glycoprotein predominantly expressed on the surface of cytotoxic T cells (CTLs) and a subset of NK cells, playing a pivotal role in immune responses. In lymphoma diagnosis and classification, CD8 holds significant clinical importance, particularly in distinguishing various subtypes of T-cell lymphomas. The presence of CD8 typically suggests that the tumor cells may originate from cytotoxic T cells. Additionally, the intensity and distribution of CD8 expression can assist in differentiating between types of T-cell lymphomas and in assessing immune cell infiltration within the tumor microenvironment. Pathologists primarily evaluate CD8 expression by examining the staining of the cell membrane and portions of the cytoplasm. A high proportion of CD8-positive cells generally indicates a tumor with a clear cytotoxic T-cell phenotype, while negative or weak expression may suggest a non-cytotoxic T-cell origin or another type of lymphoma. This study aimed to qualitatively analyze CD8 expression using 120 WSI samples. A weakly supervised classification approach was employed, utilizing ABMIL as the model framework. Training parameters were set to 50 epochs, with a batch size of 1 and a learning rate of 0.00002. The dataset was divided into training, validation, and test sets in an 8:1:1 ratio. All training operations were conducted on systems equipped with A100 GPUs, with images magnified at 20x and resized to 256$\times$256 pixels.

\noindent\textit{85. C-myc IHC Qualification}. C-myc is a nuclear transcription factor that plays a critical role in the diagnosis and prognosis of lymphoma. Overexpression of C-myc is typically associated with highly aggressive lymphomas, which often exhibit translocation or amplification of the myc gene, leading to elevated protein expression. This correlates with rapid tumor growth, increased malignancy, and poor prognosis. When qualitatively assessing C-myc expression, pathologists focus on the staining intensity within the tumor cell nuclei. High-intensity and widespread C-myc positivity usually indicates a highly aggressive tumor and is often associated with C-myc gene rearrangement or amplification. This study aims to perform a qualitative analysis of C-myc expression using 293 WSI samples. A weakly supervised classification method was employed, utilizing ABMIL as the model framework. Training parameters were set to 50 epochs, with a batch size of 1 and a learning rate of 0.00002. The dataset was divided into training, validation, and test sets in an 8:1:1 ratio. All training operations were conducted on systems equipped with A100 GPUs, with images magnified at 20x and resized to 256$\times$256 pixels.

\noindent\textit{86. CD56 IHC Qualification}. CD56 expression holds significant clinical importance in the diagnosis and classification of lymphomas, particularly in identifying subtypes associated with natural killer (NK) cells and certain T-cell subsets. CD56 expression not only helps confirm the tumor's origin but is also linked to highly aggressive lymphomas with poor prognosis. When qualitatively assessing CD56 expression, pathologists focus on the staining patterns in the tumor cell membrane and parts of the cytoplasm. Strong and widespread CD56 positivity often indicates that the tumor cells may originate from NK cells or CD56-positive T-cell subtypes, which are typically associated with higher invasiveness and worse prognosis. Conversely, negative or weak CD56 expression may suggest that the tumor originates from non-NK/T-cell lineages or subtypes not associated with CD56 expression. This study aims to conduct a qualitative analysis of CD56 expression using 88 WSI samples. A weakly supervised classification approach was employed, utilizing ABMIL as the model framework. Training parameters were set to 50 epochs, with a batch size of 1 and a learning rate of 0.00002. The dataset was divided into training, validation, and test sets in an 8:1:1 ratio. All training operations were conducted on systems equipped with A100 GPUs, with images magnified at 20x and resized to 256$\times$256 pixels.

\noindent\textit{87. GranzymeB IHC Qualification}. Granzyme B expression holds significant clinical value in the classification, diagnosis, and prognosis of lymphomas, particularly in identifying lymphomas derived from cytotoxic T cells or NK cells. The detection of Granzyme B not only aids in confirming the cytotoxic origin of the tumor but also provides insights into the tumor's biological behavior, as Granzyme B positivity is often associated with a more aggressive tumor phenotype. When qualitatively assessing Granzyme B expression, pathologists primarily examine the staining within the cytoplasm of tumor cells. A strong and widespread positive expression typically indicates active cytotoxicity in the tumor cells, which may correlate with higher aggressiveness. This study aims to perform a qualitative analysis of Granzyme B expression using 80 WSI samples. A weakly supervised classification approach was employed, utilizing ABMIL as the model framework. Training parameters were set to 50 epochs, with a batch size of 1 and a learning rate of 0.00002. The dataset was divided into training, validation, and test sets in an 8:1:1 ratio. All training operations were conducted on systems equipped with A100 GPUs, with images magnified at 20x and resized to 256$\times$256 pixels.

\noindent\textit{88. TIA-1 IHC Qualification}. The evaluation of TIA-1 expression plays a significant role in the classification, diagnosis, and prognosis of lymphomas, particularly in distinguishing lymphomas of cytotoxic T-cell or NK-cell origin. TIA-1 positivity indicates that tumor cells have cytotoxic potential, helping confirm the cytotoxic origin of the tumor and providing insights into the tumor's biological behavior. When qualitatively assessing TIA-1 expression, pathologists primarily observe the staining within the cytoplasm of tumor cells. Strong positive and widespread TIA-1 expression typically suggests that the tumor cells have active cytotoxic functions, indicating a more aggressive behavior. This study aims to qualitatively analyze TIA-1 expression using 80 WSI samples. A weakly supervised classification method was employed, utilizing ABMIL as the model framework. Training parameters were set to 50 epochs, with a batch size of 1 and a learning rate of 0.00002. The dataset was divided into training, validation, and test sets in an 8:1:1 ratio. All training operations were conducted on systems equipped with A100 GPUs, with images magnified at 20x and resized to 256$\times$256 pixels.

\noindent\textit{89. Perforin IHC Qualification}. Perforin is a protein secreted by cytotoxic T cells (CTLs) and natural killer (NK) cells, playing a crucial role in the apoptosis of target cells. Its expression is of significant clinical importance in the diagnosis and classification of lymphomas, particularly in identifying lymphoma subtypes associated with cytotoxic T cells or NK cells. When qualitatively assessing Perforin expression, pathologists focus on the staining within the cytoplasm of tumor cells. Extensive and intense Perforin expression typically indicates a strong cytotoxic function within the tumor, which may be associated with higher aggressiveness and a poorer prognosis. This study aims to perform a qualitative analysis of Perforin expression using 56 WSI samples. A weakly supervised classification approach was employed, utilizing ABMIL as the model framework. Training parameters were set to 50 epochs, with a batch size of 1 and a learning rate of 0.00002. The dataset was divided into training, validation, and test sets in an 8:1:1 ratio. All training operations were conducted on systems equipped with A100 GPUs, with images magnified at 20x and resized to 256$\times$256 pixels.

\noindent\textit{90. CD2 IHC Qualification}. CD2 is a transmembrane glycoprotein crucial in the diagnosis, classification, and prognosis of lymphomas, particularly for identifying T-cell or NK-cell lineage. In the qualitative assessment of CD2 expression, pathologists focus on the staining patterns observed on the tumor cell membrane. Extensive and strong CD2 expression typically indicates a definitive T-cell or NK-cell phenotype, supporting the corresponding lymphoma diagnosis. This study aimed to qualitatively analyze CD2 expression using a dataset of 69 WSI samples. We employed a weakly supervised classification method, utilizing ABMIL as the model framework, with training parameters set to 50 epochs, a batch size of 1, and a learning rate of 0.00002. The dataset was divided into training, validation, and test sets in an 8:1:1 ratio. All training operations were conducted on A100 GPU-equipped systems, with images magnified at 20x and resized to 256$\times$256 pixels.

\noindent\textit{91. CD30 IHC Qualification}. CD30 which is a transmembrane protein belonging to the tumor necrosis factor (TNF) receptor superfamily, is primarily expressed on activated T cells, B cells, and germ cells. CD30 expression is a hallmark of classical Hodgkin lymphoma, typically presenting as intense brown staining on the tumor cell membrane, often with Golgi area staining within the cytoplasm. Strong and widespread CD30 expression generally suggests the presence of classical Hodgkin lymphoma or anaplastic large cell lymphoma, whereas weak or negative expression may indicate other lymphoma types or non-neoplastic conditions. This study aimed to qualitatively analyze CD30 expression using 94 WSI samples. We employed a weakly supervised classification approach, utilizing ABMIL as the model framework, with training parameters set to 50 epochs, a batch size of 1, and a learning rate of 0.00002. The dataset was divided into training, validation, and test sets in an 8:1:1 ratio. All training operations were conducted on A100 GPU-equipped systems, with images magnified at 20x and resized to 256$\times$256 pixels.

\noindent\textit{92. CD7 IHC Qualification}. CD7 is a transmembrane glycoprotein widely expressed on the surface of T cells and some NK cells, serving as one of the earliest markers during T cell differentiation. CD7 expression is crucial for determining the T cell lineage of tumor cells and assessing their biological behavior. In the qualitative evaluation of CD7 expression, pathologists primarily observe staining on the tumor cell membrane. Strong and widespread CD7 expression typically indicates a clear T cell phenotype
This study aimed to qualitatively analyze CD7 expression using a dataset of 62 WSI samples. We employed a weakly supervised classification method, utilizing ABMIL as the model framework, with training parameters set to 50 epochs, a batch size of 1, and a learning rate of 0.00002. The dataset was divided into training, validation, and test sets in an 8:1:1 ratio. All training operations were conducted on A100 GPU-equipped systems, with images magnified at 20x and resized to 256$\times$256 pixels.

\noindent\textit{93. CD38 IHC Qualification}. CD38 is a transmembrane glycoprotein broadly expressed on plasma cells, activated T cells, B cells, and NK cells. Its expression suggests that tumor cells may exhibit characteristics of plasma cells or activated lymphocytes, which is particularly significant in the diagnosis of multiple myeloma, plasma cell leukemia, and certain proliferative B-cell and T-cell lymphomas. When qualitatively assessing CD38 expression, pathologists focus on the staining of the tumor cell membrane and parts of the cytoplasm. Strong and widespread CD38 expression typically indicates that the tumor cells may originate from plasma cells or possess a high activation state, which is crucial for diagnosing multiple myeloma and related disorders. This study aimed to qualitatively analyze CD38 expression using a dataset of 46 WSI samples. We employed a weakly supervised classification method, utilizing ABMIL as the model framework, with training parameters set to 50 epochs, a batch size of 1, and a learning rate of 0.00002. The dataset was divided into training, validation, and test sets in an 8:1:1 ratio. All training operations were conducted on A100 GPU-equipped systems, with images magnified at 20x and resized to 256$\times$256 pixels.

\noindent\textit{94. ICOS IHC Qualification}. ICOS (Inducible T-cell CO-Stimulator) is a co-stimulatory molecule expressed on activated T cells, playing a crucial role in regulating T-cell proliferation and function. Evaluating ICOS expression is clinically significant for diagnosing and classifying T-cell lymphomas and understanding immune regulation mechanisms within the tumor microenvironment. Positive ICOS expression, typically presenting as brown staining on the tumor cell membrane, often indicates that the tumor originates from activated T cells, which may correlate with more aggressive behavior. This study aimed to qualitatively analyze ICOS expression using a dataset of 46 WSI samples. We employed a weakly supervised classification method, utilizing ABMIL as the model framework, with training parameters set to 50 epochs, a batch size of 1, and a learning rate of 0.00002. The dataset was divided into training, validation, and test sets in an 8:1:1 ratio. All training operations were conducted on A100 GPU-equipped systems, with images magnified at 20x and resized to 256$\times$256 pixels.

\noindent\textit{95. CXCL-13 IHC Qualification}. CXCL-13 (C-X-C motif chemokine ligand 13) is a chemokine secreted by follicular helper T cells (Tfh cells) and other immune cells, playing a critical role in B cell chemotaxis and the formation and maintenance of germinal centers. The expression of CXCL-13 holds significant clinical relevance in diagnosing lymphomas associated with germinal center reactions. Positive CXCL-13 expression, typically manifesting as brown staining in the cytoplasm of tumor cells or around follicular helper T cells, often suggests a lymphoma subtype linked to follicular reactions or germinal centers, potentially correlating with specific tumor microenvironments and biological behaviors. This study aimed to qualitatively analyze CXCL-13 expression using a dataset of 52 WSI samples. We employed a weakly supervised classification method, utilizing the ABMIL framework, with training parameters set to 50 epochs, a batch size of 1, and a learning rate of 0.00002. The dataset was divided into training, validation, and test sets in an 8:1:1 ratio. All training operations were conducted on A100 GPU-equipped systems, with images magnified at 20x and resized to 256$\times$256 pixels.

\noindent\textit{96. ALK IHC Qualification}. Evaluating ALK (Anaplastic Lymphoma Kinase) expression holds critical clinical significance in the diagnosis, classification, and prognosis of lymphomas, particularly in identifying subtypes of anaplastic large cell lymphoma (ALCL) and in formulating personalized treatment strategies. ALK positivity typically presents as brown staining in both the nuclei and cytoplasm of tumor cells, with staining patterns varying from granular to diffuse. This study aimed to conduct a qualitative analysis of ALK expression using a dataset of 23 WSI samples. We employed a weakly supervised classification approach, utilizing the ABMIL framework, with training parameters set to 50 epochs, a batch size of 1, and a learning rate of 0.00002. The dataset was divided into training, validation, and test sets in an 8:1:1 ratio. All training operations were conducted on A100 GPU-equipped systems, with images magnified at 20x and resized to 256$\times$256 pixels.

\noindent\textit{97. PD-L1 IHC Qualification}. We have developed a model for distinguishing ROI-level IHC images in the context of PD-L1 qualification. To rigorously evaluate the model's performance, we employed a dataset comprising 10,150 images, with 8,120 allocated for training and 2,030 reserved for testing. The training phase consisted of 50 epochs with a batch size of 256. All training procedures were conducted on A100 GPU-equipped systems to ensure high processing speed and computational efficiency. Furthermore, to meet the input requirements of the model, the images were processed at 20x magnification and resized to 256$\times$256 pixels.

\noindent\textit{98. Breast HER2 Scoring}. In the HER2 (Human Epidermal Growth Factor Receptor 2) scoring task, we focused on classifying this protein in breast cancer. HER2 is a protein that promotes cancer cell growth, and approximately 15\% to 20\% of breast cancer patients are HER2 positive, indicating an excess of HER2 protein in their cancer cells. HER2 scoring is critical in pathology, especially for breast cancer diagnosis and treatment. The HER2 scores are categorized into four levels: 0, 1+, 2+, and 3+. The training set includes 106 ROI images for score 0, 59 ROI images for score 1+, 172 ROI images for score 2+, and 135 ROI images for score 3+. The test set includes 26 ROI images for score 0, 14 ROI images for score 1+, 43 ROI images for score 2+, and 33 ROI images for score 3+. We used Linear Probing as the training strategy, with parameters set to a maximum of 1000 iterations. ROIs were cropped at 10x magnification and resized to 256$\times$256 pixels. All training operations were conducted on NVIDIA Tesla V100 GPU-equipped systems.

\noindent\textit{99. ER IHC Qualification}. Detecting the expression of ER (estrogen receptor) in cancer cells can assess the tumor's sensitivity to hormone therapy. ER positivity indicates that tumor cells may respond to endocrine therapy, significantly improving patient prognosis. Additionally, ER status is a critical prognostic indicator, as ER-positive patients generally have a better prognosis than ER-negative patients. Accurate ER IHC classification can aid in developing personalized treatment plans, guiding therapeutic decisions, and ultimately improving survival and quality of life for breast cancer patients. The training set included 505 WSIs for ER-positive and 627 WSIs for ER-negative cases. The test set included 338 WSIs for ER-positive and 419 WSIs for ER-negative cases. Training was conducted using the ABMIL framework with 50 epochs and a learning rate of 0.00002. WSIs were cropped into patches at 10x magnification and resized to 256$\times$256 pixels. All training operations were conducted on NVIDIA Tesla V100 GPU-equipped systems (Quantitative analysis and assessment of expression intensity for ER is clinically significant. However, due to limitations in the ground truth of the dataset, we currently used qualitative assessment after thresholding. We aim to improve this issue in the future.).

\noindent\textit{100. PR IHC Qualification}. In breast cancer, detecting the expression of PR (progesterone receptor) in cancer cells can evaluate the tumor's sensitivity to hormone therapy. PR positivity suggests that the tumor cells are responsive to hormone therapy, indicating that these patients might benefit from endocrine treatments. Additionally, the qualitative classification of PR immunohistochemical staining can provide critical information for clinicians, enhancing treatment outcomes and survival rates. The training set included 333 WSIs for PR-positive and 381 WSIs for PR-negative cases. The test set included 233 WSIs for PR-positive and 255 WSIs for PR-negative cases. Training was conducted using the ABMIL framework with 50 epochs and a learning rate of 0.00002. WSIs were cropped into patches at 10x magnification and resized to 256$\times$256 pixels. All training operations were conducted on NVIDIA Tesla V100 GPU-equipped systems (Quantitative analysis and assessment of expression intensity for PR is clinically significant. However, due to limitations in the ground truth of the dataset, we currently used qualitative assessment after thresholding. We aim to improve this issue in the future.).

\noindent\textit{101. Ki67 IHC Qualification}. The significance of Ki67 in breast cancer diagnosis lies in its ability to evaluate tumor cell proliferation, which directly correlates with prognosis. A high Ki67 index is typically associated with poor prognosis, while a low Ki67 index suggests a better prognosis, thereby aiding in treatment decision-making. The Ki67-positive dataset includes 553 WSIs in the training set and 369 WSIs in the test set, while the Ki67-negative dataset comprises 399 WSIs in the training set and 267 WSIs in the test set. We utilized ABMIL for model training, with the parameters set to 50 epochs, a learning rate of 0.00002, and a batch size of 1. ROIs were cropped at 10x magnification and resized to 256$\times$256 pixels. All training operations were conducted on systems equipped with NVIDIA Tesla V100 GPUs  (Quantitative analysis and assessment of expression intensity for Ki67 is clinically significant. However, due to limitations in the ground truth of the dataset, we currently used qualitative assessment after thresholding. We aim to improve this issue in the future.).

\noindent\textit{102. HER2 IHC Qualification}. The HER2 task focuses on determining the HER2 status (positive or negative) of the provided slides. The training set includes 352 WSIs with HER2-positive data and 480 WSIs with HER2-negative data, while the test set comprises 233 WSIs with HER2-positive data and 255 WSIs with HER2-negative data. We employed ABMIL for model training, setting the parameters to 50 epochs, a learning rate of 0.00002, and a batch size of 1. ROIs were cropped at 10x magnification and resized to 256$\times$256 pixels. All training operations were conducted on systems equipped with NVIDIA Tesla V100 GPUs.

\noindent\textbf{VI. Gene Expression Prediction Tasks}\\
\noindent\textit{103. Multi-Disease Gene Expression}. We employed the HEST-Benchmark dataset~\cite{jaume2024hest}, a publicly available resource designed for predicting gene expression in human cancer samples across nine distinct sub-datasets encompassing a total of 71 samples, each corresponding to a different organ. The sub-datasets include Invasive Ductal Carcinoma (IDC), Prostate Adenocarcinoma (PRAD), Skin Cutaneous Melanoma (SKCM), Colon Adenocarcinoma (COAD), Rectal Adenocarcinoma (READ), Clear Cell Renal Cell Carcinoma (ccRCC), Hepatocellular Carcinoma (HCC), Lung Adenocarcinoma (LUAD), and Axillary Lymph Nodes in IDC patients (LYMPH\_IDC). The dataset comprises image patches captured at 20$\times$ magnification with a resolution of 224$\times$224 pixels, aiming to predict the expression levels of the top 50 genes exhibiting the highest normalized variance. For our evaluations, we conducted experiments on each of the nine sub-datasets as well as on the combined dataset (excluding Pancreatic Adenocarcinoma, PAAD, due to data limitations). We adhered to the official training and testing split methodology provided by the HEST-Benchmark, utilizing k-fold cross-validation where $k$ corresponds to the number of patients in each sub-dataset. This patient-level cross-validation ensures that data from the same patient do not appear in both training and testing sets, thereby preventing data leakage and ensuring the robustness of our evaluation. Training was performed over 50 epochs using a learning rate of 0.001 and a batch size of 32. The model was trained on a single NVIDIA A100 GPU, leveraging its computational capabilities to handle the high-dimensional data and to expedite the training process.

\noindent\textit{104. IDC Gene Expression Evaluation}. This task focuses on gene expression prediction in invasive ductal carcinoma (IDC) of the breast. We utilized a total of four samples from four patients, comprising all publicly available Xenium samples from 10x Genomics (two FFPE human breast tissue samples covering the entire sample area) and two additional samples published by Janesick et al.~\cite{janesick2023high}. All samples consist of FFPE sections imaged using the Xenium pipeline v1.

\noindent\textit{105. PRAD Gene Expression Evaluation}. For prostate adenocarcinoma (PRAD), we employed 23 Visium samples consisting of fresh frozen sections from two patients, as published by Erickson et al.~\cite{erickson2022spatially}. Both patients were diagnosed with prostatic acinar adenocarcinoma with a Gleason score of 4+3 (International Society of Urological Pathology [ISUP] group 4). These samples enabled us to assess gene expression prediction in prostate cancer tissues with a consistent histopathological grading.

\noindent\textit{106. SKCM Gene Expression Evaluation}. In the context of skin cutaneous melanoma (SKCM), we utilized two samples from different patients, sourced from the 10x Genomics repository under the title "Human Skin Data with Xenium Human Multi-Tissue and Cancer Panel." All samples are FFPE sections processed using the Xenium pipeline version 1, allowing us to evaluate gene expression prediction in melanoma tissues.

\noindent\textit{107. COAD Gene Expression Evaluation}. For colon adenocarcinoma (COAD), we analyzed six samples from three different patients, as published by Valdeolivas et al.~\cite{valdeolivas2023charting}. All samples consist of fresh frozen sections processed using the Visium spatial transcriptomics platform. This setup facilitated the assessment of gene expression prediction in colorectal cancer tissues.

\noindent\textit{108. READ Gene Expression Evaluation}. In the case of rectal adenocarcinoma (READ), we employed four samples from two different patients, also published by Valdeolivas et al.~\cite{valdeolivas2023charting}. Similar to the COAD samples, these are fresh frozen sections processed with the Visium platform, enabling us to extend our gene expression prediction evaluations to rectal cancer tissues.

\noindent\textit{109. CCRCC Gene Expression Evaluation}. For clear cell renal cell carcinoma (ccRCC), we utilized all 24 samples from 24 different patients, as published by Meylan et al.~\cite{meylan2022tertiary}. These samples consist of fresh frozen sections processed using the Visium platform, providing a substantial dataset to evaluate gene expression prediction in renal cancer.

\noindent\textit{110. HCC Gene Expression Evaluation}. In hepatocellular carcinoma (HCC), we analyzed two liver samples from two different patients, as published by Giraud et al.~\cite{giraud2022trem1}. Both samples are fresh frozen sections processed using the Visium platform, facilitating the evaluation of gene expression prediction in liver cancer tissues.

\noindent\textit{111. LUAD Gene Expression Evaluation}. For lung adenocarcinoma (LUAD), we used two samples from two different patients, sourced from 10x Genomics under the title "Preview Data: FFPE Human Lung Cancer with Xenium Multimodal Cell Segmentation." These samples are FFPE sections processed using the Xenium pipeline version 1, enabling gene expression prediction assessments in lung cancer tissues.

\noindent\textit{112. LYMPH\_IDC Gene Expression Evaluation}. This task focuses on gene expression prediction in axillary lymph nodes of IDC patients. We utilized four axillary lymph node samples from two IDC patients, as published by Liu et al.~\cite{liu2022single}. All samples consist of fresh frozen sections processed using the Visium platform, allowing us to assess gene expression prediction in metastatic lymph node tissues associated with breast cancer.

\clearpage
\noindent\textbf{\large{APPENDIX II}}

\begin{longtable}{ccccc}
	\caption{List of 112 Clinical Downstream Tasks and Datasets.} \\
	\toprule
	\textbf{Order} & \textbf{Task} & \textbf{Dataset} & \textbf{Open-Source} \\ \midrule
	\endfirsthead
	
	\toprule
	\textbf{Task} & \textbf{Name}  & \textbf{Dataset} & \textbf{Open-Source} \\ \midrule
	\endhead
	
	\midrule
	\multicolumn{5}{r}{\textit{Continued on next page}} \\
	\midrule
	\endfoot
	
	\bottomrule
	\endlastfoot
	
	\multirow{2}{*}{1} & \multirow{2}{*}{Natural vs. Pathological Image Clas.} & ImageNet-1K/ILSVRC2012 & \checkmark \\
	  &  &  Gastrointestinal H\&E and IHC & \\
	2 & Tissue Fold Det. & Gastrointestinal H\&E & \\
	3 & Bubble \& Glue Clas. & H\&E Data & \\
	4 & Contaminant Det. & H\&E Data & \\
	5 & Blur Det. & FocusPath-UofT & \checkmark \\
	6 & Pan-Organ H\&E \& IHC Recog. & Pan-Organ H\&E \\
	\multirow{2}{*}{7} & \multirow{2}{*}{Pan-Organ H\&E \& Fluorescence Recog.} & Pan-Organ Dataset \\
	  & & DeepCell & \checkmark \\
	8 & IHC Marker Type Recog. & Multi-IHCs & \\
	9 & Pan-Organ Multi-Magnification Recog. & Pan-Organ H\&E & \\
	10 & Pan-Organ Frozen vs. FFPE Recog. & TCGA-Pan-Organ & \checkmark \\
	11 & Pan-Organ Biopsy vs. Large Specimen Recog. & Pan-Organ H\&E &  \\
	12 & IHC Nuclear vs. Membrane+ Recog. & IHC Nuclear \& Membrane &  \\
	13 & Pan-Cancer Clas. (17-Subtype) & Pan-cancer H\&E & \\
	14 & FFPE Pan-Cancer Clas. (32-Subtype) & TCGA-FFPE & \checkmark \\
	15 & Frozen Pan-Cancer Clas. (32-Subtype) & TCGA-Frozen & \checkmark \\
	16 & Breast Metastasis Det. & CAMELYON16 & \checkmark \\
	17 & Breast Metastasis Det. & CAMELYON17 & \checkmark \\
	18 & Lymphoma PD-L1 Tumor Region Det. & PD-L1 Data & \\
	19 & TIL Negative vs. Positive Clas. & TCGA-TIL & \checkmark \\
	20 & Breast Cancer Clas. & PCam & \checkmark \\
	21 & Ring cell Clas. & DigestPath & \checkmark \\
	22 & SegPath (Multi-Class) Seg. & SegPath & \checkmark \\
	23 & Gland Segmentation & GlaS & \checkmark \\
	24 & Nuclei Seg. & PanNuke & \checkmark \\
	25 & Colorectal Nuclear Seg. & CoNSeP & \checkmark \\
	26 & COSAS Seg. (Adenocarcinoma) & COSAS & \checkmark \\
	27 & COSAS Seg. (Cell Type) & COSAS & \checkmark \\
	28 & SegPath (Single-Class) Det. & SegPath & \checkmark \\
	29 & TCT Tumor Lesion Det. & TCT Data &  \\
	30 & PE\&A Neoplastic Cell Few-shot Det. & PE\&A Data &  \\
	31 & Cervical Cancer Positive Cell Subtyp. & Cervical Data &  \\
	32 & Bladder Benign-Malignant Clas. & Bladder Data &  \\
	33 & TCT Benign-Malignant Clas (WSI). & TCT Data &  \\
	34 & Bladder Benign-Malignant Clas. (Patch) & Bladder Data &  \\
	35 & TCT Benign-Malignant Clas. (Patch) & TCT Data &  \\
	36 & Digestive Tract Benign-Malignant Clas. & Digestive H\&E & \\
	37 & Lymphoma vs Reactive Hyperplasia Clas. & Lymphoma H\&E & \\
	38 & Lymphoma B/T Cells Clas. & Lymphoma H\&E & \\
	39 & Intestinal Cancer Det. & Intestinal H\&E & \\
	40 & Cervical Cell Clas. (TissueNet) & TissueNet & \checkmark \\
	41 & Colon and Lung Tissues Clas. (LC25K-5) & LC25K & \checkmark \\
	42 & Lung Subtyp. (LC25K-3) & LC25K & \checkmark \\
	43 & Colon Subtyp. (LC25K-2) & LC25K & \checkmark \\
	44 & Breast Benign-Malignant Clas. (BreaKHis) & BreaKHis & \checkmark \\
	45 & NSCLC Subtyp. (TCGA+) & TCGA-NSCLC & \checkmark \\
	46 & RCC Subtyping (TCGA+) & TCGA-RCC & \checkmark \\
	47 & Lymphoma Subtype Multi-class Clas. & Lymphoma H\&E & \checkmark \\
	48 & BRCA Subtyp. (BACH) & TCGA-BACH & \checkmark \\
	49 & ESCA Subtyp. (TCGA+) & TCGA-ESCA & \checkmark \\
	50 & CRC Screen. (HunCRC) & HunCRC & \checkmark \\
	51 & Prostate ISUP Grad. (PANDA) & PANDA & \checkmark \\
	52 & Prostate Gleason Grad. & H\&E Data &  \\
	53 & Intestinal Cancer High-Grade vs. Low-Grade Clas. & Intestinal H\&E & \\
	54 & PatchGastricADC22 (Subtype-2) & PatchGastricADC22 & \checkmark \\
	55 & PatchGastricADC22 (Subtype-10) & PatchGastricADC22 & \checkmark \\
	56 & PRAD Tissue Clas. (AGGC) & AGGC & \checkmark \\
	57 & Glioma IDH1 Screen. (TCGA+) & TCGA-GBM & \checkmark \\
	58 & Pan-Organ Lymph Node Metastasis Clas. & Pan-Organ H\&E & \\
	59 & Intestinal Cancer and Polyp Type Clas. & Intestinal H\&E & \\
	60 & CRC Tissue Clas. (HunCRC) & HunCRC & \checkmark \\
	61 & RCC Tissue Clas. (TCGA+) & TCGA-RCC & \checkmark \\
	62 & CRC Tissue Clas. (CRC-100K) & CRC-100K & \checkmark \\
	63 & Colon Subtyp. (Chaoyang) & Chaoyang & \checkmark \\
	64 & WSSS4LUAD Clas. & WSSS4LUAD & \checkmark \\
	65 & CRC Clas. (Kather) & Kather & \checkmark \\
	66 & Multi-task Clas. (Ebrains) & Ebrains & \checkmark \\
	67 & Multi-IHC Qual. & Multi-IHC Dataset \\
	68 & CD5 IHC Qual. & IHC-CD5 \\
	69 & CD3 IHC Qual. & IHC-CD3 \\
	70 & CD20 IHC Qual. & IHC-CD20 \\
	71 & CD79a IHC Qual. & IHC-CD79a \\
	72 & CD21 IHC Qual. & IHC-CD21 \\
	73 & EBER Qual. & IHC-EBER \\
	74 & CD10 IHC Qual. & IHC-CD10 \\
	75 & Bcl-6 IHC Qual. & IHC-Bcl-6 \\
	76 & Bcl-2 IHC Qual. & IHC-Bcl-2 \\
	77 & MUM-1 IHC Qual. & IHC-MUM-1 \\
	78 & CD4 IHC Qual. & IHC-CD4 \\
	79 & CD23 IHC Qual. & IHC-CD23 \\
	80 & PD-1 IHC Qual. & IHC-PD-1 \\
	81 & Cyclin D1 IHC Qual. & IHC-Cyclin D1 \\
	82 & CD19 IHC Qual. & IHC-CD19 \\
	83 & CD22 IHC Qual. & IHC-CD22 \\
	84 & CD8 IHC Qual. & IHC-CD8 \\
	85 & C-myc IHC Qual. & IHC-C-myc \\
	86 & CD56 IHC Qual. & IHC-CD56 \\
	87 & Granzyme B IHC Qual. & IHC-Granzyme B \\
	88 & TIA-1 IHC Qual. & IHC-TIA-1 \\
	89 & Perforin IHC Qual. & IHC-Perforin \\
	90 & CD2 IHC Qual. & IHC-CD2 \\
	91 & CD30 IHC Qual. & IHC-CD30 \\
	92 & CD7 IHC Qual. & IHC-CD7 \\
	93 & CD38 IHC Qual. & IHC-CD38 \\
	94 & ICOS IHC Qual. & IHC-ICOS \\
	95 & CXCL-13 IHC Qual. & IHC-CXCL-13 \\
	96 & ALK IHC Qual. & IHC-ALK \\
	97 & PD-L1 IHC Qual. & IHC-PD-L1 \\
	98 & Breast HER2 Scoring & Breast-HER2 \\
	99 & ER IHC Qual. & Breast-ER \\
	100 & PR IHC Qual. & Breast-PR \\
	101 & Ki67 IHC Qual. & Breast-Ki67 \\
	102 & HER2 IHC Qual. & Breast-HER2 \\
	103 & Multi-Disease Gene Expression & HEST-Benchmark & \checkmark \\
	104 & IDC Gene Exp. & HEST-IDC & \checkmark \\
	105 & PRAD Gene Exp. & HEST-PRAD & \checkmark \\
	106 & SKCM Gene Exp. & HEST-SKCM & \checkmark \\
	107 & COAD Gene Exp. & HEST-COAD & \checkmark \\
	108 & READ Gene Exp. & HEST-READ & \checkmark \\
	109 & CCRCC Gene Exp. & HEST-CCRCC & \checkmark \\
	110 & HCC Gene Exp. & HEST-HCC & \checkmark \\
	111 & LUAD Gene Exp. & HEST-LUAD & \checkmark \\
	112 & LYMPH\_IDC Gene Exp. & HEST-LYMPH\_IDC & \checkmark\\ 
	
\end{longtable}

\clearpage

\begin{longtable}{cccc|c}
    \caption{Details of 112 Downstream Task Datasets.} \\
	\toprule
	\textbf{Task} & \textbf{Quantity} & \textbf{Level} & \textbf{Site} & \textbf{Category} \\ \midrule
	\endfirsthead
	
	\toprule
	\textbf{Task} & \textbf{Quantity} & \textbf{Level} & \textbf{Site} & \textbf{Category} \\ \midrule
	\endhead
	
	\midrule
	\multicolumn{5}{r}{\textit{Continued on next page}} \\
	\midrule
	\endfoot
	
	\bottomrule
	\endlastfoot
	1 &     43,730 &   ROI &          Multi-Organ &        \multirow{12}{*}{Slide Preprocessing} \\
	2 &     37,461 &   ROI &          Multi-Organ &                            \\
	3 &     12,432 &   ROI &          Multi-Organ &                            \\
	4 &     10,035 &   ROI &          Multi-Organ &                            \\
	5 &       864 &   ROI &          Multi-Organ &                            \\
	6 &     21,865 &   ROI &          Multi-Organ &                            \\
	7 &     15,022 &   ROI &          Multi-Organ &                            \\
	8 &      2,689 &   ROI &          Multi-Organ &                            \\
	9 &    286,712 &   ROI &          Multi-Organ &                            \\
	10 &    289,469 &   ROI &          Multi-Organ &                            \\
	11 &       726 &   ROI &          Multi-Organ &                            \\
	12 &      1,812 &   ROI &          Multi-Organ &                            \\
	\midrule
	13 &      1,071 &   WSI &           Pan-Cancer &                 \multirow{3}{*}{Pan-Cancer} \\
	14 &      1,483 &   WSI &           Pan-Cancer &                            \\
	15 &      3,038 &   WSI &           Pan-Cancer &                            \\
	\midrule
	16 &       402 &   WSI &               Breast &      \multirow{15}{*}{Lesion Identification} \\
	17 &       500 &   WSI &               Breast &                            \\
	18 &      8,540 &   ROI &             Lymphoma &                            \\
	19 &     304,097 &   ROI &             Multi-Organ &                            \\
	20 &    2,948,912 &   ROI &             Breast &                            \\
	21 &       4,477 &   ROI &             Colorectal &                            \\
	22 &     158,687 &   ROI &             Multi-Organ &                            \\
	23 &       165 &   ROI &             Intestine &                            \\
	24 &      7,901 &   ROI &             Multi-Organ &                            \\
	25 &       656 &   ROI &             Intestine &                            \\
	26 &       190 &   ROI &             Multi-Organ &                            \\
	27 &       180 &   ROI &             Multi-Organ &                            \\
	28 &      4,477 &   ROI &             Multi-Organ &                            \\
	29 &   1,378,941 &   ROI &             Cervical Cancer &                         \\
	30 &     36,544 &   ROI &               Multi-Organ &                            \\
	\midrule
	31 & 215,296 & ROI & Cervix & \\
	32 & 324 & WSI & Bladder & \\
	33 & 1,497 & WSI & Cervix & \\
	34 & 112 & ROI & Bladder & \\
	35 & 1,378,941 & ROI & Cervix & \\
	36 & 825 & WSI & Digestive Tract & \\
	37 & 688 & WSI & Lymph Nodes & \\
	38 & 645 & WSI & Lymph Nodes & \\
	39 & 275 & WSI & Intestine & \\
	40 & 1,005 & WSI & Cervix & \\
	41 & 25,000 & ROI & Lung/Intestine & \\
	42 & 25,000 & ROI & Lung & \\
	43 & 10,000 & ROI & Intestine & \\
	44 & 1,693 & WSI & Breast & \\
	45 & 922 & WSI & Lung & \\
	46 & 1,043 & WSI & Kidney & \\
	47 & 363 & WSI & Lymph Nodes & \\
	48 & 400 & ROI & Breast & \\
	49 & 367,330 & ROI & Esophageal Cancer & \multirow{1}{*}{Cancer Subtyping} \\
	50 & 200 & WSI & Intestine & \\
	51 & 10,616 & ROI & Prostate & \\
	52 & 1,042 & WSI & Prostate & \\
	53 & 110 & WSI & Intestine & \\
	54 & 133,826 & ROI & Stomach & \\
	55 & 262,777   & ROI & Stomach          & \\
	56 & 1,126,447  & ROI & Prostate         & \\
	57 & 1,698     & WSI & Brain            & \\
	58 & 922      & ROI & Lymph Nodes       & \\
	59 & 400      & WSI & Intestine        & \\
	60 & 101,398   & ROI & Intestine        & \\
	61 & 52,713    & ROI & Kidney           & \\
	62 & 100,000   & ROI & Intestine        & \\
	63 & 6,160     & ROI & Intestine        & \\
	64 & 1,176     & ROI & Lung             & \\
	65 & 5,000     & ROI & Intestine        & \\
	66 & 963      & WSI & Brain            & \\
	\midrule
	67 & 6,405  & WSI & Multi-Organ          & \multirow{36}{*}{Biomarker Evaluation} \\
	68 & 448   & WSI & Multi-Organ          & \\
	69 & 474   & WSI & Multi-Organ          & \\
	70 & 473   & WSI & Multi-Organ          & \\
	71 & 316   & WSI & Multi-Organ          & \\
	72 & 377   & WSI & Multi-Organ          & \\
	73 & 325   & WSI & Multi-Organ          & \\
	74 & 542   & WSI & Multi-Organ          & \\
	75 & 527   & WSI & Multi-Organ          & \\
	76 & 471   & WSI & Multi-Organ          & \\
	77 & 425   & WSI & Multi-Organ          & \\
	78 & 124   & WSI & Multi-Organ          & \\
	79 & 133   & WSI & Multi-Organ          & \\
	80 & 91    & WSI & Multi-Organ          & \\
	81 & 266   & WSI & Multi-Organ          & \\
	82 & 161   & WSI & Multi-Organ          & \\
	83 & 143   & WSI & Multi-Organ          & \\
	84 & 120   & WSI & Multi-Organ          & \\
	85 & 293    & WSI & Multi-Organ          & \\
	86 & 88    & WSI & Multi-Organ          & \\
	87 & 80    & WSI & Multi-Organ          & \\
	88 & 80    & WSI & Multi-Organ          & \\
	89 & 56    & WSI & Multi-Organ          & \\
	90 & 69    & WSI & Multi-Organ          & \\
	91 & 94    & WSI & Multi-Organ          & \\
	92 & 62    & WSI & Multi-Organ          & \\
	93 & 46    & WSI & Multi-Organ          & \\
	94 & 46    & WSI & Multi-Organ          & \\
	95 & 52    & WSI & Multi-Organ          & \\
	96 & 23    & WSI & Multi-Organ          & \\
	97  & 10,150  & ROI & Lymph Nodes         & \\
	98  & 588    & ROI & Breast                 & \\
	99  & 1,132   & ROI & Breast                 & \\
	100 & 714    & ROI & Breast                 & \\
	101 & 922    & ROI & Breast                 & \\
	102 & 832    & ROI & Breast                 & \\
	\midrule
	103 & 71     & WSI & Multi-Disease          & \multirow{10}{*}{Gene Expression} \\
	104 & 4      & WSI & Breast                 & \\
	105 & 23     & WSI & Prostate               & \\
	106 & 2      & WSI & Skin                   & \\
	107 & 6      & WSI & Colon                  & \\
	108 & 4      & WSI & Rectum                 & \\
	109 & 24     & WSI & Kidney                 & \\
	110 & 2      & WSI & Liver                  & \\
	111 & 2      & WSI & Lung                   & \\
	112 & 4      & WSI & Lymph Nodes   & \\
\end{longtable}

\clearpage

\begin{table}[h]
	\centering
	\begin{tabular}{p{7.5cm}|p{3cm}}
		\toprule
		Hyper-parameter & Value \\
		\midrule
		Layers & 24 \\
		Heads & 16 \\
		Patch size & 16 \\
		FFN layer & MLP \\
		Head activation & GELU \\
		Embedding dimension & 1024 \\
		Stochastic dropout rate & 0.1 \\
		\midrule
		Global crop scale & 0.48, 1.0 \\
		Global crop number \& size & 2, 224 \\
		Local crop scale & 0.16, 0.48 \\
		Local crop number \& size & 8, 96 \\
		Max masking ratio & 0.5 \\
		Min masking ratio & 0.1 \\
		Gradient clipping max norm & 3.0 \\
		Normalize last layer & yes \\
		Shared head & none \\
		\midrule
		AdamW $\beta$ & (0.9, 0.999) \\
		Batch size & 3072 \\
		Freeze last layer epochs & 1 \\
		Warmup epochs & 2 \\
		Warmup teacher temperature epochs & 6 \\
		Max Epochs & 20 \\
		Learning rate schedule & Cosine \\
		Learning rate (start) & 0 \\
		Learning rate (post warmup) & 2e-3 \\
		Learning rate (final) & 1e-6 \\
		Teacher temperature (start) & 0.04 \\
		Teacher temperature (final) & 0.4 \\
		Teacher momentum (start) & 0.992 \\
		Teacher momentum (final) & 1.000 \\
		Weight decay (start) & 0.04 \\
		Weight decay (end) & 0.4 \\
		Automatic mixed precision & FP16 \\
		\bottomrule
	\end{tabular}
	\caption{\textbf{Hyperparameters used in PathOrchestra pretraining}. 32 $\times$ 80GB NVIDIA A100 GPUs were used for training. Batch size refers to the total batch size across GPUs.}
	\label{tab:hparams_dinov2}
\end{table}

\end{document}